\documentclass{article}
\usepackage{ijcai26}

\usepackage{times}
\usepackage{soul}
\usepackage{url}
\usepackage[hidelinks]{hyperref}
\usepackage[utf8]{inputenc}
\usepackage[small]{caption}
\usepackage{graphicx}
\usepackage{amsmath}
\usepackage{amsthm}
\usepackage{booktabs}
\usepackage{algorithm}
\usepackage[switch]{lineno}
\usepackage{algorithm}
\usepackage{algpseudocode}
\usepackage{caption}
\usepackage[T1]{fontenc}    % use 8-bit T1 fonts
\usepackage{hyperref}       % hyperlinks
\usepackage{url}            % simple URL typesetting
\usepackage{booktabs}       % professional-quality tables
\usepackage{amsfonts}       % blackboard math symbols
\usepackage{nicefrac}       % compact symbols for 1/2, etc.
\usepackage{microtype}      % microtypography
\usepackage{xcolor}         % colors
\usepackage{amsmath}
\usepackage{multirow}
\usepackage{colortbl}
\usepackage{pifont}
\usepackage{wrapfig}
\usepackage{float}
\usepackage{tabularx}
\usepackage{xcolor}
\usepackage{makecell}
\definecolor{fuchsia}{rgb}{0.752, 0.173, 0.220}
\usepackage{tcolorbox}
\definecolor{color1}{HTML}{FF0000} % 红色
\definecolor{color2}{HTML}{00FF00} % 绿色
\definecolor{color3}{HTML}{0000FF} % 蓝色
\definecolor{color4}{HTML}{FF00FF} % 紫色
\definecolor{color5}{HTML}{FFA500} % 橙色
\definecolor{color6}{HTML}{808080} % 灰色
\usepackage{soul}
\definecolor{bestcolor}{HTML}{BBDEFB}
\definecolor{secondbestcolor}{HTML}{E0F7FF}
\newcommand{\mse}[2]{\ensuremath{{\text{#1}}_{\textcolor{gray}{\pm\,#2}}}}
\newcommand{\correspondingmark}{\textsuperscript{\dag}}

\usepackage{subcaption}
\title{Cochain: Balancing Insufficient and Excessive Collaboration in LLM Agent Workflows}

\author{
Jiaxing Zhao$^1$\textsuperscript{*}
\and
Hongbin Xie$^2$\textsuperscript{*}\and
Yuzhen Lei$^1$\and
Xuan Song$^{1,2}$\correspondingmark \\
Zhuoran Shi$^2$\and
Lianxin Li$^2$\and
Shuangxue Liu$^1$\and
Linguo Xie$^3$\And
Haoran Zhang$^3$\correspondingmark
\affiliations
$^1$School of Artificial Intelligence, Jilin University\\
$^2$Department of Computer Science and Engineering, Southern University of Science and Technology\\
$^3$School of Urban Planning and Design, Peking University\\
\emails
\{jiaxing25, leiyz25, sxliu25\}@mails.jlu.edu.cn,
\{12131108, 12210702, 12212055\}@mail.sustech.edu.cn,
songxuan@jlu.edu.cn,
\{xielinguo, h.zhang\}@pku.edu.cn
}

\begin{document}

\maketitle

\begingroup
\renewcommand\thefootnote{*} % 将当前脚注标记设置为星号
\footnotetext{Equal contribution.} % 星号对应的脚注内容

\renewcommand\thefootnote{\dag} % 将当前脚注标记设置为剑号
\footnotetext{Corresponding author.} % 剑号对应的脚注内容
\endgroup

\begin{abstract}
Large Language Models (LLMs) have demonstrated impressive performance in executing complex reasoning tasks. Chain-of-thought effectively enhances reasoning capabilities by unlocking the potential of large models, while multi-agent systems provide more comprehensive solutions by integrating the collective intelligence of multiple agents. However, both approaches face significant limitations. Single-agent with chain-of-thought, due to the inherent complexity of designing cross-domain prompts, faces collaboration challenges. Meanwhile, multi-agent systems consume substantial tokens and inevitably dilute the primary problem, which is particularly problematic in business workflow tasks. To address these challenges, we propose \textbf{Cochain}, a collaboration prompting framework that effectively solves the business workflow collaboration problem by combining knowledge and prompts at a reduced cost. Specifically, we construct an integrated knowledge graph that incorporates knowledge from multiple stages. Furthermore, by maintaining and retrieving a prompts tree, we can obtain prompt information relevant to other stages of the business workflow. We perform extensive evaluations of Cochain across multiple datasets, demonstrating that Cochain outperforms all baselines in both prompt engineering and multi-agent LLMs. Additionally, expert evaluation results indicate that the use of a small model in combination with Cochain outperforms GPT-4.
\end{abstract}

\section{Introduction}

\begin{figure}[t!]
    \centering
    \includegraphics[width=\linewidth]{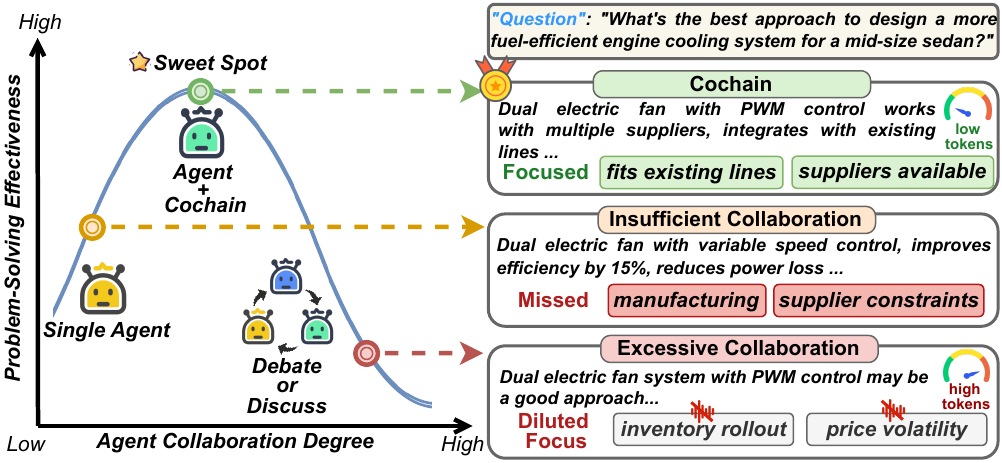}
    \caption{Collaboration has a Goldilocks region. With Insufficient Collaboration, a system overlooks cross-stage constraints and dependencies. With Excessive Collaboration, interaction injects peripheral details that dilute the core decision and increase token cost. Cochain targets effective collaboration by enforcing stage constraints while keeping the solution focused. We illustrate this with a car manufacturing query, where a single agent misses manufacturing and supplier constraints, and excessive discussion dilutes focus on issues such as inventory rollout and price volatility.}
    \label{teaser}
\end{figure}

Large Language Models (LLMs) have achieved strong performance on language understanding and complex reasoning tasks~\cite{touvron2023llama,bai2023qwen,glm2024chatglm,bi2024deepseek}. Much of this progress is unlocked at inference time. Prompting techniques such as Chain-of-Thought encourage step-by-step reasoning and often improve depth of reasoning~\cite{wei2022chain,besta2024graph}. In parallel, multi-agent frameworks extend a single model into a team of specialized roles and interaction protocols, aiming to provide broader coverage across domains such as healthcare, education, law, and finance~\cite{tang2023medagents,dan2023educhat,cui2024chatlaw,yang2023fingpt}. However, a practical question remains: How much collaboration is actually needed for high-quality decision making?

This question is especially salient in business workflow tasks. We define a \emph{Business Workflow} as a complex process consisting of many interconnected stages. The stages are specialized, and they strongly depend on each other. A decision made in one stage requires more than local expertise. It must anticipate and incorporate knowledge and constraints from other stages. For example, in an automotive manufacturing workflow, design and production choices must align with what the supply chain can actually provide. In this setting, we observe two opposite failure modes (Figure~\ref{teaser}). \textbf{Insufficient Collaboration} arises when the overall system does not effectively integrate downstream constraints into the decision process, so solutions may look locally reasonable but violate workflow level requirements~\cite{sahoo2024systematic}. \textbf{Excessive Collaboration} arises when many agents participate directly and simultaneously in each decision, making interaction expensive and unfocused, so non critical information overwhelms the main issue and the system may prefer consensus-like answers over accurate ones~\cite{du2023improving,zhang2024coa}. Existing research on multi-agent systems primarily focuses on maximizing collaboration, extensively exploring how multi-agent systems improve decision quality~\cite{zhang2025collm}, make safe decisions~\cite{piatti2024cooperate}, and solve complex problems~\cite{li2025parallelized}. In contrast, the negative effects of excessive collaboration have received insufficient attention, particularly in multi-stage business workflows~\cite{lei2025mscorebenchmarkmultistagecollaborative}. \textbf{To our best knowledge, no prior work has systematically analyzed and defined this \emph{Excessive Collaboration} phenomenon in workflow settings, nor proposed targeted solutions to mitigate it.} We therefore seek effective collaboration that is stage aware and focused, without paying the full cost of direct multi-agent chatting.

\begin{figure*}[t]
    \centering
    \includegraphics[width=0.95\linewidth]{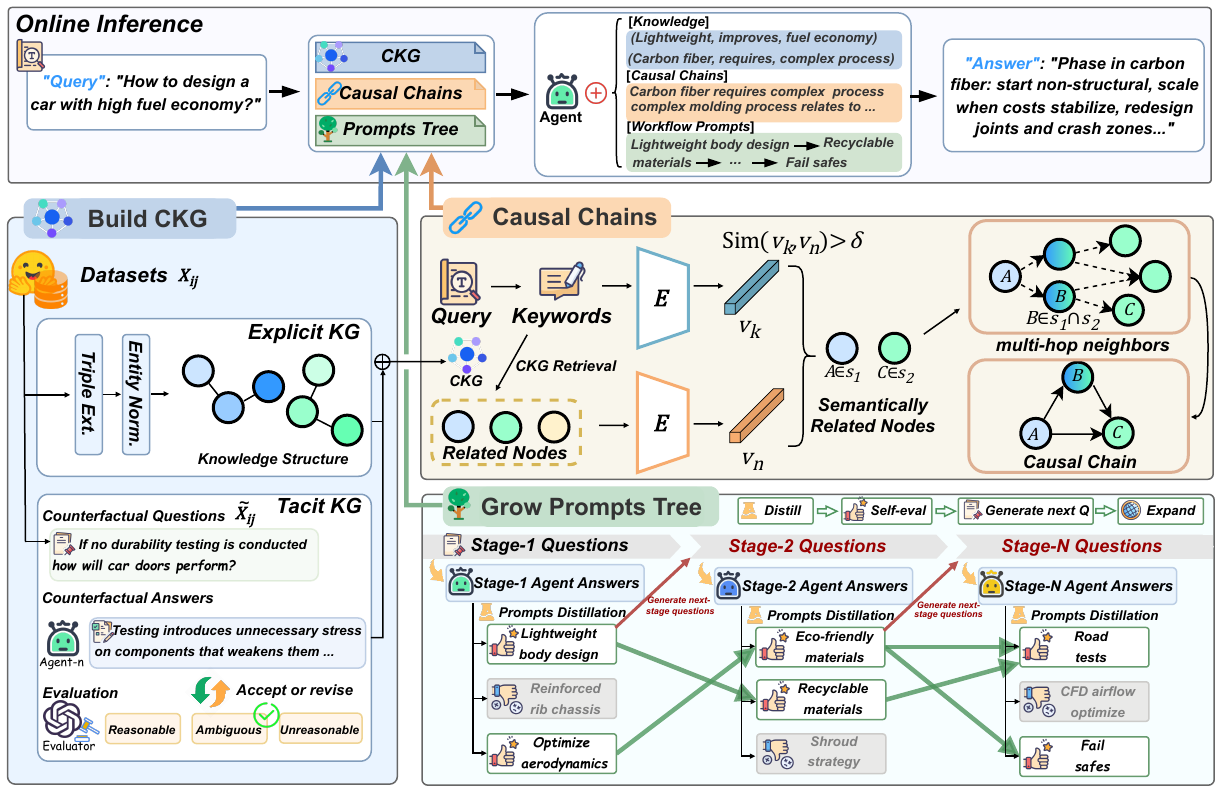}
\caption{Overview of Cochain. Offline, we construct a Collaborative Knowledge Graph (CKG) by combining explicit triples from workflow data with tacit triples elicited via counterfactual querying and iterative verification, and build a Prompts Tree through cross-stage prompt distillation. Online, Cochain retrieves stage-relevant knowledge from the CKG, composes causal chains, and leverages the Prompts Tree to propagate necessary constraints across stages, mitigating insufficient collaboration and excessive collaboration.}
    \label{pipeline}
\end{figure*}

To this end, we propose \textbf{Cochain}, a chain-of-collaboration prompting framework for business workflows. Cochain replaces token-intensive direct interaction with retrieval of reusable collaboration artifacts distilled from agents and data (Figure~\ref{pipeline}). It constructs a \textbf{collaborative knowledge graph} that fuses explicit knowledge from the original datasets with tacit knowledge elicited from specialized agents via counterfactual questioning. To mitigate the influence of irrelevant knowledge~\cite{zhou2024multifaceteval}, retrieved evidence is organized into compact \textbf{causal chains} that connect stage-specific information through shared bridge entities, supporting cross-stage reasoning while filtering noise. Cochain also builds a \textbf{prompts tree} that distills reusable, solution-oriented prompts from agent responses and organizes them across workflow stages. Inspired by how the human brain integrates fragmented knowledge into coherent thought~\cite{courellis2024abstract}, the prompt tree is designed to provide structured cross-stage guidance: querying the tree yields a \textbf{prompt chain} tailored to the input, guiding the backbone model through stage-specific reasoning in a consistent order. At inference time, Cochain retrieves stage-relevant knowledge, a causal chain from the knowledge graph, and a prompt chain from the prompt tree, then composes them into a structured prompt for the backbone LLM. This produces workflow-aware yet focused answers, balancing insufficient collaboration and excessive collaboration. 

The main contributions are summarized as follows:

\begin{itemize}
\item \textbf{Effective collaboration in business workflows.} We identify two opposite collaboration failures in business workflow tasks, insufficient collaboration that misses cross-stage constraints and excessive collaboration that dilutes the core decision with excessive interaction.
\item \textbf{Cochain framework.} We propose Cochain, which enables stage-aware collaboration via reusable artifacts rather than token-heavy interactions. Cochain combines a collaborative knowledge graph with causal-chain retrieval and a prompts tree that composes cross-stage prompt chains for coherent multi-stage reasoning.
\item \textbf{Empirical evidence and expert support.} Experiments across benchmarks and backbones show that Cochain improves quality with efficient inference. Expert evaluation further suggests a smaller model with Cochain can outperform a larger one. We will release our collaborative knowledge graph to support future research.
\end{itemize}

\section{Related Work}
%\begin{figure}[t]
%    \centering
%    \includegraphics[width=\linewidth]{figuers/pipeline.pdf}
%    \caption{Cochain connects business workflow stages. In part (a), agents address counterfactual inquiries, with answers undergoing interactive iteration. In part (b), a prompts tree is built. Part (d) presents our framework: an agent uses the prompts tree and a collaborative knowledge graph built from counterfactual answers and original data, achieving more collaborative results than part (c).}
%    \label{pipeline}
%\end{figure}

\subsection{Knowledge Graph Augmented LLMs}

LLM reasoning capabilities are critical for high-quality responses~\cite{jain2023mechanistically}. Knowledge graphs, with their structured, explicit, and interpretable nature~\cite{zhou2025reflection}, provide clear representations and transparent reasoning paths, guiding LLMs toward deeper reasoning and mitigating limitations~\cite{wang2024knowledge}. For domain-specific LLMs, knowledge graphs enable continuous knowledge updates to accommodate evolving information~\cite{lavrinovics2025knowledge}. Applications include MindMap~\cite{wen2023mindmap} in medicine and ChatLAW~\cite{cui2024chatlaw} in law. Business workflows pose an additional challenge of multi-stage, cross-domain knowledge fusion, where irrelevant knowledge can easily overwhelm the core decision and degrade performance~\cite{ouyang2022training}. To address this, Cochain organizes retrieved evidence into causal chains, improving cross-stage understanding while filtering noise.

\subsection{Prompt Engineering}

Prompt engineering guides LLMs to maximize their potential via designed prompts~\cite{sahoo2024systematic}. It enables model adaptation to new tasks through in-context learning and instruction following~\cite{brown2020language,li2023chain}. Well-designed prompts are shown to enhance performance, particularly on complex tasks~\cite{wei2022chain,wang2022self,yao2022react}. CoT~\cite{wei2022chain} enables models to generate intermediate reasoning steps via simple prompts. ToT~\cite{yao2024tree} explores coherent text units as intermediate problem-solving steps. CPO~\cite{zhang2024chain} leverages non-optimal reasoning paths from tree search to speed up inference. However, these methods are typically designed for a single decision context and do not explicitly maintain constraint awareness across stages or support cross-domain coordination in business workflows. Cochain complements this line by distilling cross-stage prompts from agent responses and organizing them into prompts tree for retrieval.

\subsection{Collaboration of LLM Agents}
Many studies in LLM research solve problems through multi-agent collaboration~\cite{du2023improving,zhao2024longagent}. Common paradigms include discussion~\cite{tang2023medagents} and debate~\cite{du2023improving} to enhance reasoning, hierarchical structures with specialized roles~\cite{zhang2025planning}, and sequential or tree-structured architectures~\cite{zhang2024coa,zhao2024longagent}. These have proven effective in domains like medicine~\cite{tang2023medagents}, long-text processing~\cite{zhang2024coa,zhao2024longagent}, social simulation~\cite{wei2024editable}, and code intelligence~\cite{huang2023agentcoder}. Most existing systems rely on richer interaction and repeated message exchanges, which increases token cost and can dilute the core decision in multi-stage workflows when peripheral opinions accumulate. Cochain instead uses reusable collaboration artifacts to deliver focused, stage-aware collaboration while keeping inference lightweight. To our best knowledge, no prior work has systematically defined \emph{Excessive Collaboration} in this setting or proposed a targeted mitigation mechanism.

\begin{table*}[t]
    \centering
    \resizebox{\textwidth}{!}{
    \begin{tabular}{@{}llccccccccc@{}}
        \toprule
        \multirow{2}{*}{Backbone} & \multirow{2}{*}{Baseline} & \multicolumn{3}{c}{Automotive} & \multicolumn{3}{c}{Pharmaceutical} & \multicolumn{3}{c}{E-commerce} \\
        \cmidrule(lr){3-5} \cmidrule(lr){6-8} \cmidrule(lr){9-11}
        & & BS-F & GLEU & ROUGE-L & BS-F & GLEU & ROUGE-L & BS-F & GLEU & ROUGE-L \\
        \midrule

        % =========================
        % Pangu-1B (new)
        % =========================
        \multirow{6}{*}{openPangu-1B}
        & PMC           & \cellcolor[HTML]{E0F7FF}\mse{72.28}{0.08} & \cellcolor[HTML]{E0F7FF}\mse{12.62}{0.10} & \cellcolor[HTML]{E0F7FF}\mse{22.89}{0.14} & \mse{75.67}{0.31} & \mse{20.66}{0.44} & \mse{29.74}{0.26} & \mse{73.96}{0.19} & \mse{11.77}{0.40} & \mse{19.94}{0.43} \\
        & MedAgents     & \mse{69.95}{0.22} & \mse{10.40}{0.31} & \mse{20.22}{0.25} & \mse{72.82}{0.17} & \mse{17.72}{0.31} & \mse{27.12}{0.37} & \mse{74.63}{0.16} & \mse{21.22}{0.34} & \mse{28.99}{0.44} \\
        & Debate(short) & \mse{71.36}{0.11} & \mse{10.72}{0.23} & \mse{21.45}{0.17} & \cellcolor[HTML]{E0F7FF}\mse{75.74}{0.41} & \cellcolor[HTML]{E0F7FF}\mse{21.87}{0.21} & \cellcolor[HTML]{E0F7FF}\mse{30.11}{0.27} & \cellcolor[HTML]{E0F7FF}\mse{75.73}{0.33} & \cellcolor[HTML]{BBDEFB}\mse{28.05}{0.28} & \cellcolor[HTML]{E0F7FF}\mse{31.05}{0.21} \\
        & Debate(long)  & \mse{70.59}{0.20} & \mse{10.21}{0.27} & \mse{20.85}{0.19} & \mse{73.77}{0.37} & \mse{18.13}{0.23} & \mse{27.61}{0.15} & \mse{75.38}{0.44} & \mse{16.98}{0.34} & \mse{25.27}{0.29} \\
        & CoA           & \mse{69.61}{0.31} & \mse{8.54}{0.47} & \mse{18.60}{0.29} & \mse{73.82}{0.33} & \mse{18.79}{0.21} & \mse{28.00}{0.33} & \mse{74.75}{0.27} & \mse{17.32}{0.28} & \mse{24.10}{0.31} \\
        & Cochain       & \cellcolor[HTML]{BBDEFB}\mse{74.35}{0.05} & \cellcolor[HTML]{BBDEFB}\mse{15.99}{0.07} & \cellcolor[HTML]{BBDEFB}\mse{26.01}{0.03} & \cellcolor[HTML]{BBDEFB}\mse{76.40}{0.10} & \cellcolor[HTML]{BBDEFB}\mse{22.10}{0.08} & \cellcolor[HTML]{BBDEFB}\mse{30.71}{0.13} & \cellcolor[HTML]{BBDEFB}\mse{77.06}{0.04} & \cellcolor[HTML]{E0F7FF}\mse{23.37}{0.14} & \cellcolor[HTML]{BBDEFB}\mse{31.64}{0.11} \\
        \midrule

        \multirow{6}{*}{openPangu-7B}
        & PMC           & \cellcolor[HTML]{E0F7FF}\mse{71.41}{0.13} & \cellcolor[HTML]{E0F7FF}\mse{11.06}{0.17} & \cellcolor[HTML]{E0F7FF}\mse{21.01}{0.21} & \mse{70.65}{0.41} & \mse{15.77}{0.39} & \mse{23.42}{0.30} & \mse{71.77}{0.23} & \cellcolor[HTML]{E0F7FF}\mse{19.55}{0.39} & \cellcolor[HTML]{BBDEFB}\mse{25.97}{0.30} \\
        & MedAgents     & \mse{70.00}{0.24} & \mse{10.63}{0.28} & \mse{20.26}{0.30} & \cellcolor[HTML]{BBDEFB}\mse{73.06}{0.26} & \cellcolor[HTML]{E0F7FF}\mse{18.20}{0.21} & \cellcolor[HTML]{BBDEFB}\mse{25.35}{0.52} & \mse{71.69}{0.18} & \mse{19.45}{0.36} & \mse{22.89}{0.53} \\
        & Debate(short) & \mse{70.11}{0.18} & \mse{10.00}{0.21} & \mse{19.78}{0.34} & \mse{70.73}{0.35} & \mse{12.34}{0.26} & \mse{20.77}{0.35} & \mse{68.56}{0.29} & \mse{18.15}{0.23} & \mse{19.62}{0.32} \\
        & Debate(long)  & \mse{69.90}{0.18} & \mse{9.43}{0.21} & \mse{19.56}{0.29} & \mse{70.67}{0.41} & \mse{13.83}{0.22} & \mse{20.73}{0.31} & \mse{68.42}{0.27} & \mse{17.43}{0.36} & \mse{19.44}{0.27} \\
        & CoA           & \mse{68.25}{0.34} & \mse{7.95}{0.35} & \mse{17.32}{0.43} & \mse{70.21}{0.40} & \mse{14.53}{0.26} & \mse{22.98}{0.24} & \mse{70.03}{0.25} & \mse{19.09}{0.29} & \mse{22.80}{0.23} \\
        & Cochain       & \cellcolor[HTML]{BBDEFB}\mse{74.17}{0.08} & \cellcolor[HTML]{BBDEFB}\mse{15.49}{0.05} & \cellcolor[HTML]{BBDEFB}\mse{25.62}{0.09} & \cellcolor[HTML]{E0F7FF}\mse{70.93}{0.09} & \cellcolor[HTML]{BBDEFB}\mse{19.40}{0.05} & \cellcolor[HTML]{E0F7FF}\mse{25.21}{0.11} & \cellcolor[HTML]{BBDEFB}\mse{73.83}{0.07} & \cellcolor[HTML]{BBDEFB}\mse{19.60}{0.25} & \cellcolor[HTML]{E0F7FF}\mse{23.00}{0.11} \\
        \midrule
        
        \multirow{6}{*}{Qwen2-7B}
        & PMC & \mse{65.86}{0.03} & \mse{13.23}{0.08} & \mse{18.35}{0.07} & \mse{66.97}{0.79} & \mse{16.25}{0.58} & \mse{24.75}{0.62} & \mse{71.02}{0.17} & \mse{24.18}{0.37} & \mse{29.93}{0.45} \\
        & MedAgents & \mse{65.16}{0.25} & \mse{12.37}{0.17} & \mse{18.33}{0.18} & \mse{65.84}{0.15} & \mse{10.68}{0.36} & \mse{20.83}{0.50} & \mse{69.45}{0.19} & \mse{17.42}{0.17} & \mse{33.38}{0.18} \\
        & Debate(short) & \mse{65.30}{0.26} & \mse{12.42}{0.37} & \mse{17.63}{0.23} & \mse{70.62}{0.41} & \mse{16.04}{0.35} & \mse{24.12}{0.49} & \mse{73.03}{0.43} & \mse{20.89}{0.32} & \mse{28.27}{0.29} \\
        & Debate(long) & \mse{65.61}{0.27} & \mse{12.85}{0.47} & \mse{18.28}{0.13} & \mse{60.57}{0.30} & \mse{10.28}{0.39} & \mse{11.21}{0.52} & \mse{72.41}{0.60} & \mse{20.82}{0.59} & \mse{28.46}{0.53} \\
        & CoA & \cellcolor[HTML]{E0F7FF}\mse{70.70}{0.53} & \cellcolor[HTML]{E0F7FF}\mse{21.51}{0.41} & \cellcolor[HTML]{E0F7FF}\mse{22.65}{0.68} & \cellcolor[HTML]{E0F7FF}\mse{76.98}{0.44} & \cellcolor[HTML]{E0F7FF}\mse{30.45}{0.42} & \cellcolor[HTML]{E0F7FF}\mse{34.00}{0.17} & \cellcolor[HTML]{E0F7FF}\mse{79.24}{0.18} & \cellcolor[HTML]{BBDEFB}\mse{38.60}{0.35} & \cellcolor[HTML]{E0F7FF}\mse{39.14}{0.62} \\
        & Cochain & \cellcolor[HTML]{BBDEFB}\mse{75.05}{0.17} & \cellcolor[HTML]{BBDEFB}\mse{27.66}{0.26} & \cellcolor[HTML]{BBDEFB}\mse{28.43}{0.23} & \cellcolor[HTML]{BBDEFB}\mse{78.50}{0.04} & \cellcolor[HTML]{BBDEFB}\mse{34.48}{0.28} & \cellcolor[HTML]{BBDEFB}\mse{35.23}{0.17} & \cellcolor[HTML]{BBDEFB}\mse{80.22}{0.16} & \cellcolor[HTML]{E0F7FF}\mse{37.83}{0.38} & \cellcolor[HTML]{BBDEFB}\mse{40.84}{0.41} \\
        \midrule

        \multirow{6}{*}{DeepSeek-R1-7B}
        & PMC & \mse{65.94}{0.13} & \mse{10.75}{0.23} & \mse{15.78}{0.13} & \mse{68.06}{0.27} & \mse{22.51}{0.29} & \mse{28.07}{0.20} & \mse{66.66}{0.43} & \mse{15.80}{0.39} & \mse{23.62}{0.30} \\
        & MedAgents & \mse{65.58}{0.60} & \mse{8.03}{0.55} & \mse{14.08}{0.59} & \mse{70.35}{0.14} & \mse{24.47}{0.19} & \mse{30.90}{0.08} & \mse{66.98}{0.49} & \mse{13.67}{0.49} & \mse{20.31}{0.60} \\
        & Debate(short) & \mse{65.93}{0.23} & \mse{11.41}{0.44} & \mse{17.33}{0.52} & \mse{71.46}{0.31} & \mse{27.13}{0.27} & \mse{30.34}{0.44} & \mse{70.71}{0.29} & \mse{23.01}{0.22} & \mse{27.02}{0.39} \\
        & Debate(long) & \mse{66.06}{0.67} & \mse{12.13}{0.49} & \cellcolor[HTML]{E0F7FF}\mse{17.70}{0.26} & \mse{71.54}{0.42} & \mse{27.34}{0.39} & \mse{30.04}{0.39} & \mse{70.48}{0.39} & \mse{23.07}{0.10} & \mse{27.00}{0.44} \\
        & CoA & \cellcolor[HTML]{E0F7FF}\mse{70.83}{0.29} & \cellcolor[HTML]{E0F7FF}\mse{13.27}{0.58} & \mse{17.51}{0.58} & \cellcolor[HTML]{E0F7FF}\mse{77.06}{0.15} & \cellcolor[HTML]{E0F7FF}\mse{38.18}{0.49} & \cellcolor[HTML]{E0F7FF}\mse{35.53}{0.23} & \cellcolor[HTML]{E0F7FF}\mse{80.43}{0.32} & \cellcolor[HTML]{E0F7FF}\mse{43.80}{0.39} & \cellcolor[HTML]{E0F7FF}\mse{42.03}{0.43} \\
        & Cochain & \cellcolor[HTML]{BBDEFB}\mse{73.20}{0.02} & \cellcolor[HTML]{BBDEFB}\mse{21.45}{0.17} & \cellcolor[HTML]{BBDEFB}\mse{23.89}{0.13} & \cellcolor[HTML]{BBDEFB}\mse{79.11}{0.09} & \cellcolor[HTML]{BBDEFB}\mse{38.27}{0.17} & \cellcolor[HTML]{BBDEFB}\mse{37.21}{0.21} & \cellcolor[HTML]{BBDEFB}\mse{82.68}{0.05} & \cellcolor[HTML]{BBDEFB}\mse{50.19}{0.13} & \cellcolor[HTML]{BBDEFB}\mse{47.07}{0.13} \\
        \midrule

        \multirow{6}{*}{Claude-3.5-haiku}
        & PMC & \mse{64.06}{0.04} & \mse{9.73}{0.56} & \mse{10.12}{0.31} & \mse{66.18}{0.16} & \mse{13.04}{0.36} & \mse{13.91}{0.18} & \mse{66.85}{0.07} & \mse{15.98}{0.60} & \mse{15.88}{0.21} \\
        & MedAgents & \mse{64.38}{0.10} & \mse{10.77}{0.56} & \mse{11.02}{0.14} & \mse{66.01}{0.17} & \mse{14.60}{0.30} & \mse{14.66}{0.29} & \mse{68.00}{0.13} & \mse{17.58}{0.31} & \mse{16.51}{0.14} \\
        & Debate(short) & \mse{64.36}{0.17} & \mse{11.21}{0.39} & \cellcolor[HTML]{E0F7FF}\mse{13.72}{0.33} & \cellcolor[HTML]{E0F7FF}\mse{69.11}{0.11} & \mse{14.53}{0.26} & \cellcolor[HTML]{E0F7FF}\mse{16.03}{0.28} & \cellcolor[HTML]{E0F7FF}\mse{68.41}{0.43} & \cellcolor[HTML]{E0F7FF}\mse{19.97}{0.32} & \mse{18.37}{0.31} \\
        & Debate(long) & \cellcolor[HTML]{E0F7FF}\mse{65.13}{0.08} & \cellcolor[HTML]{E0F7FF}\mse{11.79}{0.49} & \mse{12.13}{0.39} & \mse{67.68}{0.10} & \cellcolor[HTML]{E0F7FF}\mse{15.94}{0.31} & \mse{15.13}{0.32} & \mse{67.49}{0.67} & \mse{17.23}{0.42} & \cellcolor[HTML]{E0F7FF}\mse{18.98}{0.56} \\
        & CoA & \mse{63.78}{0.26} & \mse{9.23}{0.29} & \mse{10.28}{0.20} & \mse{66.00}{0.39} & \mse{13.33}{0.22} & \mse{14.25}{0.36} & \mse{67.03}{0.35} & \mse{15.02}{0.33} & \mse{15.12}{0.35} \\
        & Cochain & \cellcolor[HTML]{BBDEFB}\mse{69.66}{0.03} & \cellcolor[HTML]{BBDEFB}\mse{18.62}{0.06} & \cellcolor[HTML]{BBDEFB}\mse{16.83}{0.24} & \cellcolor[HTML]{BBDEFB}\mse{72.83}{0.06} & \cellcolor[HTML]{BBDEFB}\mse{23.47}{0.13} & \cellcolor[HTML]{BBDEFB}\mse{19.25}{0.22} & \cellcolor[HTML]{BBDEFB}\mse{71.56}{0.12} & \cellcolor[HTML]{BBDEFB}\mse{20.11}{0.13} & \cellcolor[HTML]{BBDEFB}\mse{20.65}{0.17} \\
        \bottomrule
    \end{tabular}}
        \caption{Performance comparison of different model backbones and multi-agent baselines across three business domains. We report the mean and standard error (SE) over five experiments.  BERTScore F1 is abbreviated as BS-F. We highlight the \sethlcolor{bestcolor}\hl{best} and \sethlcolor{secondbestcolor}\hl{second-best} results. }
    \label{tab:combined-performance}
\end{table*}

\section{Methods}

\subsection{Collaborative Knowledge Graph}
Building a collaborative knowledge graph requires abundant stage-specific knowledge.
We obtain it from the training sets of the workflow datasets, denoted as
\( D_i = \{(X_{ij}, Y_{ij})\}_{j=1}^{n_i} \),
where \(i \in \{1,\dots,N\}\) indexes workflow stages, \(N\) is the number of stages,
and \(j \in \{1,\dots,n_i\}\) indexes the \(n_i\) training instances in stage \(i\).
Here \( X_{ij} \in X_i \) is the input and \( Y_{ij} \in Y_i \) is the corresponding output.
After data cleaning and triplet extraction, we construct an explicit knowledge graph \( \mathcal{KG}_{\text{explicit}} \):
\begin{equation}
    \mathcal{KG}_{\text{explicit}} = \bigcup_{i=1}^{N} \bigcup_{j=1}^{n_i} \text{ExtractTriples}(X_{ij}, Y_{ij})
\end{equation}

Business workflows also contain \emph{tacit} knowledge that is rarely stated in the original datasets, such as feasibility constraints, implementation caveats, and stage-specific heuristics. 
Since such considerations are not explicitly available, we elicit tacit workflow knowledge from the agent’s responses by probing the agent under counterfactual perturbations and attributing response variations to unobserved internal considerations.

For each sample \( (X_{ij}, Y_{ij}) \), we generate a counterfactual input \( \tilde{X}_{ij} \) by editing \(X_{ij}\) to perturb one salient condition while preserving intent, using templates for causal, adversarial, substitution, extreme, and backward-causal variants:
\begin{equation}
    \tilde{X}_{ij} = \text{GenerateCounterfactual}(X_{ij})
\end{equation}

In our setting, a vertical domain agent is the stage-specialized LLM assigned to stage \(i\) (fine-tuned on stage-\(i\) data for open-source backbones). We query this agent with \( \tilde{X}_{ij} \) to obtain the counterfactual output \( \tilde{Y}_{ij} \). 
To capture the unobserved considerations that shape how the agent resolves \( \tilde{X}_{ij} \), we introduce a latent variable \( \theta_{ij} \) and an intermediate latent state \( h_{ij} \) influenced by \( \theta_{ij} \), where \( h_{ij} \mid \theta_{ij} \sim P(h_{ij} \mid \theta_{ij}) \), and \( \tilde{Y}_{ij} \mid h_{ij}, \tilde{X}_{ij} \sim P(\tilde{Y}_{ij} \mid h_{ij}, \tilde{X}_{ij}) \). The resulting generation process can be written as:
\begin{equation}
\begin{split}
P(\tilde{Y}_{ij} \mid \tilde{X}_{ij})
&= \int_{\Theta} P(\tilde{Y}_{ij} \mid h_{ij}, \tilde{X}_{ij}) \\
&\quad \cdot P(h_{ij} \mid \theta_{ij}) \, P(\theta_{ij} \mid \tilde{X}_{ij}) \, d\theta_{ij}
\end{split}
\end{equation}

Here, \( P(\tilde{Y}_{ij} \mid h_{ij}, \tilde{X}_{ij}) \) is the likelihood of generating the counterfactual output given \( h_{ij} \) and \( \tilde{X}_{ij} \), \( P(h_{ij} \mid \theta_{ij}) \) describes how tacit considerations shape the latent state, and \( P(\theta_{ij} \mid \tilde{X}_{ij}) \) is a prior over tacit considerations conditioned on the counterfactual input.

To improve reliability, each counterfactual answer is evaluated by a general-purpose LLM and assigned one of three labels: reasonable, ambiguous, or unreasonable. The evaluator also provides concise feedback and revision suggestions, which the vertical agent uses to regenerate a response. This loop repeats until a reasonable answer is obtained. During knowledge distillation, we apply causal enhancement by emphasizing causal cues such as ``depends on'', ``relies on'', and ``applies to'', which helps preserve cross-stage dependency signals. Using the counterfactual input-output pairs \( (\tilde{X}_{ij}, \tilde{Y}_{ij}) \) produced by the refinement loop, we extract tacit triples and build a tacit knowledge graph \( \mathcal{KG}_{\text{tacit}} \):
\begin{equation}
\mathcal{KG}_{\text{tacit}} = \bigcup_{i=1}^{N} \bigcup_{j=1}^{n_i} \text{ExtractTriples}(\tilde{X}_{ij}, \tilde{Y}_{ij}, \theta_{ij})
\end{equation}

Finally, we integrate explicit and tacit knowledge to form the collaborative knowledge graph \( \mathcal{KG} \), which captures both stage-specific knowledge and inter-stage connectivity:
\begin{equation}
\mathcal{CKG} = \mathcal{KG}_{\text{explicit}} \cup \mathcal{KG}_{\text{tacit}}
\end{equation}

\subsection{Causal Chain}

To help the agent organize cross-stage dependencies instead of merely listing retrieved facts, we introduce a causal chain mechanism. Given a user request, we decompose it and extract keywords, which are then used to retrieve relevant nodes from the collaborative knowledge graph.

Following Figure~\ref{pipeline}, we use a pre-trained text encoder~\cite{reimers-2019-sentence-bert} to embed each keyword and candidate knowledge node. For the embedding \( v_k \) of keyword \( k \) and the embedding \( v_n \) of node \( n \), we compute cosine similarity:
\begin{equation}
\text{Sim}(v_k, v_n) = \frac{v_k \cdot v_n}{\|v_k\| \|v_n\|}
\end{equation}

With a similarity threshold \( \delta \), we select nodes that are semantically related to the query. For each selected node, we expand one-hop neighbors to introduce closely associated evidence and form a local triple-based structure. To connect knowledge across workflow stages, we further perform multi-hop exploration through \emph{bridge entities}. Let the workflow stages be \( \mathcal{S} = \{s_1, s_2, s_3, \dots\} \). For a high-similarity node \( A \) belonging to stage \( s_1 \), we expand a neighbor \( B \) that appears in both \( s_1 \) and some other stage \( s_i \) with \( i \neq 1 \). We then use \( B \) as a bridge to reach a node \( C \) in stage \( s_i \), forming a cross-stage causal chain:
\begin{equation}
\text{CausalChain}(A, C) = A \xrightarrow{B} C
\end{equation}

The causal chain provides a compact, stage-connected context that highlights how constraints and decisions propagate across stages, improving coherence and feasibility in the final response.

\begin{table}[t]
\centering
\resizebox{\columnwidth}{!}{
\begin{tabular}{@{}clccc@{}}
\toprule
Backbone & Method & BS-F & GLEU & ROUGE-L \\
\midrule
\multirow{4}{*}{GPT-3.5-turbo}
& + IO      & \mse{72.84}{0.07} & \mse{15.09}{0.17} & \mse{21.98}{0.23} \\
& + CoT     & \mse{71.61}{0.17} & \mse{11.15}{0.17} & \mse{17.92}{0.16} \\
& + ToT     & \mse{66.69}{0.05} & \mse{10.08}{0.04} & \mse{17.74}{0.09} \\
& + Cochain & \textbf{\mse{76.94}{0.03}} & \textbf{\mse{28.92}{0.12}} & \textbf{\mse{30.65}{0.09}} \\
\midrule
\multirow{4}{*}{GPT-4o}
& + IO      & \mse{71.92}{0.08} & \mse{23.05}{0.17} & \mse{23.79}{0.17} \\
& + CoT     & \mse{72.57}{0.15} & \mse{23.76}{0.14} & \mse{24.15}{0.28} \\
& + ToT     & \mse{66.23}{0.17} & \mse{12.51}{0.14} & \mse{17.52}{0.19} \\
& + Cochain & \textbf{\mse{74.46}{0.07}} & \textbf{\mse{25.16}{0.25}} & \textbf{\mse{25.07}{0.14}} \\
\bottomrule
\end{tabular}}
\caption{The performance of GPT-3.5-turbo and GPT-4o when using CoT, ToT, and Cochain.}
\label{tab4}
\end{table}

\begin{table}[t]
    \centering
    \small
    \label{tab:final_comparison_resized}
    \begin{tabular}{lcc}
        \toprule
        \multirow{2}{*}{Baseline} & \multicolumn{2}{c}{Backbone} \\
        \cmidrule(lr){2-3}
        & DeepSeek-V3.1 & Claude-3.5-haiku \\
        \midrule
        PMC       & 0.12 & 0.29 \\
        MedAgents & 0.07 & 0.26 \\
        Debate    & 0.32 & 0.31 \\
        CoA       & 0.27 & 0.37 \\
        \midrule
        Cochain   & \textbf{0.33} & \textbf{0.38} \\
        \bottomrule
    \end{tabular}
    \caption{Accuracy comparison on Auto-SLURP benchmark.}
    \label{Auto-SLURP}
\end{table}

\subsection{Prompts Tree}

In business workflow tasks, effective collaboration depends on whether stage-specific reasoning can be carried forward in a consistent and reusable manner. We propose a prompt distillation procedure that offline constructs a \textbf{Prompts Tree}, which stores reusable, solution-oriented prompts and explicitly encodes cross-stage dependencies. We model the Prompts Tree as a rooted tree \(T=(V,E)\). Each node \(v\in V\) is associated with a stage label \(s(v)\in\mathcal{S}\), a distilled prompt \(p(v)\), and its supporting evidence (the Q\&A from which the prompt is distilled). Each directed edge \((u\!\rightarrow\!v)\in E\) represents a dependency where prompt \(p(u)\) from an upstream stage induces a downstream question template, whose answer leads to prompt \(p(v)\) at a later stage. Querying \(T\) returns a \textbf{prompt chain} along a root-to-leaf path, providing structured, cross-stage guidance during inference.

\textbf{Tree construction.} We start from a seed Q\&A at Stage 1. Given the Stage 1 answer, the Stage 1 agent distills a set of candidate prompts that are solution-oriented and constraint-aware, focusing on actionable decisions, feasibility conditions, and implementation requirements. Each candidate prompt is then self-evaluated for utility and executability, and we retain the top \(m\) prompts. For each retained prompt, we generate a downstream question template that asks how to implement, validate, or satisfy this prompt in the next stage. The next-stage agent answers these questions, producing new Q\&A pairs that serve as seeds for prompt distillation at the next stage. This distill--self-evaluate--generate-next-question--expand loop repeats across stages, causing the tree to grow as upstream prompts branch into multiple downstream realizations.

\textbf{Node and edge semantics.} The user need is treated as the root node. Prompts distilled at Stage 1 become children of the root. For a node \(u\) at stage \(s_k\), each outgoing edge corresponds to a generated question template \(q_{u\rightarrow}\) targeting stage \(s_{k+1}\). The resulting answer yields a child node \(v\) at stage \(s_{k+1}\) with distilled prompt \(p(v)\). This construction explicitly records cross-stage dependencies, since each child prompt is conditioned on satisfying or operationalizing its parent prompt under downstream constraints.

\textbf{Retrieval and usage.} Inspired by how the human brain integrates fragmented knowledge into coherent thought~\cite{courellis2024abstract}, the Prompts Tree organizes fragmented stage-specific guidance into an ordered chain. At inference time, we retrieve a prompt chain \( \pi = (v_0\!\rightarrow v_1\!\rightarrow \cdots \!\rightarrow v_L) \) that best matches the input query, and use the prompts \( \{p(v_\ell)\}_{\ell=0}^{L} \) as structured guidance so the backbone LLM reasons through the workflow stages in a consistent order. This avoids brittle manual prompt construction while improving cross-stage coordination with low token overhead.
\section{Experiments}

%We evaluate Cochain on multi-stage business workflow benchmarks to answer the following questions:
%\textbf{(1)} Does Cochain improve overall performance compared with single-agent prompting and representative multi-agent systems?
%\textbf{(2)} How robust are the gains across datasets, workflow stages, and backbone LLMs, including reasoning-capable and standard models?
%\textbf{(3)} Which components contribute most to the gains, and what is the cost profile in tokens, latency, and API fees?

We study when collaboration helps and when it hurts in multi-stage business workflows, and evaluate whether Cochain provides a practical middle ground via reusable collaboration artifacts. Accordingly, we benchmark overall performance against strong baselines, analyze key design factors and hyperparameters, compare jointly fine-tuned and stage-wise specialized deployment regimes, and report efficiency and cost.
\subsection{Experimental Setup}

\paragraph{Datasets.}
We evaluate Cochain on two benchmarks across four datasets: Auto-SLURP~\cite{shen2025autoslurpbenchmarkdatasetevaluating}, which assesses end-to-end multi-stage workflows, and MSCoRe~\cite{lei2025mscorebenchmarkmultistagecollaborative}, which focuses on multi-stage reasoning and collaboration. The automotive workflow test set is reviewed and refined by domain experts from a major automobile manufacturer to ensure practical industry relevance.

\paragraph{Metrics.}
We report BERTScore~\cite{zhang2019bertscore}, GLEU~\cite{wu2016google}, and ROUGE-L~\cite{lin2004rouge} to measure semantic similarity and generation quality between model outputs and reference answers.

\paragraph{LLMs.}
We use 17 backbone LLMs, covering both proprietary and open-source models. We compare reasoning-capable models (e.g., DeepSeek-R1-7B~\cite{deepseekai2025deepseekr1incentivizingreasoningcapability}) with standard models (e.g., openPangu-7B~\cite{chen2025panguembeddedefficientdualsystem}, Qwen2-7B~\cite{qwen2}). We also study using identical backbones across stages versus mixed backbones (Llama2-7B~\cite{touvron2023llama2openfoundation}, GLM4-9B, Qwen2-7B, DeepSeek-7B) for stage specialization. For open-source backbones, we perform stage-wise fine-tuning on MSCoRe: the agent at stage \(k\) is fine-tuned using only the training split of stage-\(k\) data, without using any validation/test instances. Proprietary models include GPT-3.5-turbo, GPT-4o~\cite{achiam2023gpt}, DeepSeek-V3.1~\cite{deepseekai2024deepseekv3technicalreport}, and Claude-3.5-haiku.

\begin{figure}[t]
  \centering
  \includegraphics[width=\columnwidth]{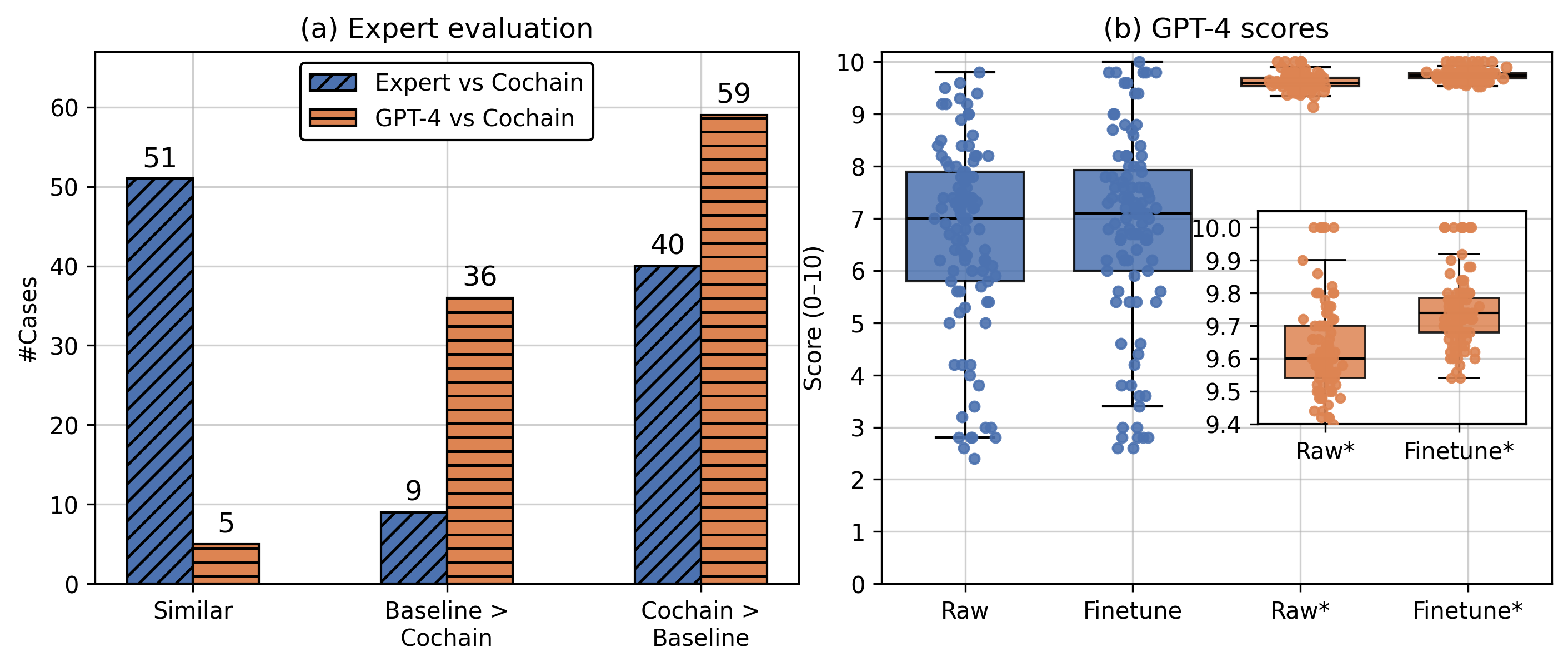}
    \caption{Expert evaluation and GPT-4 scoring results. Panel (a) compares Qwen2-7B augmented with Cochain against GPT-4 and expert-written references (100 cases). In panel (b), the asterisk indicates augmentation with Cochain.}
  \label{expert_evaluation}
\end{figure}

\paragraph{Baselines.}
Our baseline selection encompasses the predominant multi-agent collaboration methodologies. Specifically, PMC~\cite{zhang2025planning} implements multi-agent collaboration through hierarchical planning. MedAgents~\cite{tang2023medagents} facilitates collaborative decision-making via discussion and voting mechanisms. Debate~\cite{du2023improving} enables collaboration through argumentative discourse. CoA~\cite{zhang2024coa} establishes chain-based collaboration. Besides, we compare the performance of single-agent prompt engineering methods, namely CoT~\cite{wei2022chain} and ToT~\cite{yao2024tree}.

% =========================
% Table 2 (no minipage)
% =========================
\begin{table}[t]
\centering
\small
\begin{tabular}{c|c|c|c|c}
\toprule
Skip & Stage & BS-F & GLEU & ROUGE-L \\
\midrule
\multirow{1}{*}{zero} & / & 75.39 & 28.17 & 28.87 \\
\midrule
\multirow{3}{*}{one}
& $S_1 \rightarrow S_3 \rightarrow S_5$ & 74.00 & 24.66 & 25.18 \\
& $S_2 \rightarrow S_4$                 & 74.52 & 25.57 & 26.18 \\
& $S_3 \rightarrow S_5$                 & 74.18 & 25.45 & 25.95 \\
\midrule
\multirow{2}{*}{two}
& $S_1 \rightarrow S_3$ & 74.27 & 25.26 & 25.74 \\
& $S_2 \rightarrow S_4$ & 74.07 & 25.00 & 25.82 \\
\bottomrule
\end{tabular}
\caption{Result of different skip stages on evaluation metrics.}
\label{skip}
\end{table}

\subsection{Result}

%\subsection{Single-Stage Relational Reasoning}
%In Table~\ref{tab:combined-performance}, we evaluate the performance of different baseline large models in the automotive design stage after applying Cochain. After applying Cochain, the original LLMs and the domain-fine-tuned LLMs exhibit significant performance improvements. Especially for Qwen2-7B, Llama3-8B, and Qwen2.5-14B, after applying Cochain, the performance of the original models surpass that of the fine-tuned models that do not apply Cochain, further demonstrating the effectiveness of Cochain.
%In Table~\ref{tab4}, we perform a comparative analysis of different methods. Although both COT and TOT demonstrate improved reasoning abilities, they do not show significant advantages in terms of collaboration. %%%
%人工评估放附录，这里还需要加引用

\paragraph{Overall Results.}
Table~\ref{tab:combined-performance} summarizes performance across three business domains and four backbones. The results support our central claim that effective workflow collaboration requires balancing \emph{Insufficient} and \emph{Excessive} Collaboration. Compared with multi-agent baselines that rely on rich interaction (often leading to excessive collaboration and diluted decisions), Cochain consistently achieves higher quality across domains. Meanwhile, Cochain also improves substantially on a small backbone, indicating that it mitigates insufficient collaboration by injecting stage-aware constraints and dependencies through reusable artifacts rather than relying on stronger backbones alone. For instance, with openPangu-1B on Automotive, Cochain raises BS-F from 72.28 (PMC) to 74.35 and improves ROUGE-L from 22.89 to 26.01, suggesting that the framework helps the agent stay \emph{constraint-aware} while remaining \emph{focused} on the core decision.

\paragraph{Benchmark Transfer on Auto-SLURP.}
Table~\ref{Auto-SLURP} confirms the advantage on an end-to-end workflow benchmark. Cochain achieves the highest accuracy under both DeepSeek-V3.1 and Claude-3.5-haiku backbones, reaching 0.33 and 0.38, respectively. This result supports that Cochain generalizes beyond MSCoRe-style text similarity metrics and remains effective when collaboration must be executed across workflow stages.

\paragraph{Comparison to Prompting Baselines.}
Table~\ref{tab4} compares Cochain with IO, CoT, and ToT on proprietary backbones. While CoT and ToT enhance reasoning in a single-agent setting, they provide limited support for carrying stage constraints across a workflow. Cochain yields stronger generation quality under both GPT-3.5-turbo and GPT-4o. On GPT-3.5-turbo, Cochain increases BS-F from 72.84 (IO) to 76.94 and improves ROUGE-L from 21.98 to 30.65, demonstrating that the benefits go beyond deeper single-path reasoning.

\paragraph{Expert Evaluation and GPT-4 Scoring.}
Figure~\ref{expert_evaluation} provides external validation beyond automatic metrics. In panel (a), domain experts judge Cochain-augmented answers to be better in 40 cases and worse in 9 cases, with 51 cases considered similar. Panel (b) shows higher GPT-4 scores for settings augmented with Cochain, and the fine-tuned augmented setting reaches near-ceiling scores, corroborating the expert preference trend.

% -------------------------
% Figure 1: Parameter analysis
% -------------------------
\begin{figure}[t]
  \centering
  \includegraphics[width=\columnwidth]{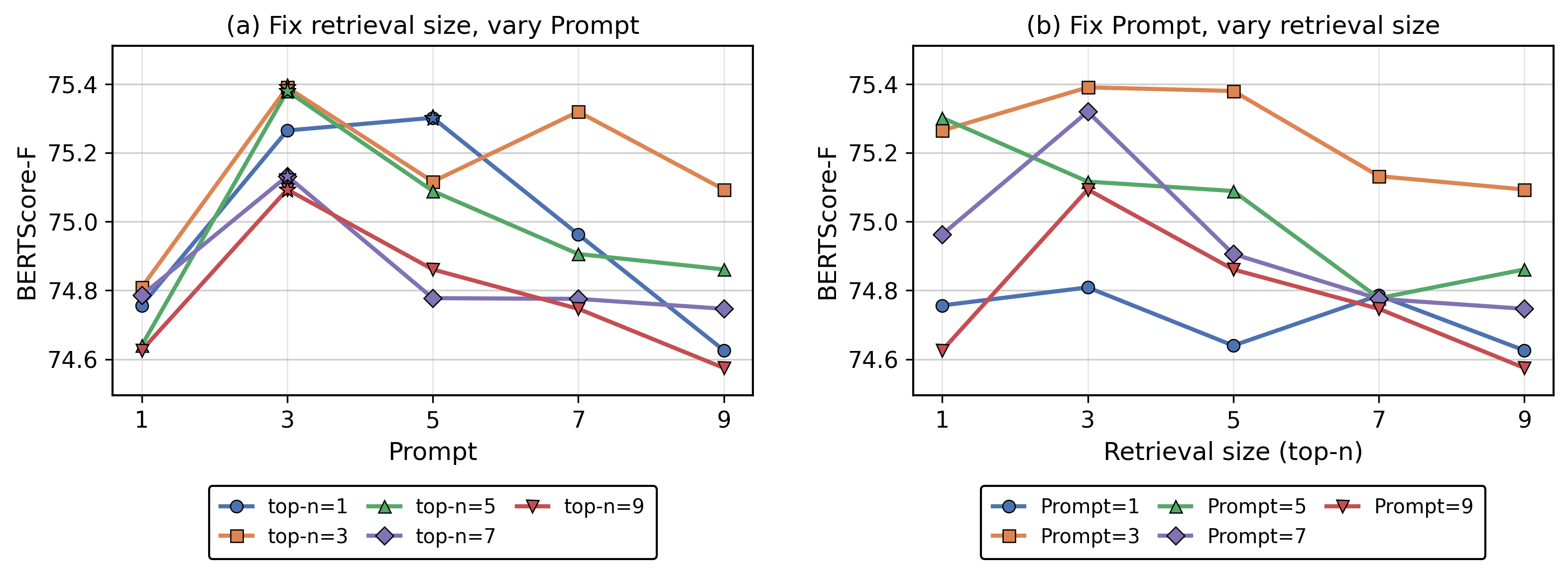}
  \caption{Hyperparameter influence of prompt budget and retrieval size. (a) Fix top-$n$ retrieved nodes and vary prompt count. (b) Fix prompt count and vary top-$n$.}
  \label{fig:parameter_analysis}
\end{figure}

% -------------------------
% Figure 2: Model scaling (7B -> 70B)
% -------------------------
\begin{figure}[t]
  \centering
  \includegraphics[width=0.85\columnwidth]{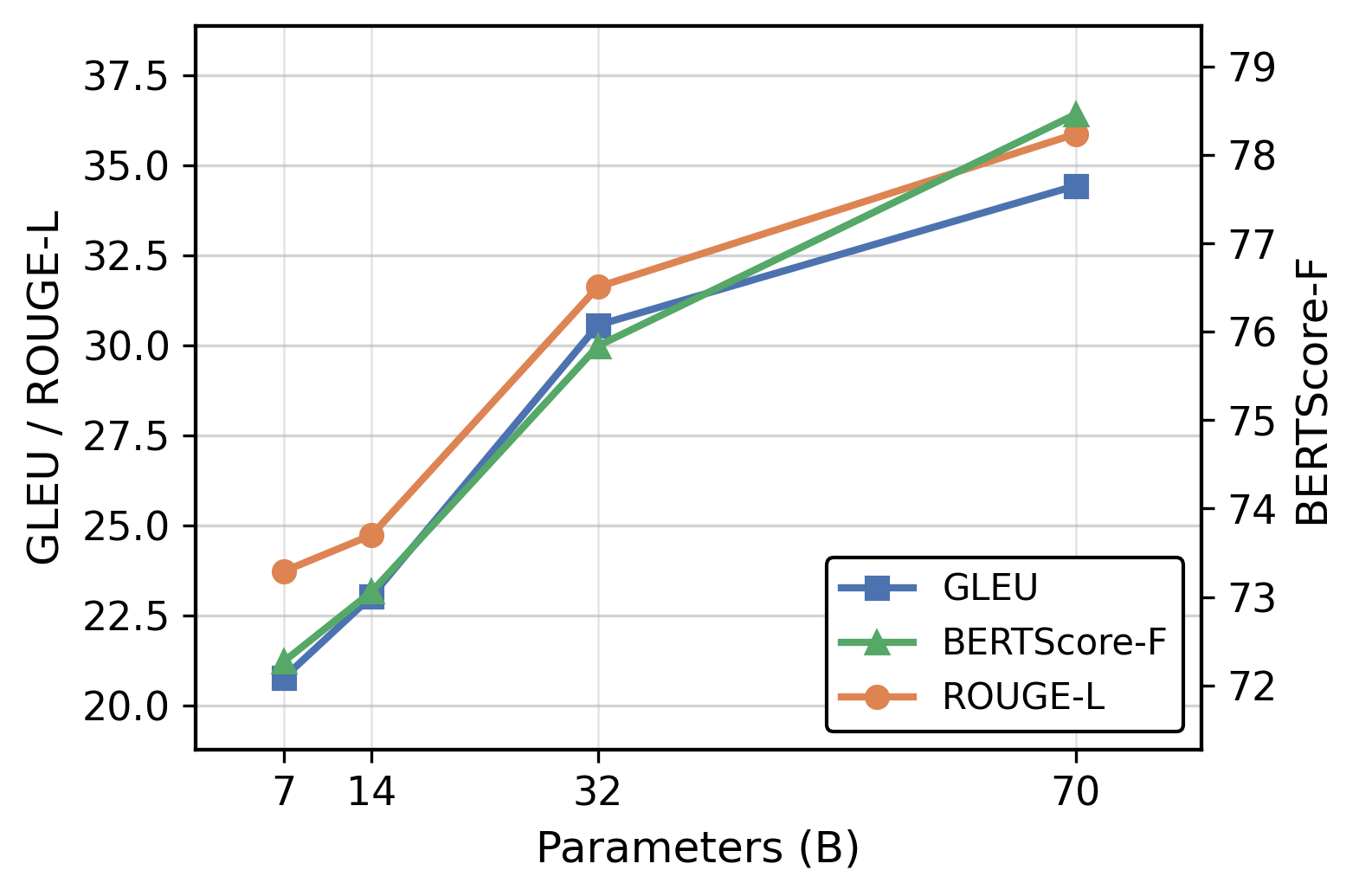}
  \caption{Comparative analysis of metrics scaling with increasing model size from DeepSeek-R1 7B to 70B.}
  \label{fig:70b}
\end{figure}

%In the qualitative assessment, we design 100 questions and ask both the fine-tuned Qwen2-7B applying Cochain and GPT-4 to answer them, with the additional requirement that GPT-4 not only solve these problems but also consider collaboration. We invite five domain experts from a large automotive manufacturing company, with each expert responsible for evaluating 20 questions. The evaluation results are shown in Figure~\ref{expert_evaluation}(a). For most of the questions, the performance of the small model combined with Cochain outperforms GPT-4. Additionally, we use GPT-4 to score the generated answers under four different conditions five times and take the average score as a supplement to the quantitative experiment. The results, shown in Figure~\ref{expert_evaluation}(b), are consistent with the quantitative evaluation.
% required packages:
% \usepackage{booktabs}
% \usepackage{multirow}
% \usepackage[table]{xcolor}
% \usepackage{subcaption}

\begin{figure}[t]
    \centering
    \includegraphics[width=\columnwidth]{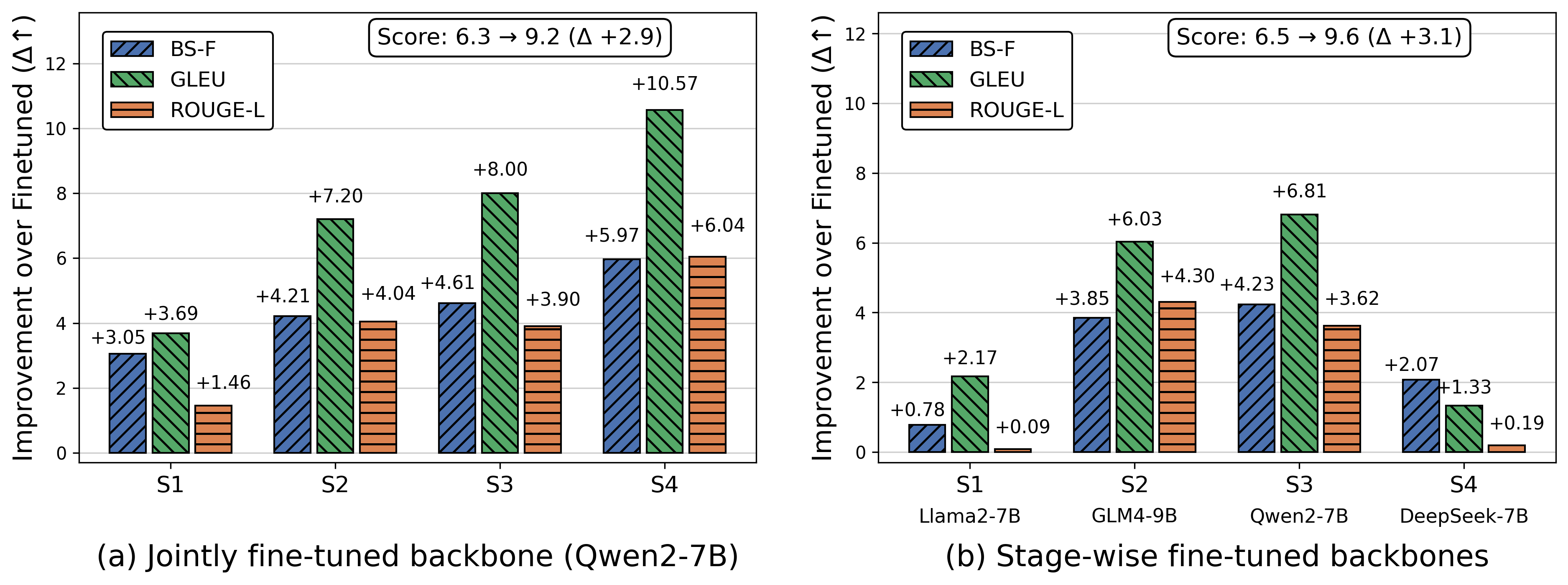}
\caption{Stage-wise gains of Cochain over fine-tuning on the Automotive workflow ($\Delta=\text{+Cochain}-\text{Finetuned}$). (a) Jointly fine-tuned backbone (Qwen2-7B). (b) Stage-wise fine-tuned backbones. Bars show improvements in BS-F, GLEU, and ROUGE-L. The box reports the overall GPT-4 score gain of the integrated solution.}
    \label{fig:stage_improvement}
\end{figure}

\begin{table*}[t]
    \centering
    \resizebox{\textwidth}{!}{
    \begin{tabular}{llccccccccc}
        \toprule
        \multirow{2}{*}{Backbone} & \multirow{2}{*}{Method} & \multicolumn{3}{c}{Automotive} & \multicolumn{3}{c}{Pharmaceutical} & \multicolumn{3}{c}{E-commerce} \\
        \cmidrule(lr){3-5} \cmidrule(lr){6-8} \cmidrule(lr){9-11} 
        & & BS-F & GLEU & ROUGE-L & BS-F & GLEU & ROUGE-L & BS-F & GLEU & ROUGE-L \\
        \midrule
        \multirow{4}{*}{Qwen2-7B} 
        & Cochain & 75.39 & 28.17 & 28.87 & 78.48 & 33.98 & 35.03 & 79.91 & 37.08 & 40.06 \\
        & w/o \(\mathcal{CKG}\) & 75.21 & 27.81 & 28.74 & 76.34 & \cellcolor{red!10}25.92 & \cellcolor{red!10}28.09 & \cellcolor{red!10}76.72 & \cellcolor{red!10}32.40 & \cellcolor{red!10}32.74 \\
        & w/o Causal Chain & \cellcolor{red!10}75.20 & \cellcolor{red!10}27.55 & \cellcolor{red!10}28.42 & \cellcolor{red!10}76.20 & \cellcolor{red!30}25.38 & \cellcolor{red!30}27.76 & 76.86 & 33.30 & 32.93 \\
        & w/o Prompts Tree & \cellcolor{red!30}73.08 & \cellcolor{red!30}22.45 & \cellcolor{red!30}24.56 & \cellcolor{red!30}76.19 & 26.84 & 28.91 & \cellcolor{red!30}76.59 & \cellcolor{red!30}31.83 & \cellcolor{red!30}32.38 \\
        \midrule
        \multirow{4}{*}{DeepSeek-R1-7B} 
        & Cochain & 73.19 & 21.73 & 24.01 & 78.97 & 37.94 & 36.75 & 82.68 & 50.30 & 47.24 \\
        & w/o \(\mathcal{CKG}\) & \cellcolor{red!10}72.41 & \cellcolor{red!10}19.71 & \cellcolor{red!10}22.63 & \cellcolor{red!10}78.22 & \cellcolor{red!30}37.33 & \cellcolor{red!10}36.40 & \cellcolor{red!30}77.36 & \cellcolor{red!10}36.54 & \cellcolor{red!10}35.85 \\
        & w/o Causal Chain & 72.67 & 19.96 & 22.86 & 78.41 & \cellcolor{red!10}37.36 & 36.74 & \cellcolor{red!10}77.60 & \cellcolor{red!30}36.30 & \cellcolor{red!30}35.74 \\
        & w/o Prompts Tree & \cellcolor{red!30}72.27 & \cellcolor{red!30}16.54 & \cellcolor{red!30}20.60 & \cellcolor{red!30}76.96 & 37.98 & \cellcolor{red!30}35.61 & 79.02 & 41.03 & 38.84 \\
        \midrule
        \multirow{4}{*}{Claude-3.5-haiku} 
        & Cochain & 69.61 & 18.73 & 17.30 & 72.87 & 23.59 & 19.67 & 71.36 & 20.15 & 20.78 \\
        & w/o \(\mathcal{CKG}\) & \cellcolor{red!10}68.98 & \cellcolor{red!10}16.63 & \cellcolor{red!10}16.31 & \cellcolor{red!10}70.93 & \cellcolor{red!10}16.42 & \cellcolor{red!10}18.45 & \cellcolor{red!10}71.13 & \cellcolor{red!10}18.85 & \cellcolor{red!10}20.22 \\
        & w/o Causal Chain & 69.33 & 17.31 & 16.65 & 71.33 & 17.43 & 18.78 & 71.23 & 18.89 & 20.45 \\
        & w/o Prompts Tree & \cellcolor{red!30}65.79 & \cellcolor{red!30}12.84 & \cellcolor{red!30}13.71 & \cellcolor{red!30}70.85 & \cellcolor{red!30}16.24 & \cellcolor{red!30}18.31 & \cellcolor{red!30}69.76 & \cellcolor{red!30}15.55 & \cellcolor{red!30}18.59 \\
        \bottomrule
    \end{tabular}}
    \caption{Result of ablation study. We highlight the \sethlcolor{red!30}\hl{largest} and \sethlcolor{red!10}\hl{second-largest} performance drops caused by removing each component.}
    \label{Ablation Study}
\end{table*}

\subsection{Key Design Factors}
\paragraph{Ablation Study.}
To validate Cochain's effectiveness, we conduct ablation experiments with three variants: (1) w/o Knowledge Graph; (2) w/o Causal Chain; and (3) w/o Prompts Tree. As shown in Table~\ref{Ablation Study}, all variants exhibit performance degradation across all metrics, confirming each component's importance. Notably, removing the prompts tree causes the most significant drop, highlighting its critical role in collaboration. The knowledge graph and causal chain components also prove essential, demonstrating their complementary functions in the reasoning process.

\paragraph{Hyperparameter Influence.}
We vary two hyperparameters: the prompt count and the retrieval nodes top-$n$ from the knowledge graph. Figure~\ref{fig:parameter_analysis} shows an inverted U-shape when fixing top-$n$ and increasing prompt count, indicating an optimal prompt budget rather than monotonic gains. When fixing the prompt budget and varying top-$n$, the trend is not monotonic. Overall, performance is more sensitive to prompt budgeting than simply retrieving more nodes, consistent with Table~\ref{Ablation Study} where Prompts Tree yields the largest contribution.

\paragraph{Stage Dependency.}
Table~\ref{skip} shows that skipping any stage consistently degrades all metrics, indicating that each stage contributes non-redundant constraints for end-to-end workflow reasoning.

%[height=3.5cm]
%[width=\linewidth]

\begin{figure}[t]
    \centering
    \includegraphics[width=\linewidth]{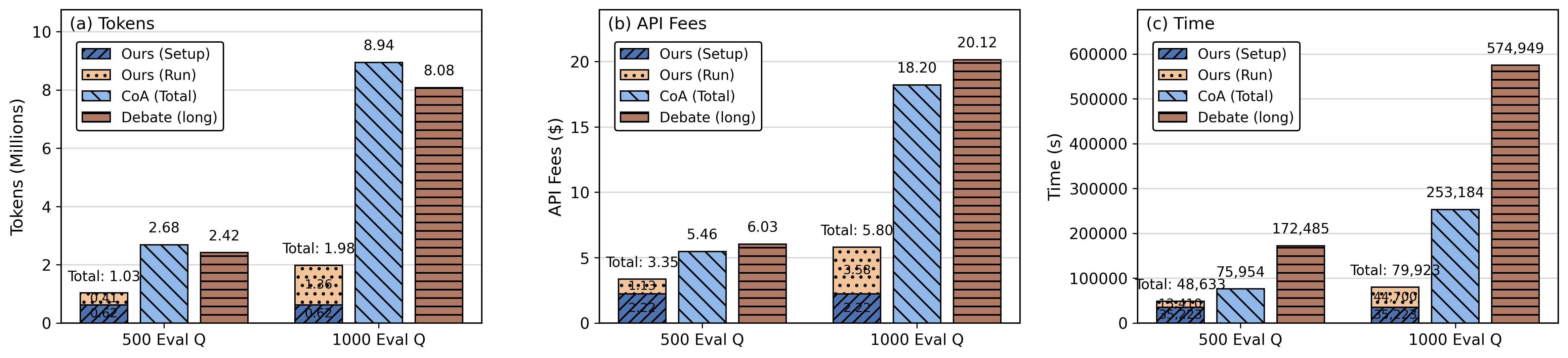}
    \caption{Cost comparison (Tokens, API Fees, Time) across baselines on
pharmaceutical dataset. Cochain's total costs include the one-time Prompts Tree setup and costs for varying evaluation queries.}
    \label{cost111}
\end{figure}

\subsection{Cross-Domain and Specialized Settings}

We evaluate Cochain on the Automotive workflow under two regimes: a jointly fine-tuned backbone trained on the union of all stage data, and stage-wise fine-tuned backbones specialized per stage. Figure~\ref{fig:stage_improvement} reports stage-wise gains over the corresponding fine-tuned baselines, where $\Delta=\text{+Cochain}-\text{Finetuned}$. With joint fine-tuning (Figure~\ref{fig:stage_improvement}(a)), Cochain improves all stages and metrics, with larger gains in later stages where downstream constraints are more tightly coupled with upstream decisions, indicating better constraint propagation without direct, token-intensive interaction. With stage-wise fine-tuning (Figure~\ref{fig:stage_improvement}(b)), Cochain consistently boosts each specialized backbone, showing it can coordinate domain-expert agents into a coherent workflow solution. The integrated four-stage outputs also receive higher GPT-4 scores in both regimes, supporting end-to-end quality improvements beyond stage-local changes.

\subsection{Efficiency and Cost Analysis}
%As shown in Figure~\ref{time}, the reasoning time differential between thinking and non-thinking models across increasing reasoning rounds reveals that with PMC, this differential grows dramatically as rounds increase, while with Cochain, it remains relatively moderate. This demonstrates Cochain's superior efficiency in handling complex reasoning tasks, with its performance advantage becoming increasingly pronounced as reasoning complexity increases. As shown in Figure~\ref{cost111}, although its construction introduces some initial overhead, we view this as a one-time investment that functions as a cost amortization strategy. We analyzed this by comparing Cochain with other baselines on the pharmaceutical dataset. Traditional multi-agent systems have no upfront setup cost but incur a very high cost for every inference. In contrast, Cochain's per-inference cost is substantially lower. As the number of uses increases, Cochain's initial investment is quickly amortized, leading to significant long-term advantages in total cost and time.

%Figure~\ref{time} shows that as reasoning rounds increase, the time gap between thinking and non-thinking models grows dramatically for PMC but remains moderate for Cochain, demonstrating Cochain's superior efficiency as complexity increases. 

Figure~\ref{cost111} breaks down efficiency into one-time setup overhead and per-query running cost. In contrast to interaction-heavy baselines, Cochain shifts most computation to an offline setup stage and keeps inference lightweight. This yields consistent savings across budgets: for 1000 evaluation queries, Cochain consumes only 1.98M tokens in total, compared with 8.94M for CoA and 8.08M for Debate(long). The same trend holds in monetary cost, where Cochain stays at \$5.80 versus \$18.20 and \$20.12, and in wall-clock time, where Cochain finishes in 79{,}923s versus 253{,}184s and 574{,}949s. While the setup stage introduces an upfront cost, it is paid once and then reused, so the marginal cost grows slowly with more queries and the total advantage widens at scale. This shows that Cochain delivers high-quality solutions at low cost, effectively alleviating excessive collaboration.

%\subsection{Overcollaboration is More Serious than Undercollaboration}

%Business workflow tasks are particularly susceptible to the ``overcollaboration'' phenomenon. When multiple agents collaborate, this manifests as responses deviating from core issues and reducing answer quality. Comparative analysis of Table~\ref{tab:combined-performance}, Table~\ref{tab2}, and additional single-agent experimental results (Appendix~\ref{app:third.1}) demonstrates that, despite consuming significantly more computational resources, overcollaboration performs substantially worse than undercollaboration. Overcollaboration's challenging nature is its covertness—systems maintain high activity and apparent collaboration, making the problem hard to detect and rectify promptly. While task decomposition, such as approaches like PMC, and summarization, such as strategies like CoA, are considered effective methods for focusing on core issues~\cite{wu2024autogen,chen2024reconcileroundtableconferenceimproves}, Cochain's collaboration mechanism exhibits superior performance in controlling excessive collaboration, enabling a more precise focus on critical task requirements.

\section{Conclusion}

In this paper, we propose Cochain, a collaboration prompting framework that balances Insufficient Collaboration and Excessive Collaboration in LLM agent workflows. Cochain bridges single-agent depth and multi-agent breadth via reusable collaboration artifacts, achieving low operational cost, fast inference, strong interpretability, and robust end-to-end reasoning. It builds a Prompts Tree to distill and propagate stage-wise constraints, enabling coherent multi-stage solutions without token-heavy interactions. Cochain also constructs a cross-stage Collaborative Knowledge Graph with tacit knowledge elicited through counterfactual querying, which strengthens relational reasoning and reduces hallucinations. In addition, the causal chain mechanism exposes explicit dependency traces, improving transparency and controllability. Extensive experiments on specialized and cross-domain settings show consistent improvements in scenarios affected by both collaboration extremes, and expert evaluation further indicates that smaller models equipped with Cochain can outperform GPT-4 on workflow tasks.

\bibliographystyle{named}
\bibliography{ijcai26}

@article{ouyang2022training,
  title={Training language models to follow instructions with human feedback},
  author={Ouyang, Long and Wu, Jeffrey and Jiang, Xu and Almeida, Diogo and Wainwright, Carroll and Mishkin, Pamela and Zhang, Chong and Agarwal, Sandhini and Slama, Katarina and Ray, Alex and others},
  journal={Advances in neural information processing systems},
  volume={35},
  pages={27730--27744},
  year={2022}
}

@inproceedings{wen2023mindmap,
  title={MindMap: Knowledge Graph Prompting Sparks Graph of Thoughts in Large Language Models},
  author={Wen, Yilin and Wang, Zifeng and Sun, Jimeng},
  booktitle={Proceedings of the 62nd Annual Meeting of the Association for Computational Linguistics},
  year={2024}
}

@article{cui2024chatlaw,
  title={Chatlaw: A multi-agent collaborative legal assistant with knowledge graph enhanced mixture-of-experts large language model},
  author={Cui, Jiaxi and Ning, Munan and Li, Zongjian and Chen, Bohua and Yan, Yang and Li, Hao and Ling, Bin and Tian, Yonghong and Yuan, Li},
  journal={arXiv preprint arXiv:2306.16092},
  year={2024}
}

@article{zhou2024multifaceteval,
  title={MultifacetEval: Multifaceted Evaluation to Probe LLMs in Mastering Medical Knowledge},
  author={Zhou, Yuxuan and Liu, Xien and Ning, Chen and Wu, Ji},
  journal={arXiv preprint arXiv:2406.02919},
  year={2024}
}

@misc{tang2023medagents,
      title={MedAgents: Large Language Models as Collaborators for Zero-shot Medical Reasoning}, 
      author={Xiangru Tang and Anni Zou and Zhuosheng Zhang and Ziming Li and Yilun Zhao and Xingyao Zhang and Arman Cohan and Mark Gerstein},
      year={2024},
      eprint={2311.10537},
      archivePrefix={arXiv},
      primaryClass={cs.CL},
      url={https://arxiv.org/abs/2311.10537}, 
}

@article{dan2023educhat,
  title={Educhat: A large-scale language model-based chatbot system for intelligent education},
  author={Dan, Yuhao and Lei, Zhikai and Gu, Yiyang and Li, Yong and Yin, Jianghao and Lin, Jiaju and Ye, Linhao and Tie, Zhiyan and Zhou, Yougen and Wang, Yilei and others},
  journal={arXiv preprint arXiv:2308.02773},
  year={2023}
}

@article{yang2023fingpt,
  title={FinGPT: Open-Source Financial Large Language Models},
  author={Yang, Hongyang and Liu, Xiao-Yang and Wang, Christina Dan},
  journal={FinLLM Symposium at IJCAI 2023},
  year={2023}
}

@inproceedings{jain2023mechanistically,
    title={Mechanistically analyzing the effects of fine-tuning on procedurally defined tasks},
    author={Samyak Jain and Robert Kirk and Ekdeep Singh Lubana and Robert P. Dick and Hidenori Tanaka and Tim Rockt{\"a}schel and Edward Grefenstette and David Krueger},
    booktitle={The Twelfth International Conference on Learning Representations},
    year={2024},
    url={https://openreview.net/forum?id=A0HKeKl4Nl}
}

@inproceedings{wang2024knowledge,
  title={Knowledge graph prompting for multi-document question answering},
  author={Wang, Yu and Lipka, Nedim and Rossi, Ryan A and Siu, Alexa and Zhang, Ruiyi and Derr, Tyler},
  booktitle={Proceedings of the AAAI Conference on Artificial Intelligence},
  volume={38},
  number={17},
  pages={19206--19214},
  year={2024}
}

@article{brown2020language,
  title={Language models are few-shot learners Advances in neural information processing systems 33},
  author={Brown, T and Mann, B and Ryder, N and Subbiah, M and Kaplan, JD and Dhariwal, P and Neelakantan, A and Shyam, P and Sastry, G and Askell, A and others},
  year={2020}
}

@article{wei2022chain,
  title={Chain-of-thought prompting elicits reasoning in large language models},
  author={Wei, Jason and Wang, Xuezhi and Schuurmans, Dale and Bosma, Maarten and Xia, Fei and Chi, Ed and Le, Quoc V and Zhou, Denny and others},
  journal={Advances in neural information processing systems},
  volume={35},
  pages={24824--24837},
  year={2022}
}

@inproceedings{wang2022self,
title={Self-Consistency Improves Chain of Thought Reasoning in Language Models},
author={Xuezhi Wang and Jason Wei and Dale Schuurmans and Quoc V Le and Ed H. Chi and Sharan Narang and Aakanksha Chowdhery and Denny Zhou},
booktitle={The Eleventh International Conference on Learning Representations },
year={2023},
url={https://openreview.net/forum?id=1PL1NIMMrw}
}

@inproceedings{
yao2022react,
title={ReAct: Synergizing Reasoning and Acting in Language Models},
author={Shunyu Yao and Jeffrey Zhao and Dian Yu and Nan Du and Izhak Shafran and Karthik R Narasimhan and Yuan Cao},
booktitle={The Eleventh International Conference on Learning Representations },
year={2023},
url={https://openreview.net/forum?id=WE_vluYUL-X}
}

@article{yao2024tree,
  title={Tree of thoughts: Deliberate problem solving with large language models},
  author={Yao, Shunyu and Yu, Dian and Zhao, Jeffrey and Shafran, Izhak and Griffiths, Tom and Cao, Yuan and Narasimhan, Karthik},
  journal={Advances in Neural Information Processing Systems},
  volume={36},
  year={2024}
}

@inproceedings{zhang2024chain,
title={Chain of Preference Optimization: Improving Chain-of-Thought Reasoning in {LLM}s},
author={Xuan Zhang and Chao Du and Tianyu Pang and Qian Liu and Wei Gao and Min Lin},
booktitle={The Thirty-eighth Annual Conference on Neural Information Processing Systems},
year={2024},
url={https://openreview.net/forum?id=2cczgOfMP4}
}

@article{achiam2023gpt,
  title={Gpt-4 technical report},
  author={Achiam, Josh and Adler, Steven and Agarwal, Sandhini and Ahmad, Lama and Akkaya, Ilge and Aleman, Florencia Leoni and Almeida, Diogo and Altenschmidt, Janko and Altman, Sam and Anadkat, Shyamal and others},
  journal={arXiv preprint arXiv:2303.08774},
  year={2023}
}

@article{zhang2019bertscore,
  title={Bertscore: Evaluating text generation with bert},
  author={Zhang, Tianyi and Kishore, Varsha and Wu, Felix and Weinberger, Kilian Q and Artzi, Yoav},
  journal={arXiv preprint arXiv:1904.09675},
  year={2019}
}

@inproceedings{papineni2002bleu,
  title={Bleu: a method for automatic evaluation of machine translation},
  author={Papineni, Kishore and Roukos, Salim and Ward, Todd and Zhu, Wei-Jing},
  booktitle={Proceedings of the 40th annual meeting of the Association for Computational Linguistics},
  pages={311--318},
  year={2002}
}

@inproceedings{banerjee2005meteor,
  title={METEOR: An automatic metric for MT evaluation with improved correlation with human judgments},
  author={Banerjee, Satanjeev and Lavie, Alon},
  booktitle={Proceedings of the acl workshop on intrinsic and extrinsic evaluation measures for machine translation and/or summarization},
  pages={65--72},
  year={2005}
}

@inproceedings{lin2004rouge,
  title={Rouge: A package for automatic evaluation of summaries},
  author={Lin, Chin-Yew},
  booktitle={Text summarization branches out},
  pages={74--81},
  year={2004}
}

@article{touvron2023llama,
  title={Llama: Open and efficient foundation language models},
  author={Touvron, Hugo and Lavril, Thibaut and Izacard, Gautier and Martinet, Xavier and Lachaux, Marie-Anne and Lacroix, Timoth{\'e}e and Rozi{\`e}re, Baptiste and Goyal, Naman and Hambro, Eric and Azhar, Faisal and others},
  journal={arXiv preprint arXiv:2302.13971},
  year={2023}
}

@article{bai2023qwen,
  title={Qwen technical report},
  author={Bai, Jinze and Bai, Shuai and Chu, Yunfei and Cui, Zeyu and Dang, Kai and Deng, Xiaodong and Fan, Yang and Ge, Wenbin and Han, Yu and Huang, Fei and others},
  journal={arXiv preprint arXiv:2309.16609},
  year={2023}
}

@inproceedings{wei2024editable,
  title={Editable scene simulation for autonomous driving via collaborative llm-agents},
  author={Wei, Yuxi and Wang, Zi and Lu, Yifan and Xu, Chenxin and Liu, Changxing and Zhao, Hao and Chen, Siheng and Wang, Yanfeng},
  booktitle={Proceedings of the IEEE/CVF Conference on Computer Vision and Pattern Recognition},
  pages={15077--15087},
  year={2024}
}

@article{glm2024chatglm,
  title={Chatglm: A family of large language models from glm-130b to glm-4 all tools},
  author={GLM, Team and Zeng, Aohan and Xu, Bin and Wang, Bowen and Zhang, Chenhui and Yin, Da and Zhang, Dan and Rojas, Diego and Feng, Guanyu and Zhao, Hanlin and others},
  journal={arXiv preprint arXiv:2406.12793},
  year={2024}
}

@misc{du2023improving,
title={Improving Factuality and Reasoning in Language Models through Multiagent Debate},
author={Yilun Du and Shuang Li and Antonio Torralba and Joshua B. Tenenbaum and Igor Mordatch},
year={2024},
url={https://openreview.net/forum?id=QAwaaLJNCk}
}

@article{bi2024deepseek,
  title={Deepseek llm: Scaling open-source language models with longtermism},
  author={Bi, Xiao and Chen, Deli and Chen, Guanting and Chen, Shanhuang and Dai, Damai and Deng, Chengqi and Ding, Honghui and Dong, Kai and Du, Qiushi and Fu, Zhe and others},
  journal={arXiv preprint arXiv:2401.02954},
  year={2024}
}

@article{zhang2024coa,
  title={Chain of agents: Large language models collaborating on long-context tasks},
  author={Zhang, Yusen and Sun, Ruoxi and Chen, Yanfei and Pfister, Tomas and Zhang, Rui and Arik, Sercan},
  journal={Advances in Neural Information Processing Systems},
  volume={37},
  pages={132208--132237},
  year={2024}
}

@inproceedings{zhang2025planning,
  title={Planning with Multi-Constraints via Collaborative Language Agents},
  author={Zhang, Cong and Goh, Xin Deik and Li, Dexun and Zhang, Hao and Liu, Yong},
  booktitle={Proceedings of the 31st International Conference on Computational Linguistics},
  pages={10054--10082},
  year={2025}
}

@misc{deepseekai2025deepseekr1incentivizingreasoningcapability,
      title={DeepSeek-R1: Incentivizing Reasoning Capability in LLMs via Reinforcement Learning}, 
      author={DeepSeek-AI},
      year={2025},
      eprint={2501.12948},
      archivePrefix={arXiv},
      primaryClass={cs.CL},
      url={https://arxiv.org/abs/2501.12948}, 
}

@article{qwen2,
  title={Qwen2 Technical Report},
  year={2024}
}

@article{touvron2023llama2openfoundation,
  title={Llama 2: Open foundation and fine-tuned chat models},
  author={Touvron, Hugo and Martin, Louis and Stone, Kevin and Albert, Peter and Almahairi, Amjad and Babaei, Yasmine and Bashlykov, Nikolay and Batra, Soumya and Bhargava, Prajjwal and Bhosale, Shruti and others},
  journal={arXiv preprint arXiv:2307.09288},
  year={2023}
}

@article{zhao2024longagent,
  title={Longagent: scaling language models to 128k context through multi-agent collaboration},
  author={Zhao, Jun and Zu, Can and Xu, Hao and Lu, Yi and He, Wei and Ding, Yiwen and Gui, Tao and Zhang, Qi and Huang, Xuanjing},
  journal={arXiv preprint arXiv:2402.11550},
  year={2024}
}

@misc{chen2024reconcileroundtableconferenceimproves,
      title={ReConcile: Round-Table Conference Improves Reasoning via Consensus among Diverse LLMs}, 
      author={Justin Chih-Yao Chen and Swarnadeep Saha and Mohit Bansal},
      year={2024},
      eprint={2309.13007},
      archivePrefix={arXiv},
      primaryClass={cs.CL},
      url={https://arxiv.org/abs/2309.13007}, 
}

@article{huang2023agentcoder,
  title={Agentcoder: Multi-agent-based code generation with iterative testing and optimisation},
  author={Huang, Dong and Zhang, Jie M and Luck, Michael and Bu, Qingwen and Qing, Yuhao and Cui, Heming},
  journal={arXiv preprint arXiv:2312.13010},
  year={2023}
}

@misc{
zhou2025reflection,
title={Reflection on Knowledge Graph for Large Language Models Reasoning},
author={Yigeng Zhou and Yifan Lu and Jing Li and Fangming Liu and Meishan Zhang and Yequan Wang and Daojing He and Min Zhang},
year={2025},
url={https://openreview.net/forum?id=6f7RoeQ7Go}
}

@article{lavrinovics2025knowledge,
  title={Knowledge Graphs, Large Language Models, and Hallucinations: An NLP Perspective},
  author={Lavrinovics, Ernests and Biswas, Russa and Bjerva, Johannes and Hose, Katja},
  journal={Journal of Web Semantics},
  volume={85},
  pages={100844},
  year={2025},
  publisher={Elsevier}
}

@article{sahoo2024systematic,
  title={A systematic survey of prompt engineering in large language models: Techniques and applications},
  author={Sahoo, Pranab and Singh, Ayush Kumar and Saha, Sriparna and Jain, Vinija and Mondal, Samrat and Chadha, Aman},
  journal={arXiv preprint arXiv:2402.07927},
  year={2024}
}

@article{li2023chain,
  title={Chain of code: Reasoning with a language model-augmented code emulator},
  author={Li, Chengshu and Liang, Jacky and Zeng, Andy and Chen, Xinyun and Hausman, Karol and Sadigh, Dorsa and Levine, Sergey and Fei-Fei, Li and Xia, Fei and Ichter, Brian},
  journal={arXiv preprint arXiv:2312.04474},
  year={2023}
}

@inproceedings{besta2024graph,
  title={Graph of thoughts: Solving elaborate problems with large language models},
  author={Besta, Maciej and Blach, Nils and Kubicek, Ales and Gerstenberger, Robert and Podstawski, Michal and Gianinazzi, Lukas and Gajda, Joanna and Lehmann, Tomasz and Niewiadomski, Hubert and Nyczyk, Piotr and others},
  booktitle={Proceedings of the AAAI Conference on Artificial Intelligence},
  volume={38},
  number={16},
  pages={17682--17690},
  year={2024}
}

@article{li2025parallelized,
  title={Parallelized Planning-Acting for Efficient LLM-based Multi-Agent Systems},
  author={Li, Yaoru and Liu, Shunyu and Zheng, Tongya and Song, Mingli},
  journal={arXiv preprint arXiv:2503.03505},
  year={2025}
}

@article{zhang2025collm,
  title={Collm: Integrating collaborative embeddings into large language models for recommendation},
  author={Zhang, Yang and Feng, Fuli and Zhang, Jizhi and Bao, Keqin and Wang, Qifan and He, Xiangnan},
  journal={IEEE Transactions on Knowledge and Data Engineering},
  year={2025},
  publisher={IEEE}
}

@article{piatti2024cooperate,
  title={Cooperate or collapse: Emergence of sustainable cooperation in a society of llm agents},
  author={Piatti, Giorgio and Jin, Zhijing and Kleiman-Weiner, Max and Sch{\"o}lkopf, Bernhard and Sachan, Mrinmaya and Mihalcea, Rada},
  journal={Advances in Neural Information Processing Systems},
  volume={37},
  pages={111715--111759},
  year={2024}
}

@article{courellis2024abstract,
  title={Abstract representations emerge in human hippocampal neurons during inference},
  author={Courellis, Hristos S and Minxha, Juri and Cardenas, Araceli R and Kimmel, Daniel L and Reed, Chrystal M and Valiante, Taufik A and Salzman, C Daniel and Mamelak, Adam N and Fusi, Stefano and Rutishauser, Ueli},
  journal={Nature},
  volume={632},
  number={8026},
  pages={841--849},
  year={2024},
  publisher={Nature Publishing Group UK London}
}

@inproceedings{reimers-2019-sentence-bert,
    title = "Sentence-BERT: Sentence Embeddings using Siamese BERT-Networks",
    author = "Reimers, Nils and Gurevych, Iryna",
    booktitle = "Proceedings of the 2019 Conference on Empirical Methods in Natural Language Processing",
    month = "11",
    year = "2019",
    publisher = "Association for Computational Linguistics",
    url = "http://arxiv.org/abs/1908.10084",
}

@article{hu2022lora,
  title={Lora: Low-rank adaptation of large language models.},
  author={Hu, Edward J and Shen, Yelong and Wallis, Phillip and Allen-Zhu, Zeyuan and Li, Yuanzhi and Wang, Shean and Wang, Lu and Chen, Weizhu and others},
  journal={ICLR},
  volume={1},
  number={2},
  pages={3},
  year={2022}
}

@article{dettmers2023qlora,
  title={Qlora: Efficient finetuning of quantized llms},
  author={Dettmers, Tim and Pagnoni, Artidoro and Holtzman, Ari and Zettlemoyer, Luke},
  journal={Advances in neural information processing systems},
  volume={36},
  pages={10088--10115},
  year={2023}
}

@inproceedings{
wu2024autogen,
title={AutoGen: Enabling Next-Gen {LLM} Applications via Multi-Agent Conversations},
author={Qingyun Wu and Gagan Bansal and Jieyu Zhang and Yiran Wu and Beibin Li and Erkang Zhu and Li Jiang and Xiaoyun Zhang and Shaokun Zhang and Jiale Liu and Ahmed Hassan Awadallah and Ryen W White and Doug Burger and Chi Wang},
booktitle={First Conference on Language Modeling},
year={2024},
url={https://openreview.net/forum?id=BAakY1hNKS}
}

@article{wu2016google,
  title={Google's neural machine translation system: Bridging the gap between human and machine translation},
  author={Wu, Yonghui and Schuster, Mike and Chen, Zhifeng and Le, Quoc V and Norouzi, Mohammad and Macherey, Wolfgang and Krikun, Maxim and Cao, Yuan and Gao, Qin and Macherey, Klaus and others},
  journal={arXiv preprint arXiv:1609.08144},
  year={2016}
}

@misc{shen2025autoslurpbenchmarkdatasetevaluating,
      title={Auto-SLURP: A Benchmark Dataset for Evaluating Multi-Agent Frameworks in Smart Personal Assistant}, 
      author={Lei Shen and Xiaoyu Shen},
      year={2025},
      eprint={2504.18373},
      archivePrefix={arXiv},
      primaryClass={cs.CL},
      url={https://arxiv.org/abs/2504.18373}, 
}

@misc{lei2025mscorebenchmarkmultistagecollaborative,
      title={MSCoRe: A Benchmark for Multi-Stage Collaborative Reasoning in LLM Agents}, 
      author={Yuzhen Lei and Hongbin Xie and Jiaxing Zhao and Shuangxue Liu and Xuan Song},
      year={2025},
      eprint={2509.17628},
      archivePrefix={arXiv},
      primaryClass={cs.CL},
      url={https://arxiv.org/abs/2509.17628}, 
}

@misc{deepseekai2024deepseekv3technicalreport,
      title={DeepSeek-V3 Technical Report}, 
      author={DeepSeek-AI},
      year={2024},
      eprint={2412.19437},
      archivePrefix={arXiv},
      primaryClass={cs.CL},
      url={https://arxiv.org/abs/2412.19437}, 
}

@inproceedings{10.1145/2959100.2959190,
author = {Covington, Paul and Adams, Jay and Sargin, Emre},
title = {Deep Neural Networks for YouTube Recommendations},
year = {2016},
isbn = {9781450340359},
publisher = {Association for Computing Machinery},
address = {New York, NY, USA},
url = {https://doi.org/10.1145/2959100.2959190},
doi = {10.1145/2959100.2959190},
booktitle = {Proceedings of the 10th ACM Conference on Recommender Systems},
pages = {191–198},
numpages = {8},
location = {Boston, Massachusetts, USA},
series = {RecSys '16}
}

@misc{chen2025panguembeddedefficientdualsystem,
      title={Pangu Embedded: An Efficient Dual-system LLM Reasoner with Metacognition}, 
      author={Hanting Chen and Yasheng Wang and Kai Han and Dong Li and Lin Li and Zhenni Bi and Jinpeng Li and Haoyu Wang and Fei Mi and Mingjian Zhu and Bin Wang and Kaikai Song and Yifei Fu and Xu He and Yu Luo and Chong Zhu and Quan He and Xueyu Wu and Wei He and Hailin Hu and Yehui Tang and Dacheng Tao and Xinghao Chen and Yunhe Wang},
      year={2025},
      eprint={2505.22375},
      archivePrefix={arXiv},
      primaryClass={cs.CL},
      url={https://arxiv.org/abs/2505.22375}, 
}

\clearpage

\appendix
\onecolumn

\section*{Summary of the Supplementary Material}

This appendix contains additional details for the paper. The appendix is organized as follows:

\begin{itemize}
    \item \S~\ref{algor} describes algorithms and examples to better understand Cochain.
    \item \S~\ref{app:first} describes the probabilistic formulations and mathematical proof for modeling an agent's unobservable tacit knowledge through the generation of counterfactual outputs.
    \item \S~\ref{app:second} introduces more details on Cochain.
    \item \S~\ref{app:third} presents more experiments and analysis on Cochain.
    \item \S~\ref{app44444} provides a qualitative analysis to interpret the outputs of Cochain, alongside a comparative case study against baseline methods.
    \item \S~\ref{app:template} presents the prompt templates for Cochain and other baselines.
\end{itemize}

\section{Algorithms and Examples to Better Understand Cochain}
\label{algor}

% In your preamble, you'll need these packages:
% \usepackage{algorithm}
% \usepackage{algpseudocode}
% \usepackage{amsmath}
\begin{algorithm}[h]
\caption{Collaborative Knowledge Graph Construction}
\label{alg:kg_construction}
\begin{algorithmic}[1] % The [1] enables line numbers

\Require 
    Datasets $D = \{D_1, \dots, D_N\}$; 
    LLM agents $A = \{A_1, \dots, A_N\}$; 
    Evaluation LLM $E$; 
    Stage Labels $L = \{L_1, \dots, L_N\}$.
\Ensure A collaborative knowledge graph $\mathrm{KG}_{\text{collab}}$.

\State \textbf{Initialize:} $\mathrm{KG}_{\text{explicit}} \gets \emptyset$, $\mathrm{KG}_{\text{tacit}} \gets \emptyset$

\Statex
\State \Comment{\textbf{Phase 1: Build Explicit Knowledge Graph}}
\For{$i \gets 1$ \textbf{to} $N$}
    \ForAll{$(\text{instruction}, \text{response}) \in D_i$}
        \State $\mathrm{triples} \gets \text{ExtractTriples}(\text{instruction}, \text{response})$
        \State $\text{LabelNodesInTriples}(\mathrm{triples}, L_i)$
        \State $\mathrm{KG}_{\text{explicit}} \gets \mathrm{KG}_{\text{explicit}} \cup \mathrm{triples}$
    \EndFor
\EndFor

\Statex
\State \Comment{\textbf{Phase 2: Build Tacit Knowledge Graph}}
\For{$i \gets 1$ \textbf{to} $N$}
    \ForAll{$(\text{instruction}, \text{response}) \in D_i$}
        \State $\mathrm{cf\_instruction} \gets \text{GenerateCounterfactual}(\text{instruction})$
        \State \textbf{repeat}
            \State $\mathrm{cf\_response} \gets A_i.\text{query}(\mathrm{cf\_instruction})$
            \State $(\text{evaluation}, \text{feedback}) \gets E.\text{evaluate}(\mathrm{cf\_response})$
            \If{$\text{evaluation} \neq \text{"reasonable"}$}
                \State $\mathrm{cf\_instruction}.\text{append}(\text{feedback})$
            \EndIf
        \State \textbf{until} {$\text{evaluation} = \text{"reasonable"}$}
        \State $\mathrm{tacit\_triples} \gets \text{ExtractTriples}(\mathrm{cf\_instruction}, \mathrm{cf\_response})$
        \State $\text{LabelNodesInTriples}(\mathrm{tacit\_triples}, L_i)$
        \State $\mathrm{KG}_{\text{tacit}} \gets \mathrm{KG}_{\text{tacit}} \cup \mathrm{tacit\_triples}$
    \EndFor
\EndFor

\Statex
\State \Comment{\textbf{Phase 3: Merge Graphs}}
\State $\mathrm{KG}_{\text{collab}} \gets \text{Merge}(\mathrm{KG}_{\text{explicit}}, \mathrm{KG}_{\text{tacit}})$
\State \Return $\mathrm{KG}_{\text{collab}}$
\end{algorithmic}
\end{algorithm}

\begin{algorithm}[h!]
\caption{Causal Chain Retrieval and Construction}
\label{alg:causal_chain}
\begin{algorithmic}[1] % The [1] enables line numbers

\Require 
    User query $\mathrm{query}$; 
    Stage-labeled knowledge graph $\mathrm{KG}_{\text{collab}}$; 
    Text encoder $\mathrm{Encoder}$; 
    Seed nodes count $\mathrm{top\_n}$; 
    Chain length limit $\mathrm{max\_depth}$.
\Ensure A textual causal chain $\mathrm{causal\_chain\_text}$.

\Statex
\State \Comment{\textbf{Phase 1: Two-Stage Seed Node Retrieval}}
\State $\mathrm{keywords} \gets \text{ExtractKeywords}(\mathrm{query})$
\State $\mathrm{query\_vector} \gets \mathrm{Encoder}.\text{encode}(\mathrm{query})$
\State $\mathrm{candidate\_nodes} \gets \mathrm{KG}_{\text{collab}}.\text{KeywordSearch}(\mathrm{keywords})$
\State $\mathrm{candidate\_vectors} \gets \mathrm{Encoder}.\text{encode}(\mathrm{candidate\_nodes})$
\State $\mathrm{similarities} \gets \text{CosineSimilarity}(\mathrm{query\_vector}, \mathrm{candidate\_vectors})$
\State $\mathrm{seed\_nodes} \gets \text{GetTopN}(\mathrm{candidate\_nodes}, \mathrm{similarities}, \mathrm{top\_n})$

\Statex
\State \Comment{\textbf{Phase 2: Stage-Aware Multi-hop Expansion}}
\State $\mathrm{all\_paths} \gets \emptyset$
\For{each $\mathrm{start\_node}$ in $\mathrm{seed\_nodes}$}
    \State \Comment{Search paths crossing stages via bridge nodes with multiple labels}
    \State $\mathrm{paths} \gets \mathrm{KG}_{\text{collab}}.\text{FindCrossStagePaths}(\mathrm{start\_node}, \mathrm{min\_depth} \gets 2, \mathrm{max\_depth})$
    \State $\mathrm{all\_paths} \gets \mathrm{all\_paths} \cup \mathrm{paths}$
\EndFor

\Statex
\State \Comment{\textbf{Phase 3: Format and Output}}
\State $\mathrm{causal\_chain\_text} \gets \text{VerbalizePaths}(\mathrm{all\_paths})$
\State \Return $\mathrm{causal\_chain\_text}$
\end{algorithmic}
\end{algorithm}

\begin{algorithm}[h]
\caption{Prompts Tree Construction}
\label{alg:prompts_tree}
\begin{algorithmic}[1] % The [1] enables line numbers

\Require 
    Workflow stages $S = \{S_1, \dots, S_n\}$; 
    Domain agents $A = \{A_1, \dots, A_n\}$; 
    Seed Q-A pair $\mathrm{InitialSeedQA}$; 
    Best prompts count $m$.
\Ensure A prompt tree $\mathrm{PromptsTree}$.

\State \textbf{Initialize:}
\State $\mathrm{rootNode} \gets \text{CreateNode}(\mathrm{InitialSeedQA}.\text{question})$
\State $\mathrm{PromptsTree} \gets \text{InitializeTree}(\mathrm{rootNode})$
\State $\mathrm{queue} \gets [(\mathrm{rootNode}, \mathrm{InitialSeedQA}.\text{answer}, S_1)]$

\While{$\mathrm{queue} \neq \emptyset$}
    \State $\mathrm{parentNode}, \mathrm{currentAnswer}, \mathrm{currentStage} \gets \mathrm{queue}.\text{pop\_front}()$
    
    \State \Comment{Phase 1: Distill Prompts}
    \State $\mathrm{distilledPrompts} \gets A_{\mathrm{currentStage}}.\text{DistillPrompts}(\mathrm{currentAnswer})$
    
    \State \Comment{Phase 2: Select Best Prompts}
    \State $\mathrm{bestPrompts} \gets A_{\mathrm{currentStage}}.\text{SelfEvaluate}(\mathrm{distilledPrompts}, m)$
    
    \State \Comment{Phase 3: Grow Tree}
    \If{$\mathrm{currentStage} \neq S_n$} \Comment{Proceed if not the last stage}
        \State $\mathrm{nextStage} \gets S_{\text{index}(\mathrm{currentStage}) + 1}$
        \For{each $\mathrm{promptText}$ in $\mathrm{bestPrompts}$}
            \State $\mathrm{childNode} \gets \text{CreateNode}(\mathrm{promptText})$
            \State $\mathrm{parentNode}.\text{add\_child}(\mathrm{childNode})$
            \State $\mathrm{newQuestion} \gets \text{GenerateQuestionForNextStage}(\mathrm{promptText}, \mathrm{nextStage})$
            \State $\mathrm{newAnswer} \gets A_{\mathrm{nextStage}}.\text{query}(\mathrm{newQuestion})$
            \State $\mathrm{queue}.\text{push\_back}((\mathrm{childNode}, \mathrm{newAnswer}, \mathrm{nextStage}))$
        \EndFor
    \EndIf
\EndWhile

\State \Return $\mathrm{PromptsTree}$
\end{algorithmic}
\end{algorithm}

\clearpage
\newpage

As shown in the Figure~\ref{example111111}, after applying Cochain, the agent anticipates that carbon fiber requires complex molding processes, completely different from traditional metal stamping, and that qualified suppliers are scarce, thus requiring consideration of supply chain stability and high procurement costs significantly impacting vehicle pricing and profit margins. The prompts tree anticipates existing production processes, guiding the agent to consider compatibility constraints between new equipment and existing stamping-welding lines. Based on this, Cochain generates superior answers compared to single agents.

\begin{figure}[h!] 
    \centering
    \includegraphics[width=\textwidth]{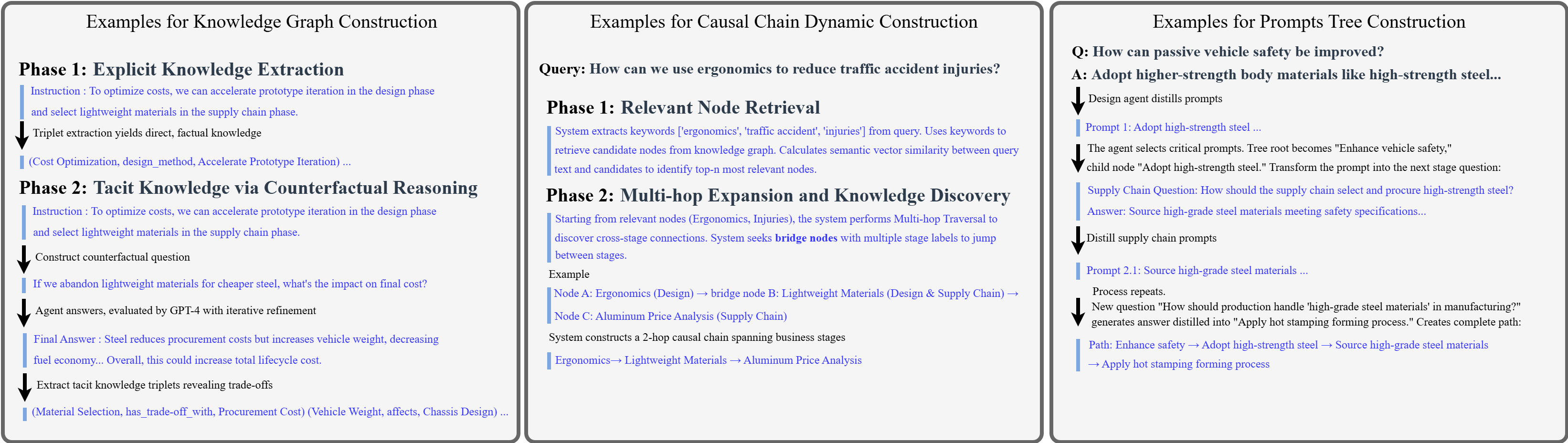}
    \caption{Examples of implementing a core component in the Cochain framework.} 
\end{figure}

\begin{figure}[h!] 
    \centering
    \includegraphics[scale=0.95]{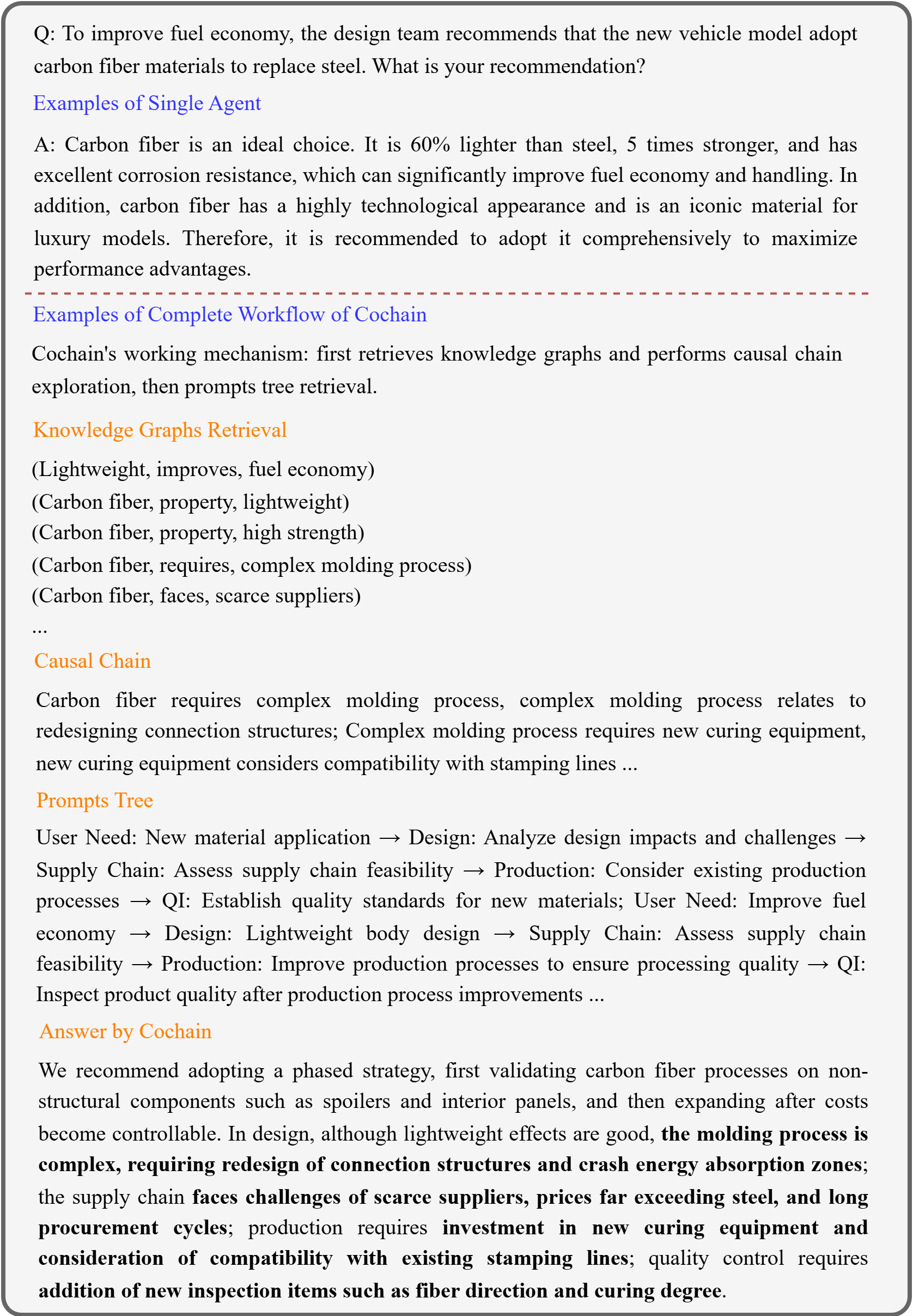}
    \caption{Examples of single agent and complete workflow of Cochain}
    \label{example111111}
\end{figure}

%\section*{Limitations and Future Work}

%While our Cochain framework excels in the evaluated business workflow tasks, its out-of-the-box applicability to other, distinct domains is contingent upon the existence of a high-quality knowledge graph. The construction of such a graph, in turn, relies heavily on access to high-quality, domain-specific knowledge. Furthermore, while our framework makes significant strides in mitigating both under-collaboration and over-collaboration, thereby achieving a demonstrably improved balance and strong collaborative performance, we acknowledge that this represents a notable advancement rather than the definitive apex of collaborative optimization. Defining and reaching what might be considered the ``optimal'' or ``best-in-class'' state for collaboration in these complex workflows remains an ongoing research challenge. This particular challenge stems from a few key issues. Firstly, there is currently limited in-depth research focused on the nuanced spectrum between under- and over-collaboration. Consequently, established benchmarks are also lacking. Such benchmarks would be crucial for precisely defining and measuring what constitutes an optimal point of collaboration in these contexts. To support further research in business workflow tasks, particularly in understanding and navigating these collaborative dynamics to potentially identify higher optima, we will publicly release our developed knowledge graph and associated high-quality datasets.
\clearpage
\newpage
\section{Proof: Tacit Knowledge \& Counterfactual Output Model}
\label{app:first}
Given the inherent difficulty in directly extracting hidden layer knowledge from agents operating within complex systems, our approach introduces counterfactual reasoning as a means to probe and subsequently infer tacit knowledge. This section elucidates the probabilistic formulations developed to model these inferred internal aspects of each agent. It details the conceptualization of tacit knowledge, the generation of counterfactual outputs by the agent, and the overall generative process used to understand these unobservable agent workings.

The foundation of our approach to modeling tacit knowledge within each agent lies in its conditional generation based on latent tacit variables. We define $h_{ij}$ as the tacit knowledge specific to a sample $(X_{ij}, Y_{ij})$ processed by a particular agent. Within an agent, $h_{ij}$ represents an internal state or representation that is not directly accessible. While $h_{ij}$ is an abstract construct for modeling these unobservable aspects, for illustrative purposes, one might conceptualize it as akin to intermediate computational states, learned feature representations, or internal configurations within the Agent that influence its output generation. Concurrently, $\theta_{ij}$ is defined as a latent tacit variable associated with the same sample and agent. This variable $\theta_{ij}$ is also an abstract construct, encapsulating underlying factors hypothesized to influence $h_{ij}$, such as specific input characteristics processed by the Agent, its operational modes, or conditioning contexts it operates under. The introduction of $\theta_{ij}$ and $h_{ij}$ is an attempt to create a probabilistic model for the Agent's internal, unobservable precursors to decision-making or generation.

A core assumption of our model is that this tacit knowledge $h_{ij}$ is not deterministically derived but is rather a stochastic realization conditioned on $\theta_{ij}$. This is particularly relevant for complex Agents, where output generation can be influenced by internal stochastic mechanisms or intricate decision processes. This probabilistic relationship is formally expressed as:
\begin{equation}
h_{ij} \mid \theta_{ij} \sim P(h_{ij} \mid \theta_{ij}) \label{eq8}
\end{equation}
This notation signifies that $h_{ij}$ is a random variable whose probability distribution $P(h_{ij} \mid \theta_{ij})$ is conditional upon the specific value or state of $\theta_{ij}$. The nature of this distribution would be specific to the agent and reflects the inherent complexities and uncertainties in how its internal states are formed. For instance, $\theta_{ij}$ might represent an abstract control signal or a high-level interpretation of input patterns by the agent, and $P(h_{ij} \mid \theta_{ij})$ would describe the distribution of possible internal representations resulting from it. This formulation acknowledges that even with a defined influencing factor $\theta_{ij}$, the precise tacit knowledge $h_{ij}$ (the agent's internal processing focus or state) can vary. The characteristics of $P(h_{ij} \mid \theta_{ij})$ can be explored and potentially learned through systematic counterfactual probing of the agent.

Following the generation of tacit knowledge, we model the generation of the counterfactual output $\tilde{Y}_{ij}$ by the agent. This output is conditioned not only on the counterfactual input $\tilde{X}_{ij}$ but also critically on the realized tacit knowledge $h_{ij}$ derived from Equation~\ref{eq8}. We define $\tilde{Y}_{ij}$ as the Agent's generated output when presented with the counterfactual input $\tilde{X}_{ij}$, and $h_{ij}$ acts as the mediating internal state through which $\tilde{X}_{ij}$ is processed to produce $\tilde{Y}_{ij}$.

The generation of $\tilde{Y}_{ij}$ by such agents is often an inherently probabilistic process. Even with a given counterfactual input $\tilde{X}_{ij}$ and a specific internal tacit knowledge state $h_{ij}$, the agent may not produce a single, deterministic output. This inherent stochasticity is captured by the following conditional probability distribution, whose parameters might also be inferred from counterfactual observations:
\begin{equation}
\tilde{Y}_{ij} \mid h_{ij}, \tilde{X}_{ij} \sim P(\tilde{Y}_{ij} \mid h_{ij}, \tilde{X}_{ij}) \label{eq9}
\end{equation}
This expression states that $\tilde{Y}_{ij}$ (the counterfactual output) is a random variable whose distribution $P(\tilde{Y}_{ij} \mid h_{ij}, \tilde{X}_{ij})$ depends on both the specific tacit knowledge state $h_{ij}$ and the counterfactual input $\tilde{X}_{ij}$. The tacit knowledge $h_{ij}$ here serves as a crucial intermediary. For agents producing sequential outputs, $P(\tilde{Y}_{ij} \mid h_{ij}, \tilde{X}_{ij})$ might be represented as an autoregressive factorization, $P(\tilde{Y}_{ij} \mid h_{ij}, \tilde{X}_{ij}) = \prod_t P(\tilde{y}_{ij,t} \mid \tilde{y}_{ij,<t}, h_{ij}, \tilde{X}_{ij})$.

To determine the overall probability of observing a counterfactual output $\tilde{Y}_{ij}$ from an agent given only the counterfactual input $\tilde{X}_{ij}$—a common scenario when internal states are not directly observable—we must account for the entire hypothesized generative pathway. This involves marginalizing out the unobserved latent tacit variable $\theta_{ij}$ and, by extension, the intermediate tacit knowledge $h_{ij}$. The final generation process is thus expressed as:
\begin{equation}
P(\tilde{Y}_{ij} \mid \tilde{X}_{ij}) = \int_{\Theta} P(\tilde{Y}_{ij} \mid h_{ij}, \tilde{X}_{ij}) P(h_{ij} \mid \theta_{ij}) P(\theta_{ij} \mid \tilde{X}_{ij}) d\theta_{ij}  \label{eq10}
\end{equation}
This equation can be derived by considering the dependencies established in Equations~\ref{eq8} and~\ref{eq9}.

The derivation proceeds as follows:
First, we apply the law of total probability to marginalize over the latent tacit variable $\theta_{ij}$. The probability of $\tilde{Y}_{ij}$ given $\tilde{X}_{ij}$ can be written as an integral over all possible values of $\theta_{ij}$ in its domain $\Theta$:
\[ P(\tilde{Y}_{ij} \mid \tilde{X}_{ij}) = \int_{\Theta} P(\tilde{Y}_{ij}, \theta_{ij} \mid \tilde{X}_{ij}) d\theta_{ij} \]
Using the definition of conditional probability, $P(A, B \mid C) = P(A \mid B, C) P(B \mid C)$, we can rewrite the integrand $P(\tilde{Y}_{ij}, \theta_{ij} \mid \tilde{X}_{ij})$ as:
\[ P(\tilde{Y}_{ij}, \theta_{ij} \mid \tilde{X}_{ij}) = P(\tilde{Y}_{ij} \mid \theta_{ij}, \tilde{X}_{ij}) P(\theta_{ij} \mid \tilde{X}_{ij}) \]
Here, $P(\theta_{ij} \mid \tilde{X}_{ij})$ represents the prior probability distribution of the tacit variable $\theta_{ij}$ given the counterfactual input $\tilde{X}_{ij}$. This term reflects how a given counterfactual input might stochastically lead to different internal configurations or ``operational modes'' $\theta_{ij}$ within the Agent that are not directly observable. The nature of this prior can also be investigated through counterfactual analysis.
Substituting this back, we get:
\begin{equation}
P(\tilde{Y}_{ij} \mid \tilde{X}_{ij}) = \int_{\Theta} P(\tilde{Y}_{ij} \mid \theta_{ij}, \tilde{X}_{ij}) P(\theta_{ij} \mid \tilde{X}_{ij}) d\theta_{ij}  \label{eq11}
\end{equation}
The term $P(\tilde{Y}_{ij} \mid \theta_{ij}, \tilde{X}_{ij})$ is the probability of generating $\tilde{Y}_{ij}$ given both $\theta_{ij}$ and $\tilde{X}_{ij}$. Our model posits a specific structure for this, reflecting a Markov chain-like dependency: $\theta_{ij} \rightarrow h_{ij} \rightarrow \tilde{Y}_{ij}$ (all conditioned on $\tilde{X}_{ij}$ where appropriate). This structure implies two key conditional independence assumptions:
\begin{enumerate}
    \item Given $\theta_{ij}$, $h_{ij}$ is independent of $\tilde{X}_{ij}$ if all influence of $\tilde{X}_{ij}$ on $h_{ij}$ is mediated through $\theta_{ij}$. More commonly, $P(h_{ij} \mid \theta_{ij})$ is defined as the direct influence of $\theta_{ij}$ on $h_{ij}$.
    \item Given $h_{ij}$ and $\tilde{X}_{ij}$, $\tilde{Y}_{ij}$ is independent of $\theta_{ij}$. That is, $P(\tilde{Y}_{ij} \mid h_{ij}, \theta_{ij}, \tilde{X}_{ij}) = P(\tilde{Y}_{ij} \mid h_{ij}, \tilde{X}_{ij})$. The tacit knowledge $h_{ij}$ fully mediates the influence of $\theta_{ij}$ on $\tilde{Y}_{ij}$.
\end{enumerate}
Under these assumptions, we can decompose $P(\tilde{Y}_{ij} \mid \theta_{ij}, \tilde{X}_{ij})$:
\[ P(\tilde{Y}_{ij} \mid \theta_{ij}, \tilde{X}_{ij}) = \int P(\tilde{Y}_{ij}, h_{ij} \mid \theta_{ij}, \tilde{X}_{ij}) dh_{ij} \]
\[ = \int P(\tilde{Y}_{ij} \mid h_{ij}, \theta_{ij}, \tilde{X}_{ij}) P(h_{ij} \mid \theta_{ij}, \tilde{X}_{ij}) dh_{ij} \]
Applying the conditional independence assumptions:
\[ = \int P(\tilde{Y}_{ij} \mid h_{ij}, \tilde{X}_{ij}) P(h_{ij} \mid \theta_{ij}) dh_{ij} \]
The form in Equation~\ref{eq10} implies that for a given $\theta_{ij}$ in the outer integral, the relevant $h_{ij}$ is the one generated from that $\theta_{ij}$, effectively collapsing the inner integral. Thus, $P(\tilde{Y}_{ij} \mid \theta_{ij}, \tilde{X}_{ij})$ is represented by the product $P(\tilde{Y}_{ij} \mid h_{ij}, \tilde{X}_{ij}) P(h_{ij} \mid \theta_{ij})$, where $h_{ij}$ is the specific realization tied to $\theta_{ij}$ within the integral.
Substituting this construction into Equation~\ref{eq11} directly yields Equation~\ref{eq10}:
\[ P(\tilde{Y}_{ij} \mid \tilde{X}_{ij}) = \int_{\Theta} P(\tilde{Y}_{ij} \mid h_{ij}, \tilde{X}_{ij}) P(h_{ij} \mid \theta_{ij}) P(\theta_{ij} \mid \tilde{X}_{ij}) d\theta_{ij} \]
This integral sums the probabilities of $\tilde{Y}_{ij}$ occurring via all possible latent tacit variables $\theta_{ij}$. Each path's contribution is weighted by the prior probability of $\theta_{ij}$, the probability of the corresponding tacit knowledge $h_{ij}$ arising from that $\theta_{ij}$, and finally the probability of the output $\tilde{Y}_{ij}$ given that $h_{ij}$ and the input $\tilde{X}_{ij}$.

\section{Implementation Details}
\label{app:second}
\textbf{Definition of Business Workflow:} It is a complex process with many interconnected stages. These stages are specialized. The key feature is that these stages strongly depend on each other. A decision in one stage requires more than just expert knowledge of that area. It must also anticipate and use knowledge and constraints from the other stages.
%Table~\ref{tab:dataset-statistics} presents a detailed overview of the datasets selected from the MSCoRe benchmark, which include three business workflow datasets comprising eleven subsets. Notably, we invited experts from a major automotive manufacturing enterprise to further calibrate the automotive datasets, enabling more effective evaluations aligned with real-world scenarios.

Table~\ref{tabf3} presents a comprehensive comparison of Cochain against other baseline models. Cochain demonstrates superior agent collaboration, characterized by faster inference speeds and lower hallucination rates. More critically, Cochain exhibits decomposability, ensuring that the collaboration process is resilient and does not terminate due to failures in individual components. It is important to note that for CoA, source documents are not included due to the differing task orientation.

%\begin{table}[h]
%  \caption{Dataset Statistics. The table shows detailed information about three domain-specific business workflow datasets.} %$^{\dagger}$Seed sources indicate the origin of initial data from which the complete datasets were generated using the Self-Instruct~\cite{wang2022self} methodology.} %Pharmaceutical R\&D includes drug screening, clinical trials, and regulatory approval processes; Materials covers API procurement and excipient supply; Manufacturing includes GMP production and quality control; Supply chain involves cold chain transportation and warehouse management; Sales \& Distribution covers hospitals, pharmacies, and e-commerce channels.}
%  \centering
%\resizebox{\textwidth}{!}{
%\begin{tabular}{lccccccccccc}
%\toprule
%\multirow{2}{*}{Datasets} & \multicolumn{4}{c}{Automotive} & \multicolumn{3}{c}{Pharmaceutical} & \multicolumn{4}{c}{E-commerce} \\
%\cmidrule(lr){2-5} \cmidrule(lr){6-8} \cmidrule(lr){9-12}
%& Design & Production & Supply & QC & R\&D & Mfg. & Sales & Procure. & Logistics & Operations & UX \\
%\midrule
%Total samples & 24884 & 20798 & 19264 & 7252 & 4881 & 4776 & 4851 & 4627 & 4630 & 4312 & 3910 \\
%\midrule
%Avg. tokens & 958 & 943 & 984 & 870 & 1656 & 1601 & 1623 & 919 & 901 & 917 & 880 \\
%\bottomrule
%\end{tabular}}
%\label{tab:dataset-statistics}
%\end{table}
%Seed source$^{\dagger}$ & \multicolumn{4}{c}{Industry reports \& expert reviewes} & \multicolumn{3}{c}{Clinical databases} & \multicolumn{4}{c}{Platform analytics} \\
The LLMs referenced in Figure 5 (Main Paper) are fine-tuned using QLoRA~\cite{dettmers2023qlora}, configured with a LoRA rank of 8 and trained for 3 epochs, on two H100 GPUs, with models quantized to 4-bit precision via bitsandbytes. Other open-source models are fine-tuned for 10 epochs using LoRA~\cite{hu2022lora}. The inference times reported in Figure~\ref{time} and Table~\ref{tabf2} are benchmarked by deploying the open-source models on a single A40 GPU to mitigate network variability.

\begin{table}[h]
  \caption{Comparison of different multi-agent collaboration methods}
  \centering
  \resizebox{\textwidth}{!}{
  \begin{tabular}{lcccccc}
    \toprule
    Baseline & Collaboration Method & Decomposable & Reasoning Speed & Hallucination Rate & Agent  \\
    \midrule
    PMC~\cite{zhang2025planning} & Hierarchical Planning & \ding{55} & Low & Middle & Multiple  \\
    MedAgents~\cite{tang2023medagents} & Discussion \& Voting & \ding{55} & Low & Middle & Multiple  \\
    Debate(short)~\cite{du2023improving} & Debate & \ding{55} & Low & Middle & Multiple  \\
    Debate(long)~\cite{du2023improving} & Debate & \ding{55} & Low & Middle & Multiple  \\
    CoA~\cite{zhang2024coa} & Chain Collaboration & \ding{55} & Low & Middle & Multiple  \\
    Original & - & \checkmark & High & High & Single  \\
    \midrule
    Cochain & Knowledge Fusion & \checkmark & High & Low & Multiple  \\
    \bottomrule
  \end{tabular}
  }
  \label{tabf3}
\end{table}

The appendix provides a more extensive report on experimental results across additional metrics. These include BERTScore-P, BERTScore-R, BLEU-4~\cite{papineni2002bleu} for measuring text alignment, METEOR~\cite{banerjee2005meteor} which considers semantically equivalent phrases, ROUGE-2 for assessing content breadth, and ROUGE-L~\cite{lin2004rouge} which evaluates deep semantic alignment, particularly for long texts.

\section{More Experiments and Analysis}
\label{app:third}

\subsection{ More Efficiency and Cost Analysis}
\label{app:third.3}

\begin{figure}[h] 
    \centering
    \includegraphics[width=0.40\textwidth]{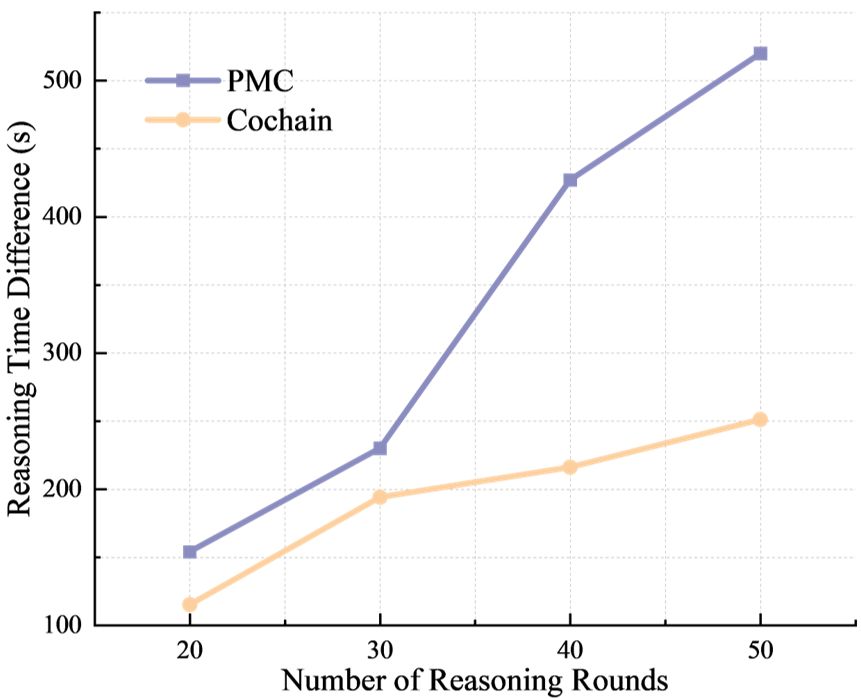}
    \caption{ExampReasoning time differential.}
    \label{time}
\end{figure}

Table~\ref{tabf2} presents a comparative analysis of Cochain against other multi-agent systems, focusing on key operational metrics: Inference Time, Token Throughput, and Cost. A noteworthy observation on the Automotive and E-commerce datasets is that Cochain's average input token count is approximately one-twentieth of that utilized by PMC and CoA. This disparity suggests that these alternative systems expend a considerable volume of tokens on inter-agent communication. In contrast, Cochain innovatively transforms such communication into structured knowledge graphs and prompts tree, a mechanism that effectively curtails operational costs. The overall results compellingly demonstrate that Cochain achieves substantial reductions in inference time, token throughput, and aggregate cost when benchmarked against other multi-agent approaches. \textbf{Remarkably, the efficiency metrics for Cochain closely approximate those observed for single-agent systems, and its average cost is even lower}, underscoring a significant advantage of our proposed framework in terms of computational resource utilization and operational expenditure. Table~\ref{tab:latency} presents the retrieval contribution of each component during the inference process. The prompt tree retrieval employs a depth-first search (DFS) strategy. In terms of time complexity, this is equivalent to a simple path traversal with a complexity of $O(D)$, where $D$ denotes the depth of the business process. Figure~\ref{time} shows that as reasoning rounds increase, the time gap between thinking and non-thinking models grows dramatically for PMC but remains moderate for Cochain, demonstrating Cochain's superior efficiency as complexity increases.

\begin{table}[h]
\caption{Comparative analysis of inference time, token throughput, and cost. $^{\dagger}$Inference time is measured using a locally deployed Qwen2-7B model to obviate network latency associated with API calls. Conversely, token throughput and cost metrics are derived using Claude-3.5-Haiku.}
\centering
\begin{tabular}{lcccc}
\toprule
Baselines & $^{\dagger}$Inference Time (s) & Avg.Input & Avg.Output & Avg.Cost (\$)\\
\midrule
\multicolumn{5}{c}{\textbf{Automotive}} \\
\midrule
PMC~\cite{zhang2025planning} & 241.56 & 11637.94 & 6644.22 & 0.0359 \\
MedAgents~\cite{tang2023medagents} & 196.53 & 3901.03 & 3483.24 & 0.0170 \\
Debate(short)~\cite{du2023improving} & 187.16 & 3938.35 & 4809.27 & 0.0223 \\
Debate(long)~\cite{du2023improving} & 162.57 & 4357.87 & 5278.53 & 0.0246 \\
CoA~\cite{zhang2024coa} & 227.21 & 11428.06 & 5446.62 & 0.0309 \\
Original & 37.22 & 93.49 & 1041.38 & \underline{0.0042} \\
Cochain & 38.77 & 501.62 & 541.74 & \textbf{0.0026} \\
\midrule
\multicolumn{5}{c}{\textbf{Pharmaceutical}} \\
\midrule
PMC~\cite{zhang2025planning} & 198.49 & 3796.07 & 2613.73 & 0.0135 \\
MedAgents~\cite{tang2023medagents} & 169.53 & 3504.85 & 3504.85 & 0.0168 \\
Debate(short)~\cite{du2023improving} & 548.66 & 3372.56 & 3772.79 & 0.0178 \\
Debate(long)~\cite{du2023improving} & 574.95 & 3810.87 & 4268.74 & 0.0201 \\
CoA~\cite{zhang2024coa} & 253.18 & 5493.94 & 3450.45 & 0.0182 \\
Original & 60.96 & 77.28 & 1091.32 & \underline{0.0044} \\
Cochain & 44.70 & 533.88 & 830.34 & \textbf{0.0037}\\
\midrule
\multicolumn{5}{c}{\textbf{E-commerce}} \\
\midrule
PMC~\cite{zhang2025planning} & 256.84 & 11319.74 & 6452.89 & 0.0349 \\
MedAgents~\cite{tang2023medagents} & 181.61 & 3816.26 & 3445.34 & 0.0169 \\
Debate(short)~\cite{du2023improving} & 318.51 & 4414.23 & 5660.94 & 0.0261 \\
Debate(long)~\cite{du2023improving} & 293.42 & 5438.52 & 6911.85 & 0.0321 \\
CoA~\cite{zhang2024coa} & 195.49 & 11446.05 & 5353.24 & 0.0306 \\
Original & 35.83 & 60.90 & 1080.92 & \underline{0.0044} \\
Cochain & 37.10 & 561.80 & 620.15 & \textbf{0.0029} \\
\bottomrule
\end{tabular}
\label{tabf2}
\end{table}

\begin{table}[h]
\centering
\caption{Latency Breakdown Analysis of Cochain's Inference Components}
\label{tab:latency}
\begin{tabular}{lrr}
\toprule
\textbf{Inference Component} & \textbf{Average Time (s)} & \textbf{Percentage of Total Time (\%)} \\
\midrule
Knowledge Graph Retrieval & 0.6200 s & 1.57 \% \\
Prompts Tree Retrieval    & 0.0022 s & \textless 0.01 \% \\
LLM Model Inference       & 38.7700 s & 98.42 \% \\
\bottomrule
\end{tabular}
\end{table}

\clearpage

\subsection{More Experiments on Single Agent}
\label{app:third.1}
Table~\ref{tabf1} and Table~\ref{tabf4} demonstrate that integrating Cochain markedly enhances single-agent performance across diverse backbone models during two distinct stages of a business workflow. This enhancement is consistently reflected across all evaluated metrics, irrespective of whether the LLMs are in their raw, pre-trained state or have undergone fine-tuning. Notably, a synergistic effect between fine-tuning and the application of Cochain typically yields optimal performance, underscoring Cochain's robust capability to leverage specialized model knowledge. The broad-spectrum improvements observed across various model metrics highlight the effectiveness and versatility of Cochain in mitigating prior issues of insufficient collaboration.

\begin{table*}[h]
    \caption{The performance of our proposed method on single-agent. BERTScore F1 is abbreviated as BS-F, BERTScore Precision as BS-P, and BERTScore Recall as BS-R. We highlight the \sethlcolor{bestcolor}\hl{best} and \sethlcolor{secondbestcolor}\hl{second-best} results.}
    \centering
    \resizebox{\textwidth}{!}{
    \begin{tabular}{llcccccccc}
        \toprule
        Backbone & Method & BS-F & BS-P & BS-R & BLEU-4 & GLEU & METEOR & ROUGE-2 & ROUGE-L \\
        \midrule
        \multirow{4}{*}{Llama2-7B} 
        & Raw LLM & 62.25 & 64.41 & 60.54 & 6.04 & 8.28 & 17.52 & 7.78 & 15.64 \\
        & + Cochain  & 65.70 & 68.25 & 63.50 & 10.99 & 13.84 & 25.50 & 10.44 & 16.46 \\
        & Finetuned LLM & \cellcolor[HTML]{E0F7FF}70.68 & \cellcolor[HTML]{E0F7FF}70.71 & \cellcolor[HTML]{E0F7FF}70.67 & \cellcolor[HTML]{E0F7FF}14.31 & \cellcolor[HTML]{E0F7FF}18.79 & \cellcolor[HTML]{E0F7FF}35.03 & \cellcolor[HTML]{E0F7FF}11.97 & \cellcolor[HTML]{E0F7FF}21.49 \\
        & + Cochain  & \cellcolor[HTML]{BBDEFB}71.46 & \cellcolor[HTML]{BBDEFB}71.35 & \cellcolor[HTML]{BBDEFB}71.59 & \cellcolor[HTML]{BBDEFB}16.50 & \cellcolor[HTML]{BBDEFB}20.96 & \cellcolor[HTML]{BBDEFB}38.69 & \cellcolor[HTML]{BBDEFB}13.79 & \cellcolor[HTML]{BBDEFB}21.58 \\
        \midrule
        \multirow{4}{*}{Qwen2-7B}
        & Raw LLM & 71.76 & 71.79 & 71.75 & 17.23 & 22.30 & 41.13 & 12.88 & 24.00 \\
        & + Cochain  & \cellcolor[HTML]{E0F7FF}75.24 & \cellcolor[HTML]{E0F7FF}74.90 & \cellcolor[HTML]{E0F7FF}75.58 & \cellcolor[HTML]{E0F7FF}22.32 & \cellcolor[HTML]{E0F7FF}26.51 & \cellcolor[HTML]{BBDEFB}52.00 & \cellcolor[HTML]{E0F7FF}19.53 & \cellcolor[HTML]{E0F7FF}27.94 \\
        & Finetuned LLM & 71.72 & 71.75 & 71.70 & 16.90 & 21.64 & 40.77 & 13.35 & 24.07 \\
        & + Cochain  & \cellcolor[HTML]{BBDEFB}75.39 & \cellcolor[HTML]{BBDEFB}75.14 & \cellcolor[HTML]{BBDEFB}75.65 & \cellcolor[HTML]{BBDEFB}24.17 & \cellcolor[HTML]{BBDEFB}28.17 & \cellcolor[HTML]{E0F7FF}51.22 & \cellcolor[HTML]{BBDEFB}21.22 & \cellcolor[HTML]{BBDEFB}28.87 \\
        \midrule
        \multirow{4}{*}{Llama3-8B}
        & Raw LLM & 68.93 & 68.38 & 69.59 & 7.47 & 9.14 & 24.48 & 10.99 & 19.28 \\
        & + Cochain  & \cellcolor[HTML]{E0F7FF}73.75 & \cellcolor[HTML]{E0F7FF}73.36 & \cellcolor[HTML]{E0F7FF}74.17 & \cellcolor[HTML]{E0F7FF}19.47 & \cellcolor[HTML]{E0F7FF}23.52 & \cellcolor[HTML]{E0F7FF}48.43 & \cellcolor[HTML]{E0F7FF}17.57 & \cellcolor[HTML]{E0F7FF}25.10 \\
        & Finetuned LLM & 70.89 & 70.88 & 70.91 & 14.90 & 19.05 & 39.83 & 12.59 & 22.90 \\
        & + Cochain  & \cellcolor[HTML]{BBDEFB}74.56 & \cellcolor[HTML]{BBDEFB}74.13 & \cellcolor[HTML]{BBDEFB}75.01 & \cellcolor[HTML]{BBDEFB}22.12 & \cellcolor[HTML]{BBDEFB}26.15 & \cellcolor[HTML]{BBDEFB}50.59 & \cellcolor[HTML]{BBDEFB}19.41 & \cellcolor[HTML]{BBDEFB}27.19 \\
        \midrule
        \multirow{4}{*}{Llama2-13B} 
        & Raw LLM & 60.79 & 63.16 & 59.00 & 5.73 & 7.96 & 16.90 & 7.52 & 15.43 \\
        & + Cochain  & 67.47 & 70.20 & 65.15 & 13.42 & 16.88 & 29.97 & \cellcolor[HTML]{E0F7FF}13.65 & 18.57 \\
        & Finetuned LLM & \cellcolor[HTML]{E0F7FF}70.88 & \cellcolor[HTML]{E0F7FF}70.88 & \cellcolor[HTML]{E0F7FF}70.90 & \cellcolor[HTML]{E0F7FF}14.57 & \cellcolor[HTML]{E0F7FF}19.14 & \cellcolor[HTML]{E0F7FF}35.87 & 12.19 & \cellcolor[HTML]{E0F7FF}21.85 \\
        & + Cochain & \cellcolor[HTML]{BBDEFB}71.80 & \cellcolor[HTML]{BBDEFB}72.08 & \cellcolor[HTML]{BBDEFB}71.58 & \cellcolor[HTML]{BBDEFB}18.16 & \cellcolor[HTML]{BBDEFB}22.29 & \cellcolor[HTML]{BBDEFB}40.33 & \cellcolor[HTML]{BBDEFB}14.96 & \cellcolor[HTML]{BBDEFB}23.18 \\
        \midrule
        \multirow{4}{*}{Qwen2.5-14B} 
        & Raw LLM & 71.86 & 72.03 & 71.70 & 9.48 & 12.16 & 40.71 & 11.85 & 17.86 \\
        & + Cochain & \cellcolor[HTML]{BBDEFB}75.62 & \cellcolor[HTML]{BBDEFB}75.56 & \cellcolor[HTML]{BBDEFB}75.69 & \cellcolor[HTML]{E0F7FF}12.39 & \cellcolor[HTML]{E0F7FF}14.71 & \cellcolor[HTML]{E0F7FF}47.95 & \cellcolor[HTML]{E0F7FF}17.66 & \cellcolor[HTML]{E0F7FF}19.55 \\
        & Finetuned LLM & 71.50 & 71.48 & 71.53 & 9.25 & 11.75 & 40.16 & 12.01 & 17.67 \\
        & + Cochain  & \cellcolor[HTML]{E0F7FF}75.18 & \cellcolor[HTML]{E0F7FF}75.00 & \cellcolor[HTML]{E0F7FF}75.38 & \cellcolor[HTML]{BBDEFB}13.11 & \cellcolor[HTML]{BBDEFB}15.08 & \cellcolor[HTML]{BBDEFB}48.35 & \cellcolor[HTML]{BBDEFB}19.70 & \cellcolor[HTML]{BBDEFB}20.30 \\
        \bottomrule
    \end{tabular}}
    \label{tabf1}
\end{table*}

\begin{table}[h]
    \centering
    \caption{The performance of our proposed method. The results of this table are in a different stage, as compared to the stage presented in Table~\ref{tabf1}. We highlight the \sethlcolor{bestcolor}\hl{best} and \sethlcolor{secondbestcolor}\hl{second-best} results.}
    \resizebox{\textwidth}{!}{
    \begin{tabular}{lccccccccc}
        \toprule
        Backbone & Method & BS-F & BS-P & BS-R & BLEU-4 & GLEU & METEOR & ROUGE-2 & ROUGE-L \\
        \midrule
        \multirow{4}{*}{Llama2-7B} 
        & \multicolumn{1}{l}{Raw LLM} & 61.76 & 64.72 & 59.44 & 6.36 & 8.69 & 18.25 & 8.66 & 16.17 \\
        & \multicolumn{1}{l}{+ Cochain} & 63.57 & 66.31 & 61.23 & 9.49 & 12.37 & 23.12 & 8.99 & 17.46 \\
        & \multicolumn{1}{l}{Finetuned LLM} & \cellcolor[HTML]{E0F7FF}69.88 & \cellcolor[HTML]{E0F7FF}69.78 & \cellcolor[HTML]{E0F7FF}70.00 & \cellcolor[HTML]{E0F7FF}12.02 & \cellcolor[HTML]{E0F7FF}17.02 & \cellcolor[HTML]{E0F7FF}32.59 & \cellcolor[HTML]{E0F7FF}9.26 & \cellcolor[HTML]{BBDEFB}20.69 \\
        & \multicolumn{1}{l}{+ Cochain} & \cellcolor[HTML]{BBDEFB}70.47 & \cellcolor[HTML]{BBDEFB}70.43 & \cellcolor[HTML]{BBDEFB}70.55 & \cellcolor[HTML]{BBDEFB}13.79 & \cellcolor[HTML]{BBDEFB}18.58 & \cellcolor[HTML]{BBDEFB}35.44 & \cellcolor[HTML]{BBDEFB}10.91 & \cellcolor[HTML]{E0F7FF}20.26 \\
        \midrule
        \multirow{4}{*}{Qwen2-7B}
        & \multicolumn{1}{l}{Raw LLM} & 70.34 & 71.12 & 70.60 & 13.74 & 19.88 & 35.29 & 9.47 & 22.01 \\
        & \multicolumn{1}{l}{+ Cochain} & \cellcolor[HTML]{E0F7FF}73.48 & \cellcolor[HTML]{E0F7FF}72.37 & \cellcolor[HTML]{E0F7FF}74.63 & \cellcolor[HTML]{E0F7FF}15.77 & \cellcolor[HTML]{E0F7FF}19.21 & \cellcolor[HTML]{E0F7FF}50.38 & \cellcolor[HTML]{E0F7FF}16.25 & \cellcolor[HTML]{E0F7FF}21.99 \\
        & \multicolumn{1}{l}{Finetuned LLM} & 70.28 & 70.08 & 70.50 & 13.61 & 18.95 & 35.07 & 9.96 & 22.38 \\
        & \multicolumn{1}{l}{+ Cochain} & \cellcolor[HTML]{BBDEFB}74.51 & \cellcolor[HTML]{BBDEFB}74.30 & \cellcolor[HTML]{BBDEFB}74.74 & \cellcolor[HTML]{BBDEFB}21.99 & \cellcolor[HTML]{BBDEFB}25.76 & \cellcolor[HTML]{BBDEFB}50.49 & \cellcolor[HTML]{BBDEFB}20.71 & \cellcolor[HTML]{BBDEFB}26.00 \\
        \bottomrule
    \end{tabular}}
    \label{tabf4}
\end{table}

\begin{table}[t]
\centering
\caption{
    Quantifying insufficient collaboration.
}
\label{tab:quantify_collaboration}
\begin{tabular}{llccc}
\toprule
\textbf{Backbone} & \textbf{Method} & \textbf{BS-F} & \textbf{GLEU} & \textbf{ROUGE-L} \\
\midrule
\multirow{3}{*}{Qwen2-7B} & Single Agent & 71.72 & 21.64 & 24.07 \\
& + Manual Prompts & 72.54 ($\uparrow$) & 21.83 ($\uparrow$) & 24.34 ($\uparrow$) \\
& \textbf{+ Cochain} & \textbf{75.39 ($\Uparrow$)} & \textbf{28.17 ($\Uparrow$)} & \textbf{28.87 ($\Uparrow$)} \\
\midrule
\multirow{3}{*}{DeepSeek-R1-7B} & Single Agent & 71.35 & 19.55 & 22.71 \\
& + Manual Prompts & 71.83 ($\uparrow$) & 19.90 ($\uparrow$) & 23.38 ($\uparrow$) \\
& \textbf{+ Cochain} & \textbf{73.19 ($\Uparrow$)} & \textbf{21.71 ($\Uparrow$)} & \textbf{24.01 ($\Uparrow$)} \\
\bottomrule
\end{tabular}
\end{table}

As shown in Table~\ref{tab:quantify_collaboration}, to better quantify the impact of insufficient collaboration, we compare the baseline single agent to one provided with minimal communication, simulated by manually injecting five prompts containing cross-domain constraints. The results of this table clearly show: First, insufficient collaboration is a real problem, as the agent with injected communication showed a slight performance improvement over the original agent. This directly proves that adding cross-domain communication improves outcomes, thereby quantifying the performance loss from insufficient collaboration. Second, Cochain's intelligent collaboration is far superior to simple communication, as the performance improved substantially after applying Cochain. This demonstrates that Cochain's value is not just about adding information. Rather, it excels at intelligently filtering, organizing, and structuring collaborative knowledge. Its effectiveness far surpasses that of unstructured, manually added information and helps to avoid the risk of excessive collaboration.

\subsection{More Experiments on Specialized Agent Collaboration}
\label{app:third.2}

Figure 6 (Main Paper) presents the results from a distinct experimental configuration designed to further investigate the coordination capabilities of Cochain. Whereas Table~\ref{tabf1} assessed a multi-domain model (a single Qwen2-7B model fine-tuned on all domains) and specialized models (different models each fine-tuned for specific domains), Table~\ref{tabf5} exclusively focuses on a setup utilizing multiple agents of the same backbone, each independently fine-tuned for a specific domain. The findings indicate that Cochain effectively facilitates collaboration among these identically architected yet uniquely specialized agents. Performance enhancements are observed both within their respective specialized domains and in the quality of the comprehensive, integrated solutions. This underscores Cochain's distinct advantage in synergizing specialized knowledge from a team of uniquely trained agents that share a common architectural foundation.

\begin{table*}[h]
    \centering
    \caption{Results for one model at particular stages. This means that we use Qwen2-7B for four fine-tunings, and the four fine-tuned models are experimented with at their respective stages. We highlight the \sethlcolor{secondbestcolor}\hl{fine-tuned} results, \sethlcolor{bestcolor}\hl{Cochain} application, and \sethlcolor{red!10}\hl{\mbox{improvements\text{\scriptsize{$(\uparrow)$}}}}.}
    \begin{tabular}{cccccccccc}
        \toprule
        Backbone & Stage & Method & BS-F & BS-P & BS-R & BLEU-4 & GLEU & ROUGE-L \\
        \midrule
        \multirow{12}{*}{\rotatebox{90}{Qwen2-7B}} & \multirow{3}{*}{$S_1$} 
        & \cellcolor[HTML]{E0F7FF}Finetuned & \cellcolor[HTML]{E0F7FF}71.72 & \cellcolor[HTML]{E0F7FF}71.75 & \cellcolor[HTML]{E0F7FF}71.70  & \cellcolor[HTML]{E0F7FF}16.90 & \cellcolor[HTML]{E0F7FF}21.64 & \cellcolor[HTML]{E0F7FF}24.07 \\
        &  & \cellcolor[HTML]{BBDEFB}+ Cochain & \cellcolor[HTML]{BBDEFB}75.39 & \cellcolor[HTML]{BBDEFB}75.14 & \cellcolor[HTML]{BBDEFB}75.65 & \cellcolor[HTML]{BBDEFB}24.17 & \cellcolor[HTML]{BBDEFB}28.17 & \cellcolor[HTML]{BBDEFB}28.87 \\
        &  & \cellcolor{red!10}$\uparrow$ & \cellcolor{red!10}+3.67 & \cellcolor{red!10}+3.39 & \cellcolor{red!10}+3.95  & \cellcolor{red!10}+7.27 & \cellcolor{red!10}+6.53 & \cellcolor{red!10}+4.80 \\
        \cmidrule{2-9}
        & \multirow{3}{*}{$S_2$} 
        & \cellcolor[HTML]{E0F7FF}Finetuned & \cellcolor[HTML]{E0F7FF}70.24 & \cellcolor[HTML]{E0F7FF}70.47 & \cellcolor[HTML]{E0F7FF}70.16  & \cellcolor[HTML]{E0F7FF}13.19 & \cellcolor[HTML]{E0F7FF}18.29 & \cellcolor[HTML]{E0F7FF}20.95 \\
        &  & \cellcolor[HTML]{BBDEFB}+ Cochain & \cellcolor[HTML]{BBDEFB}74.23 & \cellcolor[HTML]{BBDEFB}74.01 & \cellcolor[HTML]{BBDEFB}74.46  & \cellcolor[HTML]{BBDEFB}20.05 & \cellcolor[HTML]{BBDEFB}23.57 & \cellcolor[HTML]{BBDEFB}24.71 \\
        &  & \cellcolor{red!10}$\uparrow$ & \cellcolor{red!10}+3.99 & \cellcolor{red!10}+3.54 & \cellcolor{red!10}+4.30  & \cellcolor{red!10}+6.86 & \cellcolor{red!10}+5.28 & \cellcolor{red!10}+3.76 \\
        \cmidrule{2-9}
        & \multirow{3}{*}{$S_3$} 
        & \cellcolor[HTML]{E0F7FF}Finetuned & \cellcolor[HTML]{E0F7FF}70.28 & \cellcolor[HTML]{E0F7FF}70.08 & \cellcolor[HTML]{E0F7FF}70.50  & \cellcolor[HTML]{E0F7FF}13.61 & \cellcolor[HTML]{E0F7FF}18.95 & \cellcolor[HTML]{E0F7FF}22.38 \\
        &  & \cellcolor[HTML]{BBDEFB}+ Cochain & \cellcolor[HTML]{BBDEFB}74.51 & \cellcolor[HTML]{BBDEFB}74.30 & \cellcolor[HTML]{BBDEFB}74.74  & \cellcolor[HTML]{BBDEFB}21.99 & \cellcolor[HTML]{BBDEFB}25.76 & \cellcolor[HTML]{BBDEFB}26.00 \\
        &  & \cellcolor{red!10}$\uparrow$ & \cellcolor{red!10}+4.23 & \cellcolor{red!10}+4.22 & \cellcolor{red!10}+4.24  & \cellcolor{red!10}+8.38 & \cellcolor{red!10}+6.81 & \cellcolor{red!10}+3.62 \\
        \cmidrule{2-9}
        & \multirow{3}{*}{$S_4$} 
        & \cellcolor[HTML]{E0F7FF}Finetuned & \cellcolor[HTML]{E0F7FF}69.40 & \cellcolor[HTML]{E0F7FF}69.49 & \cellcolor[HTML]{E0F7FF}69.32  & \cellcolor[HTML]{E0F7FF}13.51 & \cellcolor[HTML]{E0F7FF}18.26 & \cellcolor[HTML]{E0F7FF}21.82 \\
        &  & \cellcolor[HTML]{BBDEFB}+ Cochain & \cellcolor[HTML]{BBDEFB}75.06 & \cellcolor[HTML]{BBDEFB}74.99 & \cellcolor[HTML]{BBDEFB}75.15  & \cellcolor[HTML]{BBDEFB}23.95 & \cellcolor[HTML]{BBDEFB}27.70 & \cellcolor[HTML]{BBDEFB}27.49 \\
        &  & \cellcolor{red!10}$\uparrow$ & \cellcolor{red!10}+5.66 & \cellcolor{red!10}+5.50 & \cellcolor{red!10}+5.83  & \cellcolor{red!10}+10.44 & \cellcolor{red!10}+9.44 & \cellcolor{red!10}+5.67 \\
        \midrule
        & \multicolumn{2}{c}{Score} & \multicolumn{2}{c}{\cellcolor[HTML]{E0F7FF}6.4} & \multicolumn{2}{c}{\cellcolor[HTML]{BBDEFB}9.5} & \multicolumn{2}{c}{\cellcolor{red!10}+3.1} \\
        \bottomrule
    \end{tabular}
    \label{tabf5}
\end{table*}

%\sethlcolor{red!10}\hl{improvements\text{\scriptsize{$(\uparrow)$}}}.

\subsection{More Results from Ablation Studies}
\label{app:third.4}
On the automotive dataset, we report additional ablation study results focusing on the impact of different backbone models. As indicated in Table~\ref{tabf6}, ``w/o Prompts Tree'' markedly impaired performance. Specifically, when applied to the Llama-8B backbone, this ablation led to a 1.97\% reduction in the BS-F score, with notable degradations also observed across other evaluation metrics. These findings are consistent with the experimental results presented in Table 5 (Main Paper), further underscoring the critical role of the Prompts Tree component in our framework.

\begin{table*}[h]
    \caption{More results of the ablation study. \(\mathcal{CKG}\) represents the collaborative knowledge graph. We highlight the \sethlcolor{red!30}\hl{most} and \sethlcolor{red!10}\hl{second-most} efficient modules.}
    \centering
    \resizebox{\textwidth}{!}{
    \begin{tabular}{lccccccccc}
        \toprule
        Backbone & Method  & BS-F & BS-P & BS-R & BLEU-4 & GLEU & METEOR & ROUGE-2 & ROUGE-L \\
        \midrule
        \multirow{4}{*}{Llama2-7B} 
        & \multicolumn{1}{l}{Cochain} & 71.46 & 71.35 & 71.59 & 16.50 & 20.96 & 38.69 & 13.79 & 21.58 \\
        & \multicolumn{1}{l}{w/o \(\mathcal{CKG}\)} & \cellcolor{red!10}71.26 & 71.14 & \cellcolor{red!10}71.41 & 16.06 & 20.53 & \cellcolor{red!10}38.21 & \cellcolor{red!10}13.55 & 21.50 \\
        & \multicolumn{1}{l}{w/o Causal Chain} & \cellcolor{red!10}71.26 & \cellcolor{red!10}71.13 & 71.44 & \cellcolor{red!10}15.63 & \cellcolor{red!30}19.59 & 40.32 & 14.33 & \cellcolor{red!30}21.16 \\
        & \multicolumn{1}{l}{w/o Prompts Tree} & \cellcolor{red!30}71.06 & \cellcolor{red!30}70.92 & \cellcolor{red!30}71.20 & \cellcolor{red!30}15.20 & \cellcolor{red!10}19.61 & \cellcolor{red!30}36.66 & \cellcolor{red!30}13.07 & \cellcolor{red!10}21.38 \\
        \midrule
        \multirow{4}{*}{Qwen2-7B} 
        & \multicolumn{1}{l}{Cochain} & 75.39 & 75.14 & 75.65 & 24.17 & 28.17 & 51.22 & 21.22 & 28.87 \\
        & \multicolumn{1}{l}{w/o \(\mathcal{CKG}\)} & 75.21 & \cellcolor{red!10}74.91 & 75.53 & \cellcolor{red!10}23.56 & 27.81 & \cellcolor{red!10}49.76 & \cellcolor{red!10}20.15 & 28.74 \\
        & \multicolumn{1}{l}{w/o Causal Chain} & \cellcolor{red!10}75.20 & 74.92 & \cellcolor{red!10}75.50 & 23.68 & \cellcolor{red!10}27.55 & 51.00 & 21.14 & \cellcolor{red!10}28.42 \\
        & \multicolumn{1}{l}{w/o Prompts Tree} & \cellcolor{red!30}73.08 & \cellcolor{red!30}73.20 & \cellcolor{red!30}72.98 & \cellcolor{red!30}18.40 & \cellcolor{red!30}22.45 & \cellcolor{red!30}45.43 & \cellcolor{red!30}16.06 & \cellcolor{red!30}24.56 \\
        \midrule
        \multirow{4}{*}{Llama3-8B} 
        & \multicolumn{1}{l}{Cochain} & 74.56 & 74.13 & 75.01 & 22.12 & 26.15 & 50.59 & 19.41 & 27.19 \\
        & \multicolumn{1}{l}{w/o \(\mathcal{CKG}\)} & 74.27 & 73.80 & 74.75 & 21.45 & 25.46 & 49.99 & 19.09 & 27.13 \\
        & \multicolumn{1}{l}{w/o Causal Chain} & \cellcolor{red!10}73.63 & \cellcolor{red!10}72.79 & \cellcolor{red!10}74.50 & \cellcolor{red!10}17.34 & \cellcolor{red!30}20.70 & \cellcolor{red!10}49.15 & \cellcolor{red!10}18.28 & \cellcolor{red!30}23.78 \\
        & \multicolumn{1}{l}{w/o Prompts Tree} & \cellcolor{red!30}72.59 & \cellcolor{red!30}72.45 & \cellcolor{red!30}72.75 & \cellcolor{red!30}17.01 & \cellcolor{red!10}21.02 & \cellcolor{red!30}45.00 & \cellcolor{red!30}15.18 & \cellcolor{red!10}23.83 \\
        \midrule
        \multirow{4}{*}{Qwen2.5-14B} 
        & \multicolumn{1}{l}{Cochain} & 75.18 & 75.00 & 75.38 & 13.11 & 15.08 & 48.35 & 19.70 & 20.30 \\
        & \multicolumn{1}{l}{w/o \(\mathcal{CKG}\)} & 74.83 & \cellcolor{red!10}74.41 & 75.25 & 12.49 & 14.55 & \cellcolor{red!10}47.32 & \cellcolor{red!10}18.39 & 19.76 \\
        & \multicolumn{1}{l}{w/o Causal Chain} & \cellcolor{red!10}74.71 & 74.66 & \cellcolor{red!10}74.79 & \cellcolor{red!10}12.28 & \cellcolor{red!10}14.17 & 47.65 & 19.30 & \cellcolor{red!10}19.44 \\
        & \multicolumn{1}{l}{w/o Prompts Tree} & \cellcolor{red!30}73.01 & \cellcolor{red!30}73.01 & \cellcolor{red!30}73.02 & \cellcolor{red!30}9.98 & \cellcolor{red!30}12.25 & \cellcolor{red!30}42.72 & \cellcolor{red!30}14.15 & \cellcolor{red!30}18.00 \\
        \bottomrule
    \end{tabular}}
    \label{tabf6}
\end{table*}

\subsection{Integration of Cochain with Baselines}

We further investigate the integration of Cochain with existing baseline methods, specifically PMC and MedAgents, through experiments conducted on a pharmaceutical dataset using Claude-3.5-Haiku. As reported in Table~\ref{tabf7}, the combination of Cochain with PMC results in improvements of 2.62\% and 7.55\% in the BS-F and BLEU-4 scores, respectively. This outcome further substantiates the efficacy of Cochain in mitigating issues associated with excessive collaboration.

\begin{table}[h]
  \caption{The improvement of Cochain over other baselines on pharmaceutical datasets.}
  \centering
  \resizebox{\textwidth}{!}{
  \begin{tabular}{lcccccccc}
    \toprule
    Baseline & BS-F & BS-P & BS-R & BLEU-4 & GLEU & METEOR & ROUGE-2 & ROUGE-L \\
    \midrule
    PMC~\cite{zhang2025planning} & 66.51 & 66.06 & 66.97 & 5.17   & 12.33  & 20.31  & 5.48   & 13.56 \\
    PMC~\cite{zhang2025planning} + Cochain & \textbf{69.13} & \textbf{68.49} & \textbf{69.80} & \textbf{12.72}  & \textbf{18.48}  & \underline{27.01} & \textbf{6.93} & \textbf{15.82} \\
    MedAgents~\cite{tang2023medagents} & 66.23 & 66.02 & 66.48  & 7.17 & 14.02 & 22.59 & 5.43 & 14.32  \\
    MedAgents~\cite{tang2023medagents} + Cochain & \underline{67.45}  & \underline{66.91} & \underline{68.00}  & \underline{11.65} & \underline{17.80} & \textbf{27.06} & \underline{6.16} & \underline{15.74} \\
    \bottomrule
  \end{tabular}}
  \label{tabf7}
\end{table}

\subsection{Evaluation of Collaborative Knowledge Graph}

To evaluate the reliability of our final knowledge graph, we randomly sampled 2,000 triplets from it, consisting of 1,000 from explicit knowledge extraction and 1,000 from our counterfactual method. We used three core metrics for evaluation: Factual Correctness (Is the knowledge factually correct in the real world?),Task Relevance (How helpful is the knowledge for solving business workflow problems?), and Knowledge Depth (Is the knowledge common sense, or does it require professional insight?) We invited two domain experts for a blind review. We also used Gemini-2.5-Pro as a judge for an objective evaluation. The average scores (out of 5) are presented in the Table~\ref{tab:evaluation_scores}.

\begin{table}[h!]
\centering
\caption{Evaluation of Collaborative Knowledge Graphs by Human Experts and LLMs.}
\label{tab:evaluation_scores}
\begin{tabular}{lcc}
\toprule
\textbf{Evaluation Metric} & \textbf{Human Expert Score (Avg. / 5.0)} & \textbf{Gemini-2.5-Pro Score (Avg. / 5.0)} \\
\midrule
Factual Correctness & 4.99 & 5.00 \\
Task Relevance      & 4.76 & 4.81 \\
Knowledge Depth     & 4.31 & 4.54 \\
\bottomrule
\end{tabular}
\end{table}

\subsection{Excessive collaboration is More Serious than Insufficient collaboration}

Business workflow tasks are particularly susceptible to the ``excessive collaboration'' phenomenon. When multiple agents collaborate, this manifests as responses deviating from core issues and reducing answer quality. Comparative analysis of Table 1 (Main Paper), Figure 6 (Main Paper), and additional single-agent experimental results (Supplementary Material~\ref{app:third.1}) demonstrates that, despite consuming significantly more computational resources, excessive collaboration performs substantially worse than insufficient collaboration. excessive collaboration's challenging nature is its covertness—systems maintain high activity and apparent collaboration, making the problem hard to detect and rectify promptly. While task decomposition, such as approaches like PMC, and summarization, such as strategies like CoA, are considered effective methods for focusing on core issues~\cite{wu2024autogen,chen2024reconcileroundtableconferenceimproves}, Cochain's collaboration mechanism exhibits superior performance in controlling excessive collaboration, enabling a more precise focus on critical task requirements.

\subsection{Robustness to Incomplete Retrieval}
To validate the system's robustness against potential keyword retrieval failures, we conducted a stress test. We designed 50 structured queries in the format of ``How to use [Core Concept A] to solve problems related to [Target Concept B]''. In the test condition, we deliberately masked all keywords corresponding to [Target Concept B] to simulate a first-stage retrieval failure. For instance, in the query ``How can ergonomics be used to reduce traffic accident injuries?'', the keywords ``traffic'' and ``accident'' were blocked.

While end-to-end semantic retrieval could theoretically solve keyword-mismatch issues, it becomes computationally prohibitive for real-time interaction on massive knowledge graphs. We therefore adopt the Retrieval-Ranking two-stage paradigm, a proven industry standard~\cite{10.1145/2959100.2959190}. This approach combines the advantages of both methods: keyword-based retrieval in the first stage ensures precise matching of technical terms and proper nouns, avoiding omissions from semantic over-generalization. In the second stage, semantic ranking captures the query's overall intent, ensuring system generalization and robustness.

\begin{table}[h]
\centering
\caption{Performance comparison under simulated retrieval failure. Masking the target concept's keywords results in a negligible performance drop, demonstrating the system's robustness.}
\label{tab:robustness_test}
\begin{tabular}{lccc}
\toprule
\textbf{Test Condition} & \textbf{BS-F} & \textbf{GLEU} & \textbf{ROUGE-L} \\
\midrule
Full Query & 74.16 & 22.08 & 25.86 \\
Target Concept Masked & 73.87 & 21.42 & 24.57 \\
\midrule
\textbf{Performance Drop (\%)} & \textbf{-0.29\%} & \textbf{-0.66\%} & \textbf{-1.29\%} \\
\bottomrule
\end{tabular}
\end{table}

The results in Table~\ref{tab:robustness_test} confirm the effectiveness of this design. Even when forcing a first-stage failure, the BS-F score experienced a minimal drop of only 0.29\%, from 74.16 to 73.87, with similarly small decreases in GLEU (-0.66\%) and ROUGE-L (-1.29\%) scores. This resilience is attributed to two factors. First, as justified above, the semantic ranking stage corrects for an imperfect initial candidate set. More critically, the subsequent graph traversal and causal chain stages provide a powerful secondary path to completeness. Even if ``Traffic Accident'' is missed during retrieval, it can be discovered through pre-existing knowledge paths represented as triples (e.g., (Injuries, is\_a\_result\_of, Traffic Accident)), ensuring the final context remains robust and comprehensive.

%\section{Broader Impacts}
%\label{app:impact}
%Cochain enhances agent efficiency and effectiveness in complex business workflows, potentially increasing productivity and enabling agent systems to tackle more sophisticated multi-step tasks across various industries. By addressing both under-collaboration and over-collaboration, Chain-of-Collaboration contributes to more robust and reliable agent systems. The interpretability offered by our framework, for instance through the causal chain mechanism, can also aid in scrutinizing agent behavior, potentially reducing misuse.
%Cochain's design focuses on improving collaboration quality rather than solely on automation. Nevertheless, the broader societal implications of increased automation in workflows, where more tasks are executed by agent systems, warrant ongoing consideration. For the responsible dissemination of our data, which is created using Self-Instruct and consists of generalized, non-sensitive information that does not contain Personally Identifiable Information or private commercial data, we will provide comprehensive documentation, a clear academic research license stipulating ethical use, and a designated point of contact for inquiries.

\section{Case Study}
\label{app44444}
\subsection{Interpretability Analysis}

We investigate the interpretability characteristics of the model after the application of Cochain through a case study. As shown in Table~\ref{tabf8}, we employ the same color to annotate logically related content visually. The findings reveal that, following the application of Cochain, the model is capable of performing multi-dimensional reasoning based on knowledge graphs, causal inference chains, and contextual cues when addressing user needs, thereby generating outputs that are traceable, reliable, and interpretable. The application of the Cochain method has significantly enhanced the transparency and explainability of the model's reasoning process, providing empirical evidence for understanding the model's collaborative decision-making mechanisms.

\subsection{Comparative Analysis of Model Outputs}

\begin{figure}[h] 
    \centering
    \includegraphics[width=0.9\textwidth]{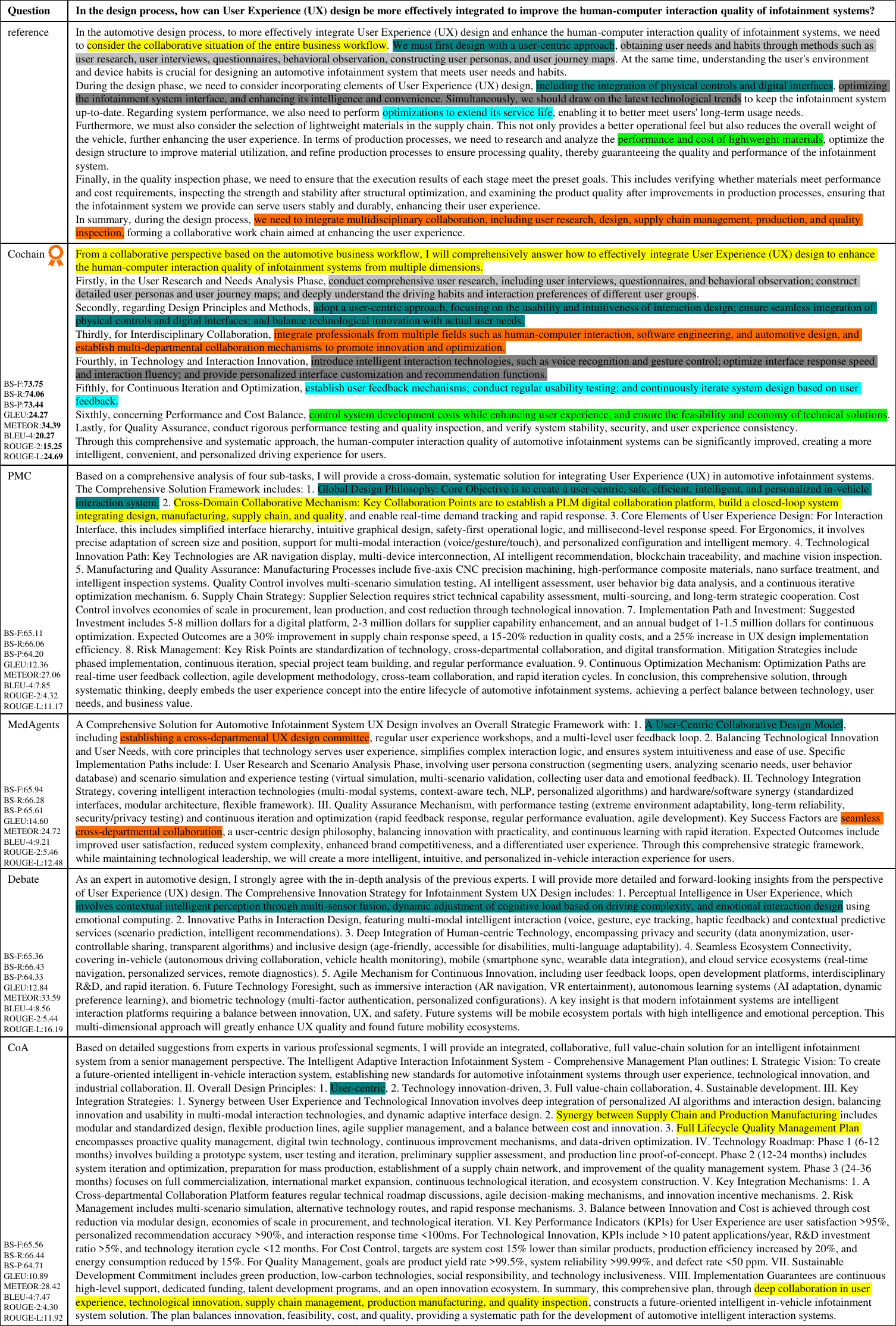}
    \caption{Case study on the Automotive dataset. Passages aligning with the reference answer are highlighted in the same color. The left side of the figure lists the scores for each baseline.} 
    \label{casestudy11}
\end{figure}

\begin{figure}[h] 
    \centering
    \includegraphics[width=0.9\textwidth]{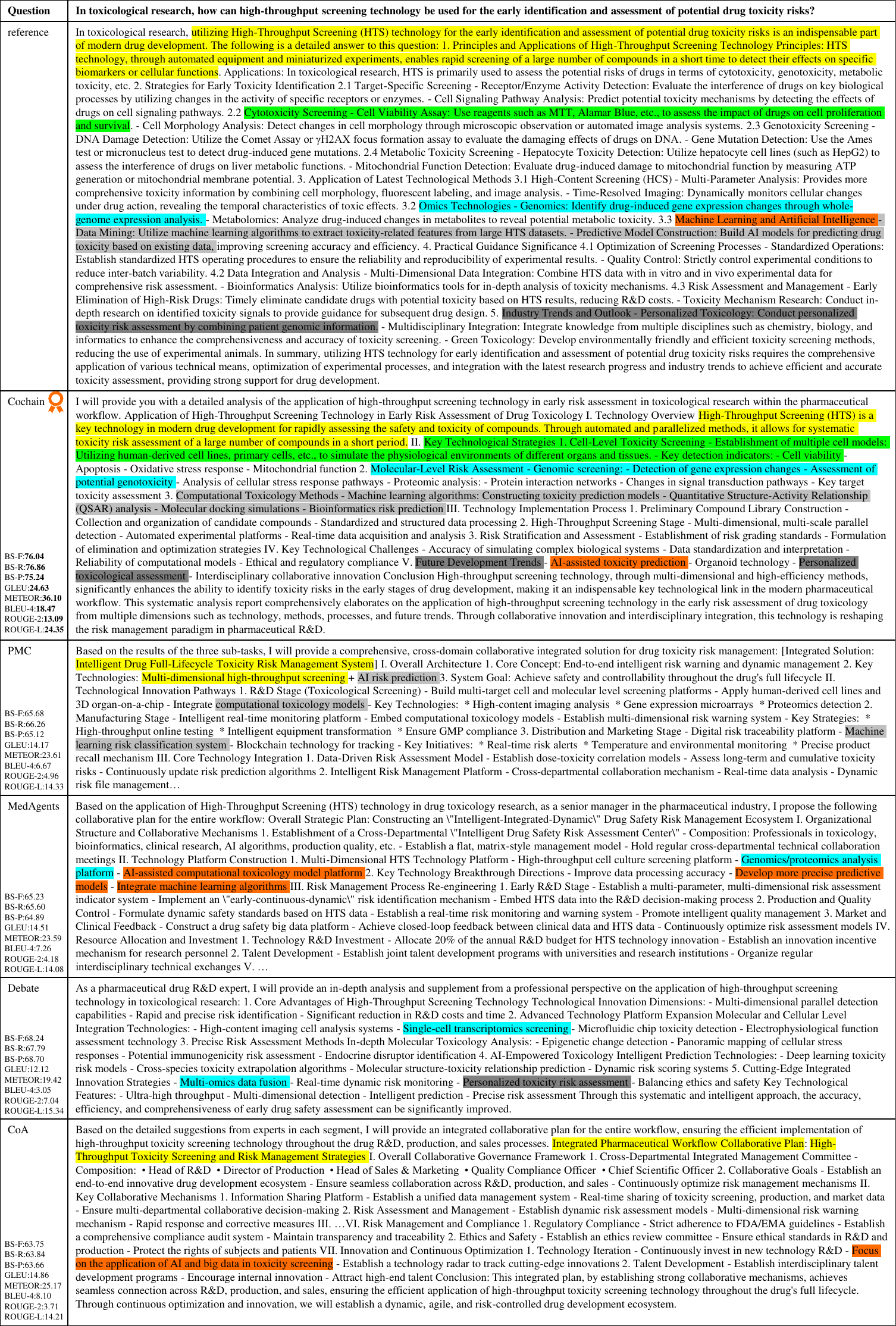}
    \caption{Case study on the Pharmaceutical dataset. Passages aligning with the reference answer are highlighted in the same color. The left side of the figure lists the scores for each baseline.} 
    \label{casestudy12}
\end{figure}

\begin{figure}[h] 
    \centering
    \includegraphics[width=0.9\textwidth]{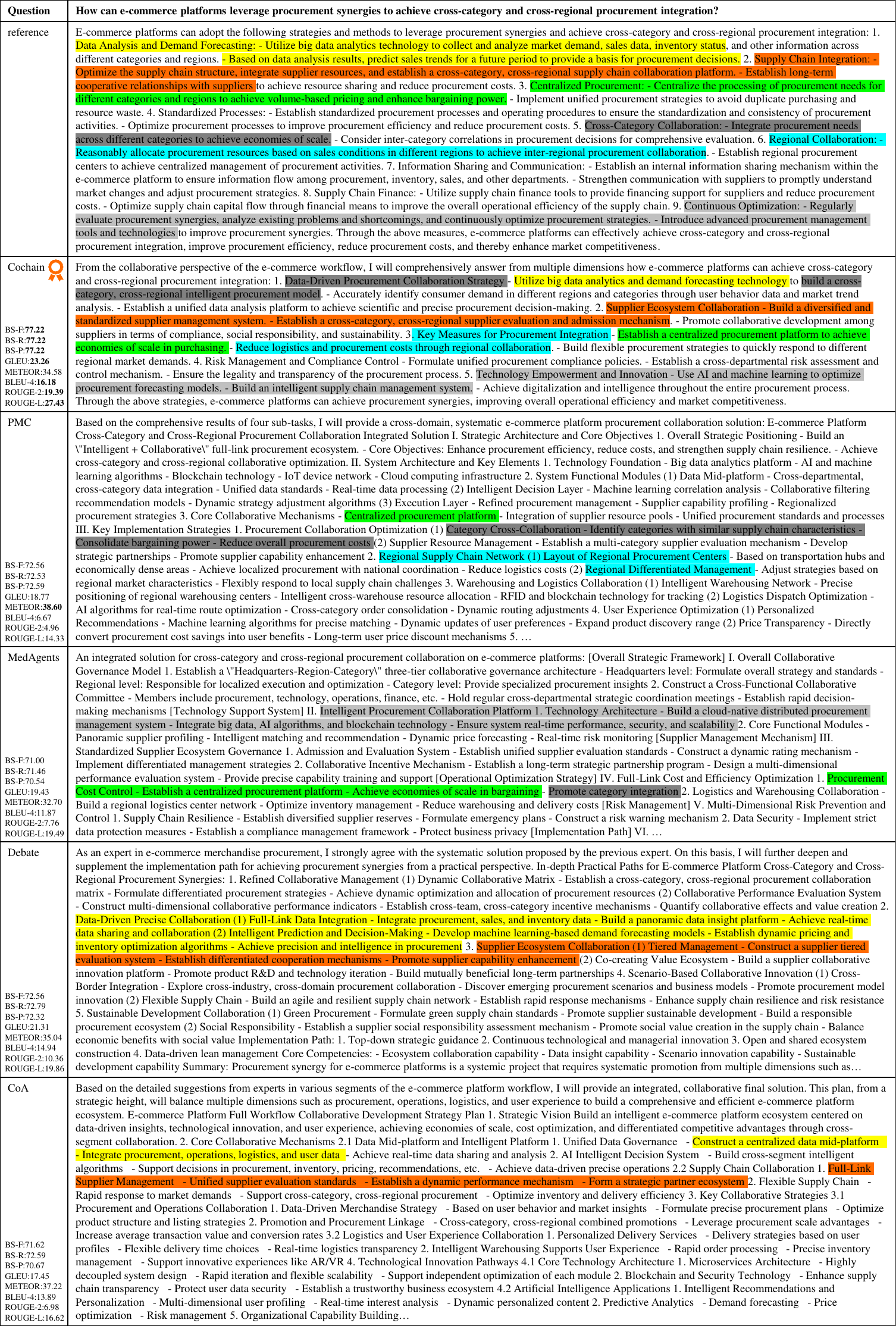}
    \caption{Case study on the E-commerce dataset. Passages aligning with the reference answer are highlighted in the same color. The left side of the figure lists the scores for each baseline.} 
    \label{casestudy13}
\end{figure}

To understand how different baselines handle complex instructions, as shown in Figure~\ref{casestudy11}, we qualitatively analyzed their outputs against a reference answer, focusing on their integration of User Experience(UX) design within the automotive business workflow. Additionally, we present case studies for the pharmaceutical business workflow and the e-commerce business workflow, as shown in Figure~\ref{casestudy12} and Figure~\ref{casestudy13}.

\paragraph{Cochain: Clear Structure, Focused on UX Integration Process.} 
Cochain's response was notably well-structured, using a seven-point list for key UX integration stages, enhancing readability and providing a clear action framework. It comprehensively covered the UX lifecycle—from user research and design principles to interdisciplinary collaboration, technological innovation, continuous iteration, and performance-cost/quality assurance. Its explicit ``collaborative perspective of the automotive business workflow'' aligned with the reference answer's holistic view, effectively capturing its spirit by offering a structured, UX-centric methodology.

\paragraph{Comparison of Cochain with Other Model Responses.} 
Cochain's primary strength is its sustained focus on the UX design integration process and principles, closely matching the prompt's core intent and the reference answer. In contrast, PMC and CoA lean towards high-level strategic planning (e.g., PLM platforms, KPIs), emphasizing system management rather than the deep UX integration across workflow stages highlighted by the reference answer. Debate showcases significant depth in UX concepts and cutting-edge technologies. However, its primary focus is on HMI innovation details, rather than systematically integrating UX across broader business workflow processes. Cochain can better balance innovation with this workflow integration. MedAgents share some structural similarities with Cochain, but Cochain is more direct in articulating the end-to-end UX lifecycle, especially ``Continuous Iteration and Optimization,'' and its business workflow perspective. While other models excel in specific areas (e.g., Debate's UX tech depth; PMC and CoA's strategic scope), Cochain demonstrates the closest overall alignment with the reference answer's advocated UX-integrated process throughout the business workflow and its pragmatic perspective, thus performing excellently in this qualitative assessment.

\begin{table}[h]
\small
\centering
\caption{Case study of interpretability. The three distinct modules, namely Knowledge, Causal Chain, and Prompts Tree, are each assigned one of the three different color schemes: \sethlcolor{red!40}\hl{red}, \sethlcolor{yellow!80}\hl{yellow}, and \sethlcolor{blue!40}\hl{blue}. The colors within the modules correspond to those in the output, and logically related content is annotated using the same color scheme.}
\begin{tabularx}{\textwidth}{l|X}
\toprule
\textbf{User need} & 
How can cost optimization be achieved in automotive manufacturing? \\ \midrule

\textbf{Knowledge} & 
Cost optimization points include technological innovation, \sethlcolor{red!20}\hl{quality control}, and environmental and energy management. Design optimization, material selection, and \sethlcolor{red!40}\hl{supply chain network construction} strategies are measures for cost optimization... \\ \midrule

\textbf{Cause chain} & 
\sethlcolor{yellow!80}\hl{Stamping} is primarily used for component forming and can be integrated with \colorbox{yellow!20}{automated and} \sethlcolor{yellow!20}\hl{intelligent production lines}, \sethlcolor{yellow!50}\hl{intelligent monitoring and optimization systems}, as well as \sethlcolor{yellow!80}\hl{smart mold maintenance}... \\ \midrule

\textbf{Prompts tree} & 
\textbf{Prompt 1:} User Need: Cost optimization $\rightarrow$ Design Method: \sethlcolor{blue!20}\hl{Accelerating prototype iteration} $\rightarrow$ Supply Chain Method: \sethlcolor{blue!10}\hl{Selection of lightweight materials} $\rightarrow$ Production Method: \sethlcolor{blue!40}\hl{Research and} \sethlcolor{blue!40}\hl{analysis of lightweight material performance and cost} $\rightarrow$ Quality Inspection Method: \sethlcolor{blue!30}\hl{ensuring materials meet performance and cost requirements.} \\ & 
\textbf{Prompt 2:} User Need: Cost optimization $\rightarrow$ Design Method: \sethlcolor{blue!20}\hl{Accelerating prototype iteration} $\rightarrow$ Supply Chain Method: \sethlcolor{blue!10}\hl{Selection of lightweight materials}  $\rightarrow$ Production Method: Design optimization to enhance material utilization $\rightarrow$ Quality Inspection Method: Testing the strength and stability of the optimized structure. \\  & 
...\\
\midrule

\textbf{Output} & 
In automotive design, it is essential to \sethlcolor{blue!20}\hl{consider prototype optimization} and innovation in design methodologies. For instance, to address customer-specific vehicle customization, cost control during the design phase can be achieved by \sethlcolor{blue!20}\hl{accelerating prototype iterations} and \sethlcolor{blue!10}\hl{utilizing lightweight materials}. Additionally, the properties and \sethlcolor{blue!30}\hl{costs of lightweight materials} should be taken into account to optimize structural design and enhance material utilization.

In supply chain management, cost optimization can also be achieved. This includes strategies for \sethlcolor{red!40}\hl{building the supply chain network} and optimizing logistics. By consolidating procurement and leveraging economies of scale, procurement costs can be reduced. Collaborative Planning, Forecasting, and Replenishment (CPFR) methods can further optimize inventory costs.

In the manufacturing stage, cost optimization can be achieved through the introduction and implementation of  \sethlcolor{yellow!20}\hl{automated and intelligent production lines}. For example, the use of automated production lines, coupled with \sethlcolor{yellow!50}\hl{intelligent monitoring and optimization systems}, can reduce redundant tasks and increase production efficiency. Furthermore, for specific manufacturing processes, such as \sethlcolor{yellow!80}\hl{stamping}, \sethlcolor{blue!40}\hl{material performance costs can be analyzed}, and \sethlcolor{yellow!80}\hl{intelligent molds and automatic maintenance systems} can be developed to optimize costs.

Finally, in the quality inspection stage, ensuring that all materials and products meet the predetermined performance and cost requirements is crucial. Additionally, improving production processes to \sethlcolor{red!20}\hl{ensure product processing quality} is an important aspect of quality inspection.\\ \bottomrule

\end{tabularx}
    \label{tabf8}
\end{table}

\clearpage

\section{Prompts Template}
\label{app:template}

We provide the prompt templates for all multi-agent baselines presented in the paper, exemplified using the automotive business workflow.
\subsection{Cochain}

\begin{figure}[h] \centering
    \includegraphics[width=\textwidth]{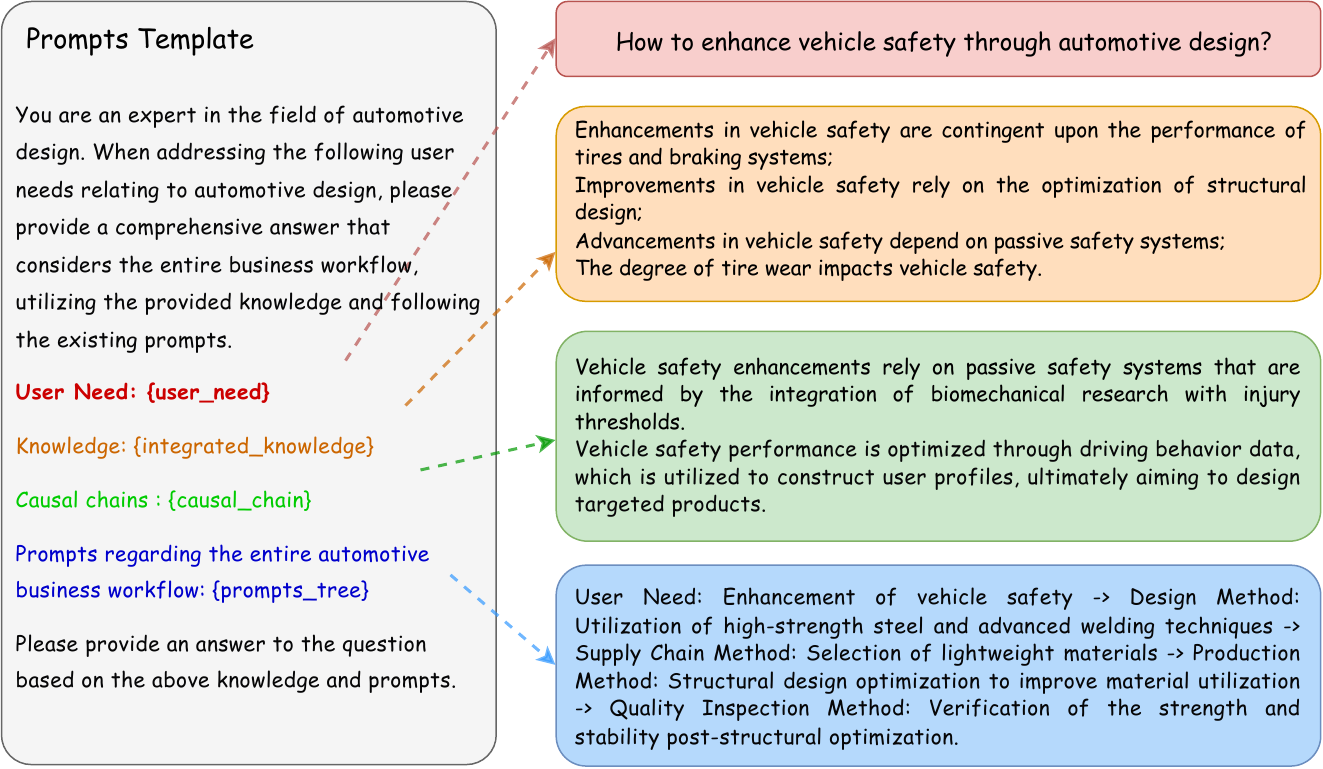}
    \caption{The prompts template of Cochain for final input to LLM.} 
    \label{tme1}
\end{figure}

We present a specific overview of the prompts template in the Cochain. As illustrated in Figure~\ref{tme1}, the prompt template adopts a modular design comprising three core components: the knowledge module retrieved from the collaborative knowledge graph, the causal chain module, and the business workflow prompts module generated through prompts tree retrieval. In the template, the gray box represents fixed prompt content, while the colored boxes denote dynamic content that adapts based on user needs.

\clearpage

\subsection{PMC}
\begin{figure}[h!] 
    \centering
    \includegraphics[width=\textwidth]{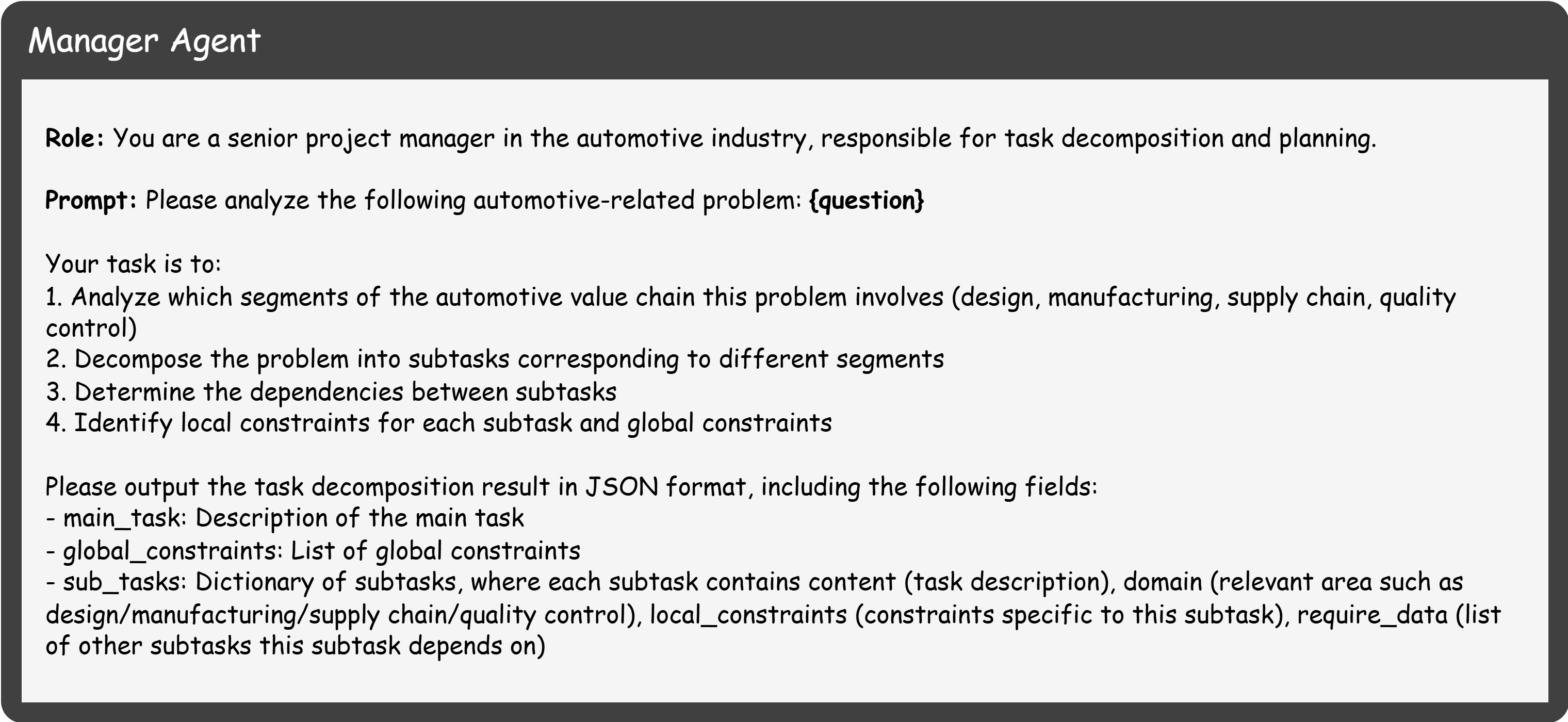}
    \caption{The manager agent prompts template of PMC.} 
\end{figure}

\begin{figure}[h!] 
    \centering
    \includegraphics[width=\textwidth]{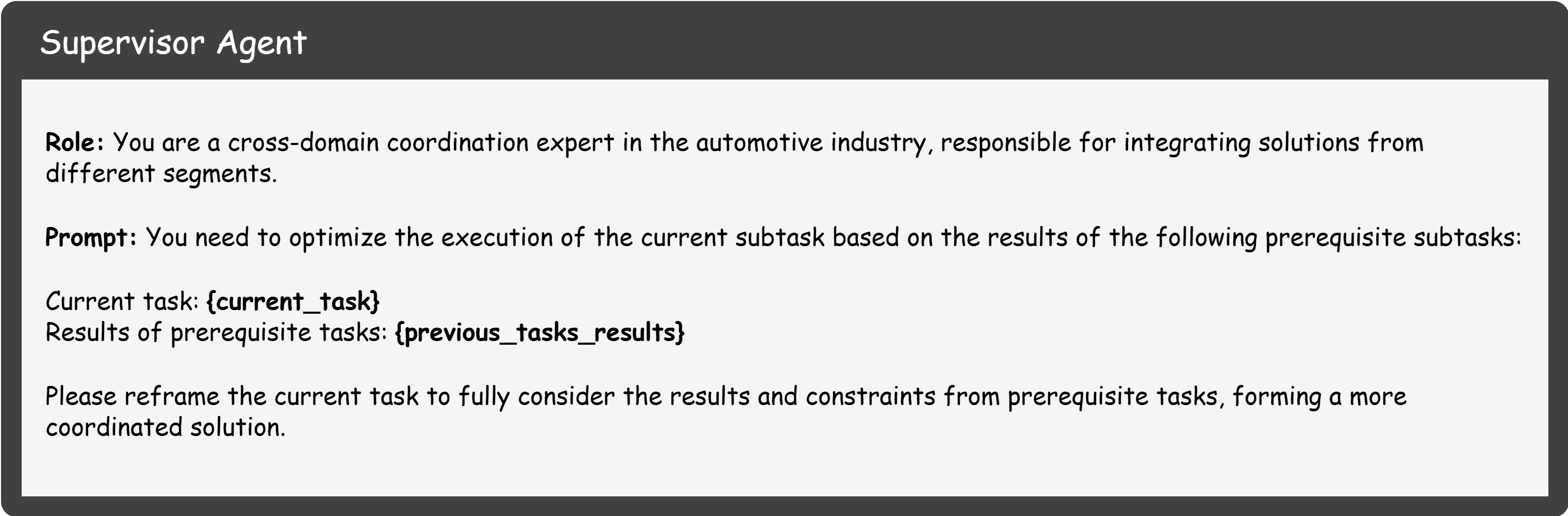}
    \caption{The supervisor agent prompts template of PMC.} 
\end{figure}

\begin{figure}[h!] 
    \centering
    \includegraphics[width=\textwidth]{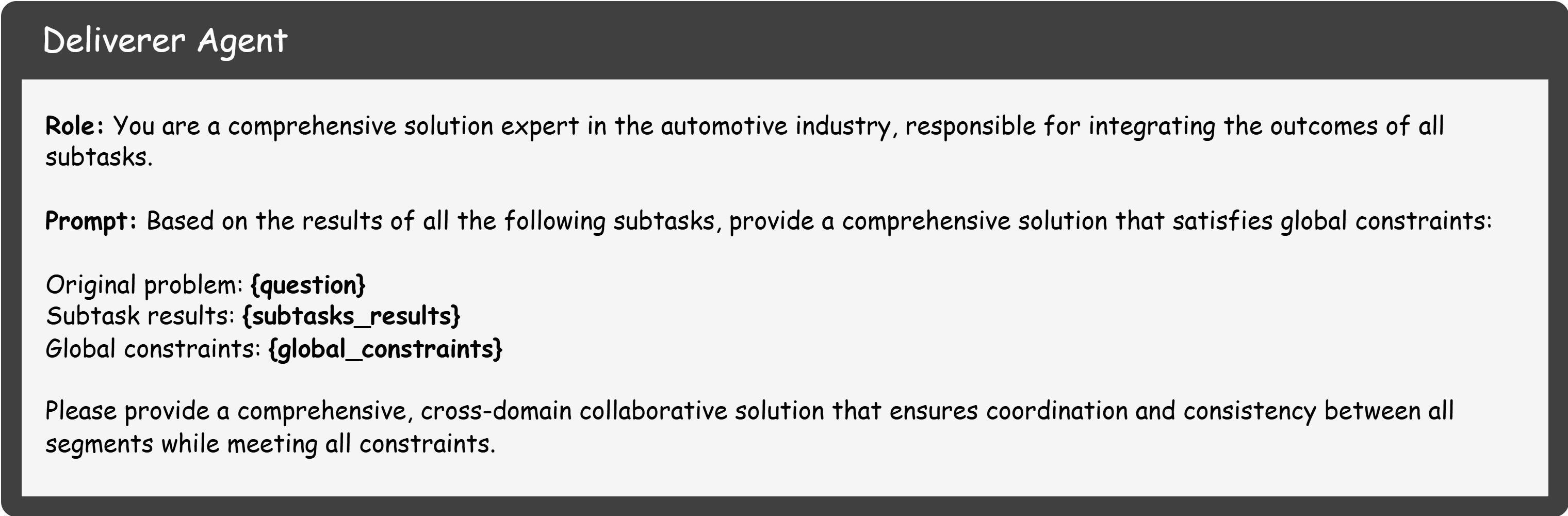}
    \caption{The deliverer agent prompts template of PMC.} 
\end{figure}

\begin{figure}[h!] 
    \centering
    \includegraphics[width=\textwidth]{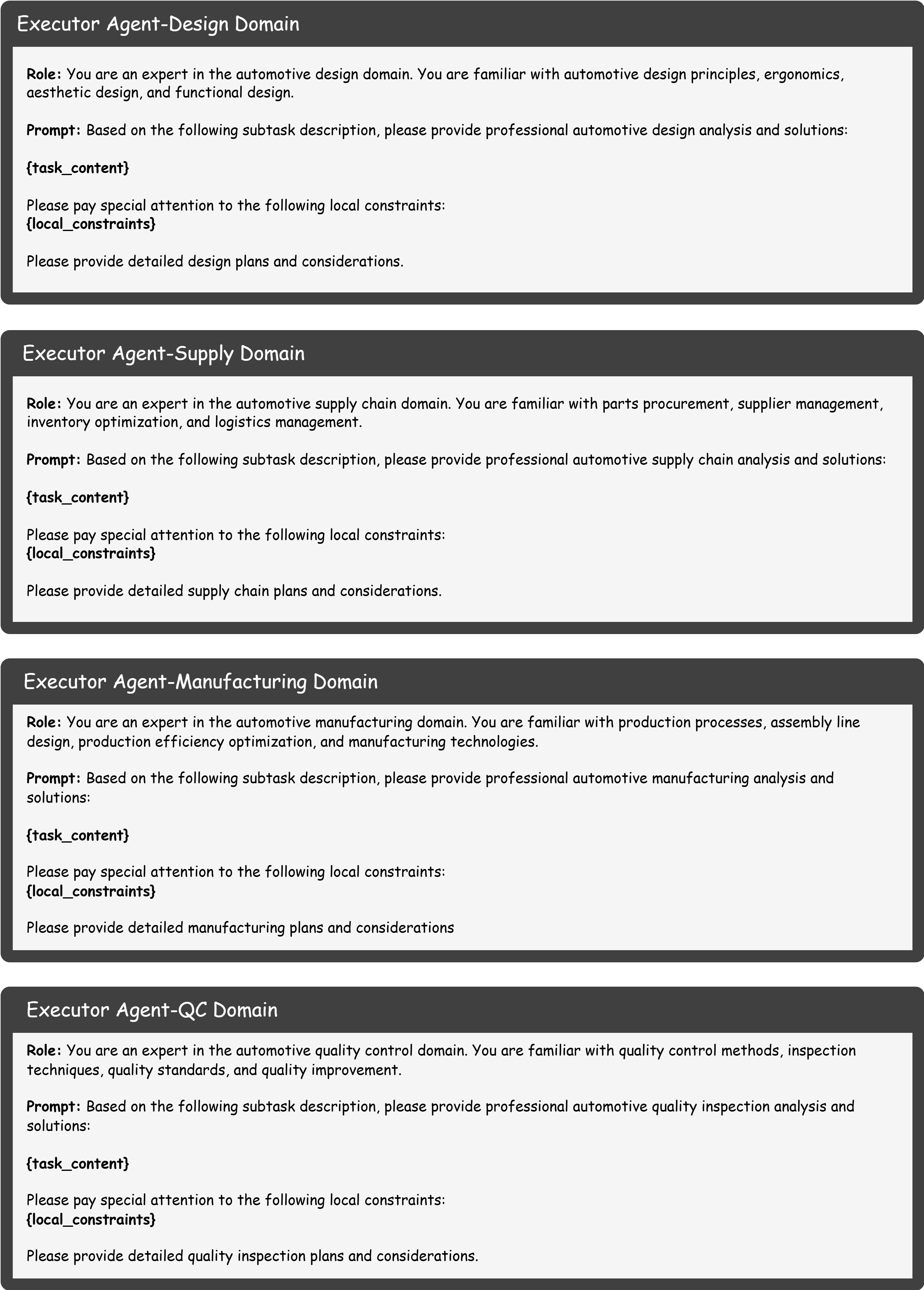}
    \caption{The executor agent prompts template of PMC.} 
\end{figure}

\clearpage

\subsection{MedAgents}

\begin{figure}[h!] 
    \centering
    \includegraphics[width=\textwidth]{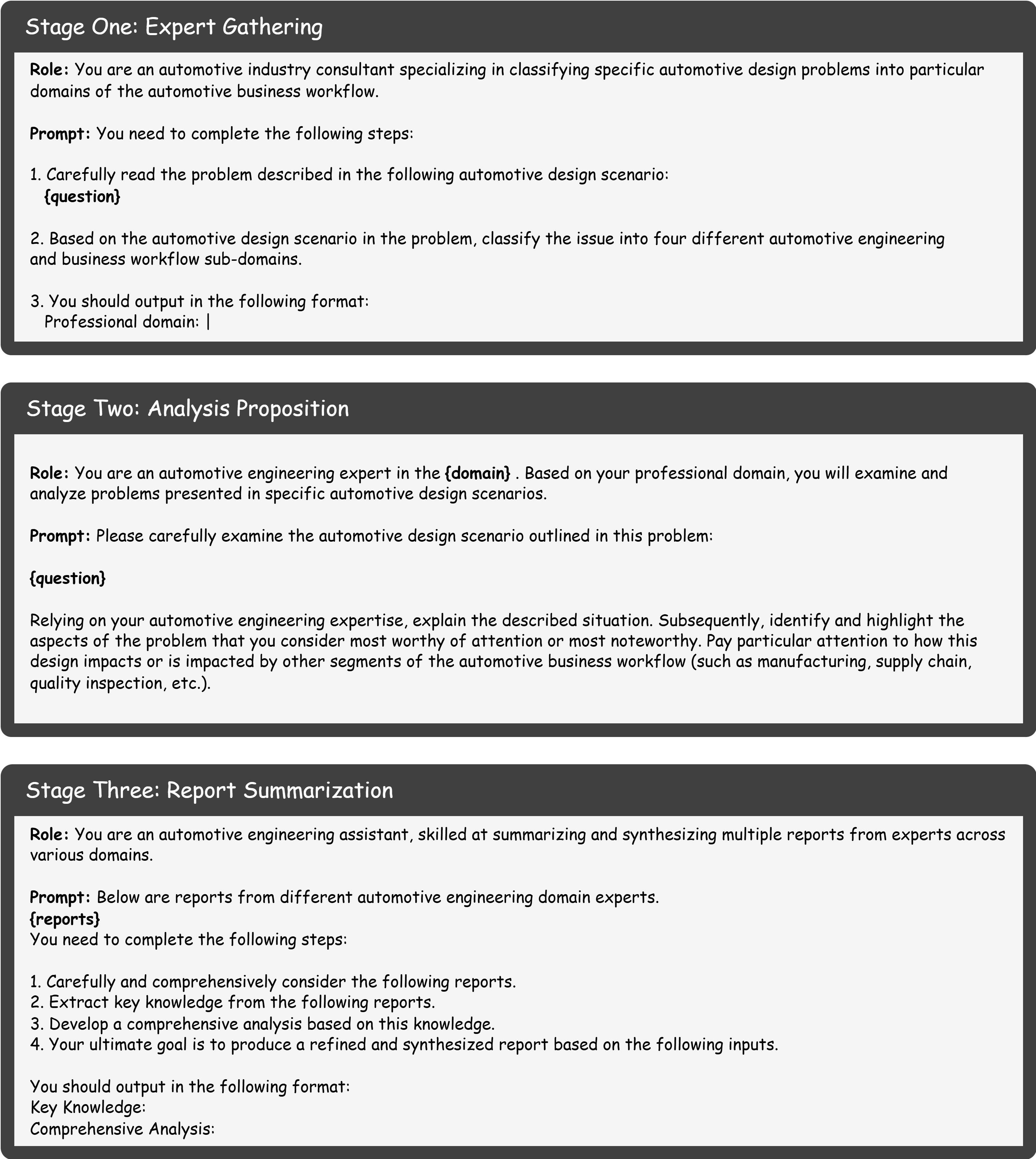}
    \caption{The prompt templates for stages 1 through 3 of MedAgents.} 
\end{figure}

\begin{figure}[h!] 
    \centering
    \includegraphics[width=\textwidth]{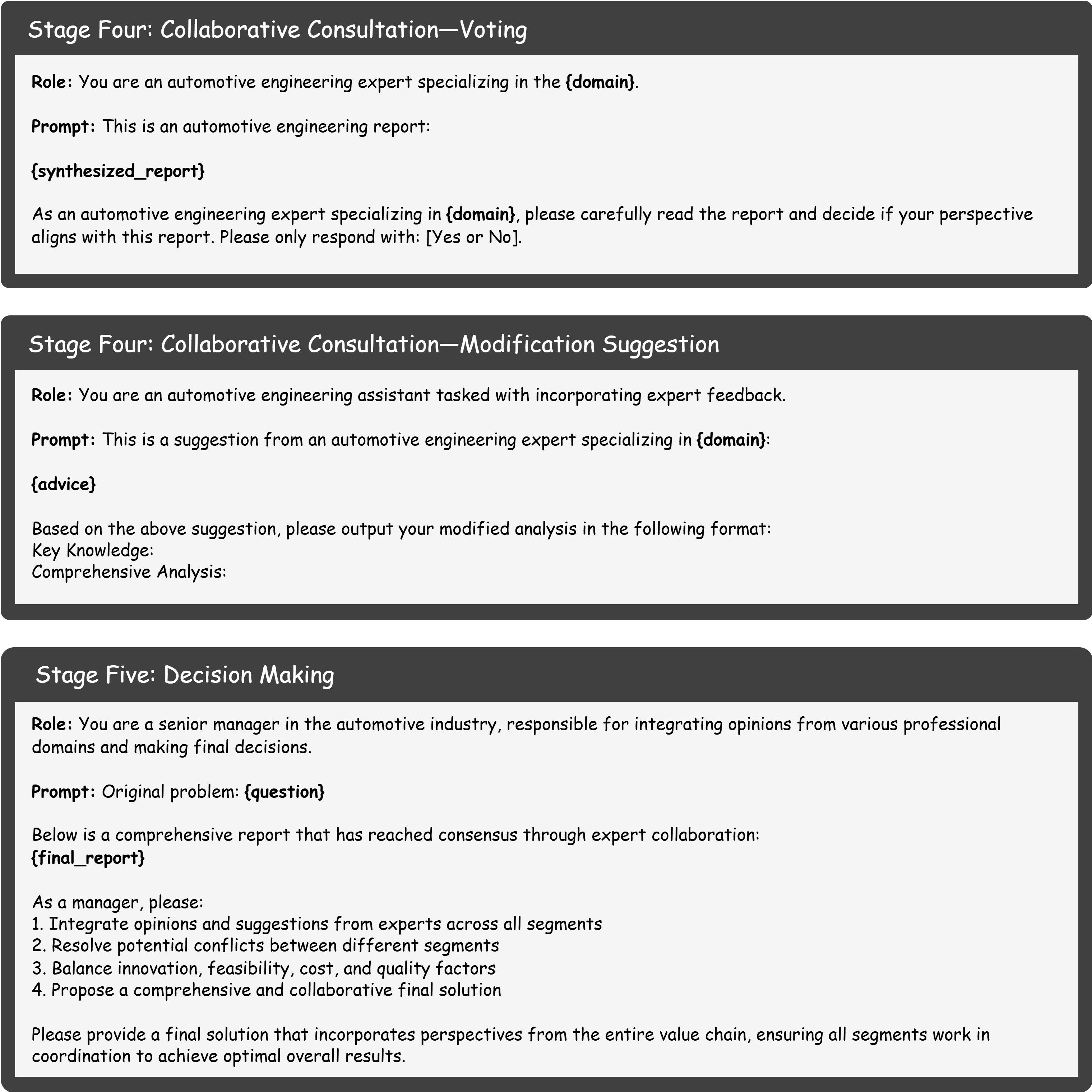}
    \caption{The prompt templates for stages 4 and 5 of MedAgents.} 
\end{figure}

\clearpage

\subsection{Debate}

\begin{figure}[h!] 
    \centering
    \includegraphics[width=\textwidth]{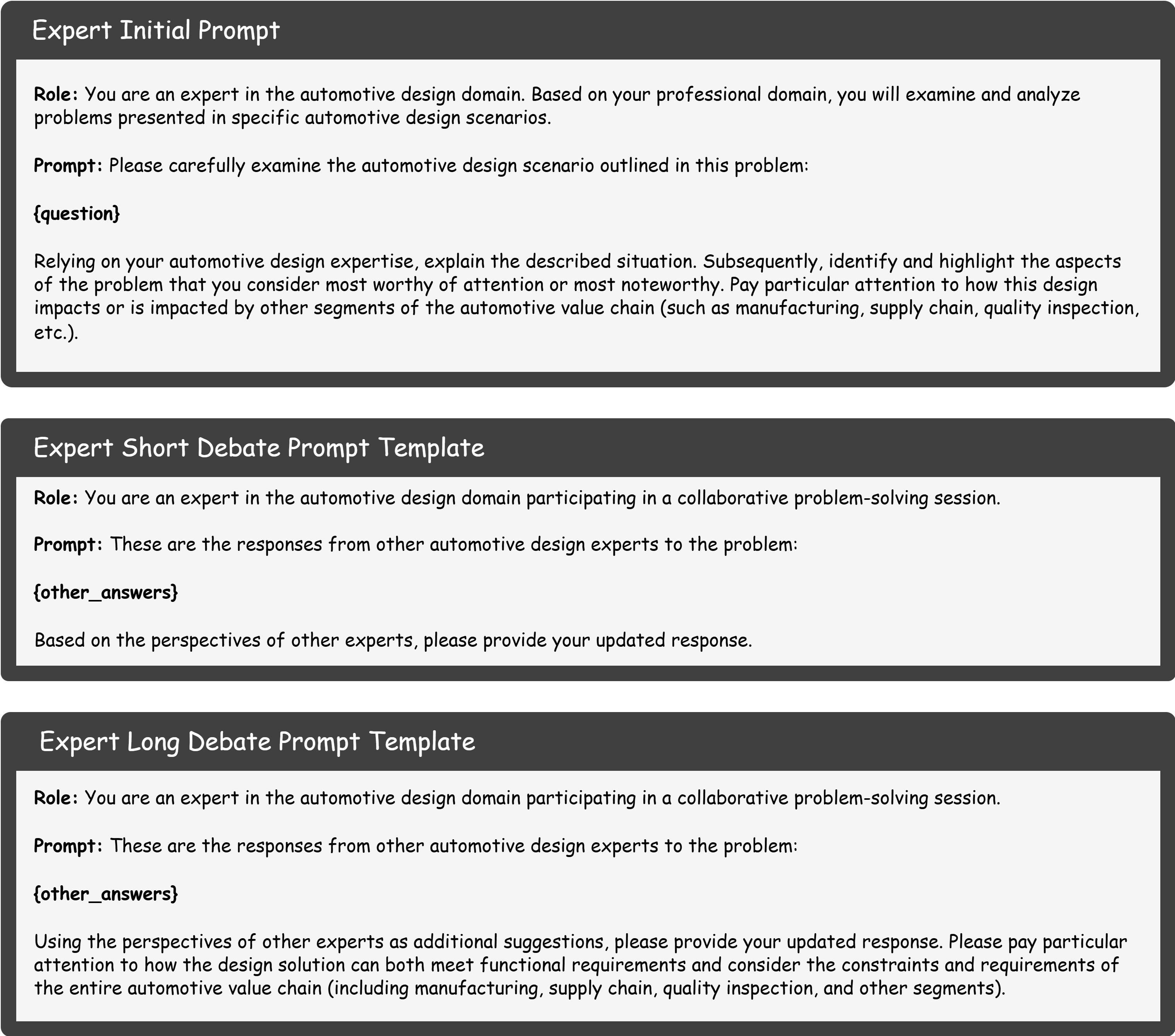}
    \caption{The prompts template of Debate for final input to LLM.} 
\end{figure}

\clearpage

\subsection{CoA}

\begin{figure}[h!] 
    \centering
    \includegraphics[width=\textwidth]{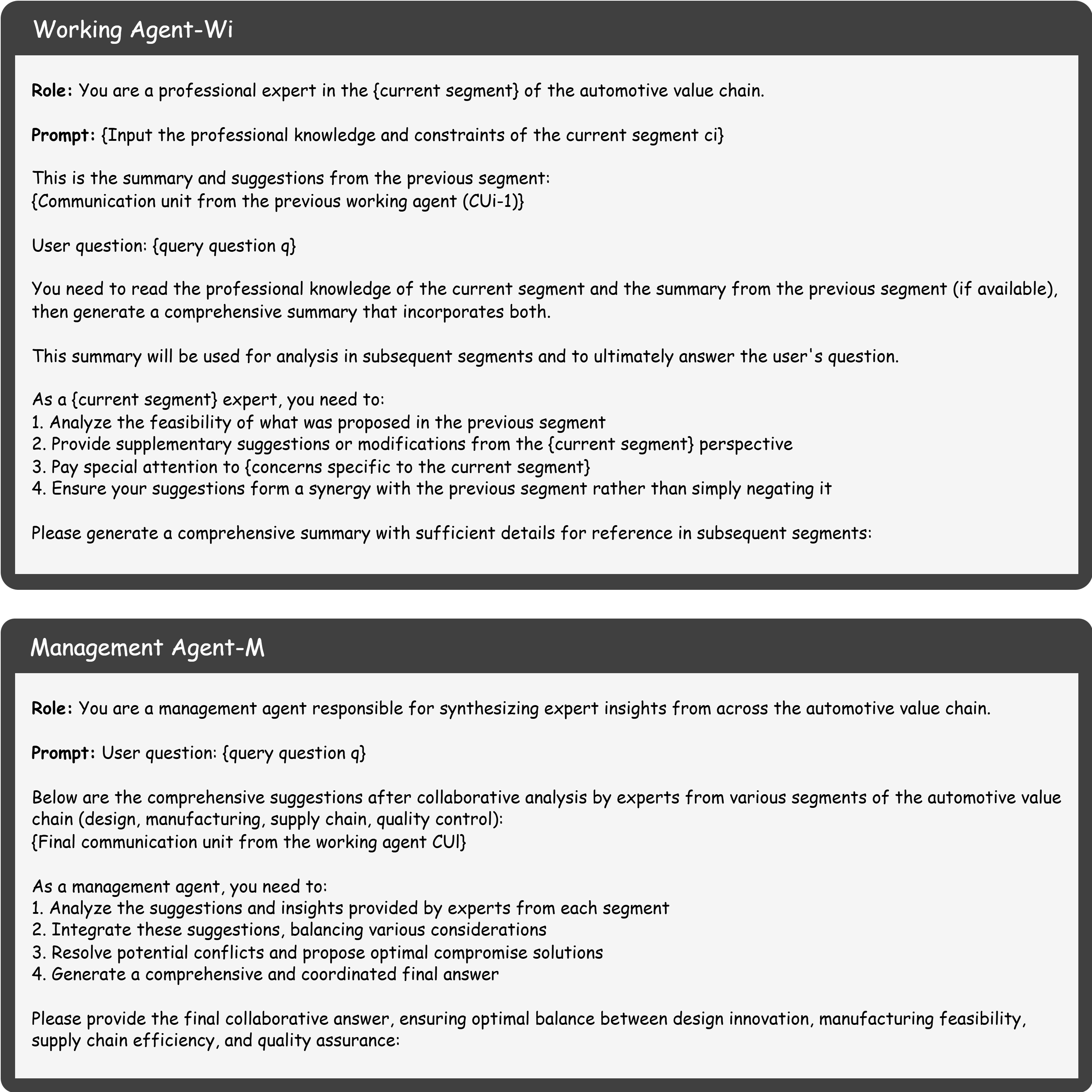}
    \caption{The prompt templates for the worker agent and manager agent of CoA.} 
\end{figure}

\begin{figure}[h!] 
    \centering
    \includegraphics[width=\textwidth]{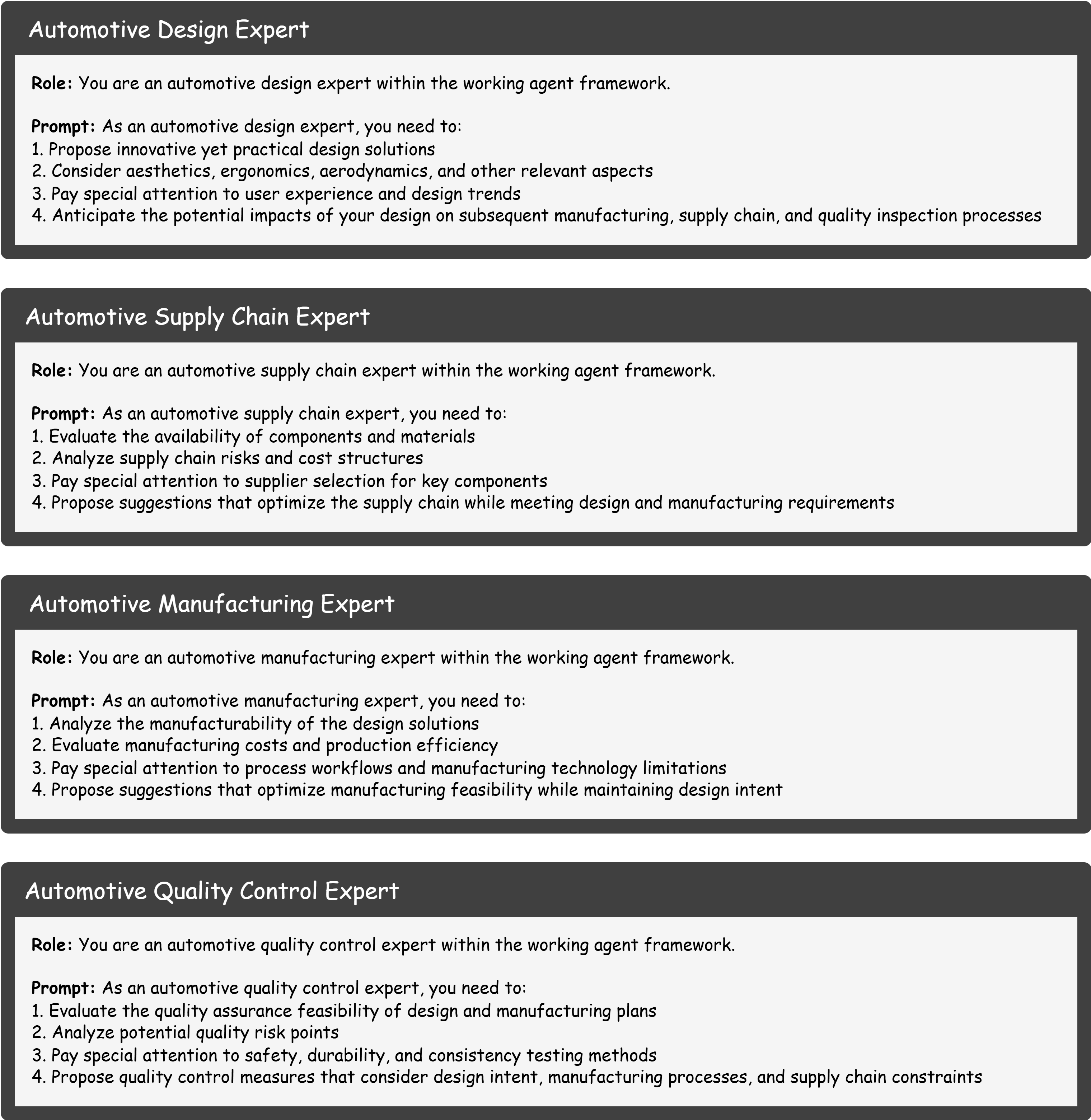}
    \caption{The prompt templates for each worker agent on the Automotive dataset.} 
\end{figure}

\end{document}